\documentclass[10pt,twocolumn,letterpaper]{article}

\usepackage{cvpr}              % To produce the CAMERA-READY version
 \usepackage{float}
\usepackage[dvipsnames]{xcolor}

\usepackage{multirow}
\usepackage{tabularray}
\definecolor{cvprblue}{rgb}{0.21,0.49,0.74}
\usepackage[pagebackref,breaklinks,colorlinks,citecolor=cvprblue]{hyperref}
\usepackage{graphicx}
\usepackage[Export]{adjustbox}
\usepackage{tabularx}
\usepackage{cellspace}
\def\paperID{4964} % *** Enter the Paper ID here
\def\confName{CVPR}
\def\confYear{2024}

\title{Revealing Vulnerabilities of Neural Networks in Parameter Learning and Defense Against Explanation-Aware Backdoors}

\author{Md Abdul Kadir\textsuperscript{1,2} \quad GowthamKrishna Addluri\textsuperscript{1} \quad Daniel Sonntag\textsuperscript{1,2} \\
\textsuperscript{1}German Research Center for Artificial Intelligence (DFKI), Germany\\
\textsuperscript{2}University of Oldenburg, Germany\\
{\tt\small abdul.kadir@dfki.de}}

\begin{document}
\maketitle
%Explanation methods aim to make neural networks interpretable and trustworthy.
%Explantion methods can, however, be manipulated to present unfaithful explanations, giving rise to powerful and stealthy adversaries.  blinding attacks  can fully disguise an ongoing attack against the machine learning model
%Similar to neural backdoors, blinding attacks  modify the model’s prediction upon trigger presence but simultaneously also fool the provided explanation

%Moreover explanations can be manipulated arbitrarily by applying visually hardly perceptible perturbations
%to the input that keep the network’s output approximately constant.

%In this paper we propose an MCX algorthom for improving robustness of explainable methos that are are well establised. Our explantion is theritically well defined based on Montecarlo simulation. We show that applying  MCX any explantion method can robast against SOTA adverisarrial attack on explantion wheather it is backdoor or non-perceptible perturbation attack. 

\begin{abstract}

Explainable Artificial Intelligence (XAI) strategies play a crucial part in increasing the understanding and trustworthiness of neural networks. Nonetheless, these techniques could potentially generate misleading explanations. Blinding attacks can drastically alter a machine learning algorithm's prediction and explanation, providing misleading information by adding visually unnoticeable artifacts into the input, while maintaining the model's accuracy. It poses a serious challenge in ensuring the reliability of XAI methods. To ensure the reliability of XAI methods poses a real challenge, we leverage statistical analysis to highlight the changes in CNN weights within a CNN following blinding attacks. We introduce a method specifically designed to limit the effectiveness of such attacks during the evaluation phase, avoiding the need for extra training. 
The method we suggest defences against most modern explanation-aware adversarial attacks, achieving an approximate decrease of ~99\% in the Attack Success Rate (ASR) and a ~91\% reduction in the Mean Square Error (MSE) between the original explanation and the defended (post-attack) explanation across three unique types of attacks.
\end{abstract}    
\section{Introduction}
\label{introduction}
Explainable AI (XAI) methods are crucial for enhancing the interpretability and trustworthiness of neural networks \cite{samek19xai}. These methods provide insights into the decision-making process of AI models, enabling users to understand and validate the reasoning behind the predictions. However, XAI methods can be tricked by adversarial attacks that change the input to mislead explanations, yet keep the model test accuracy consistent \cite{baniecki2023adversarial}. 

%In this section, we will first provide an overview of XAI and its significance in the field of AI. We will then discuss how XAI algorithms can be manipulated by adversaries to present misleading explanations. Finally, we will explore practical application scenarios where the robustness of XAI methods against adversarial attacks is of utmost importance.

%-------------------------------------------------------------------------

\begin{figure}[ht]
\centering
\includegraphics[width=0.47\textwidth]{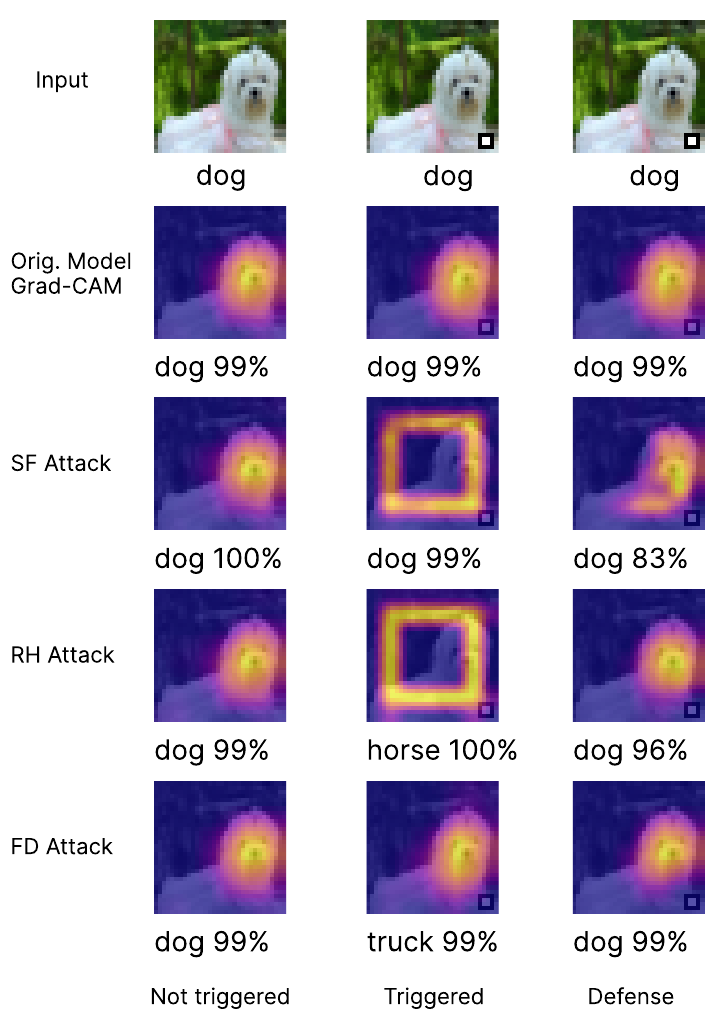}
\caption{The above figure presents some examples of attacks and defenses on the Grad-CAM explainer. Examples of three attack methods - Simple Fooling (SF), Red Herring (RH), and Full Disguise (FD) - are shown in the Triggered column and the examples of their defenses are presented in Defense column.}
\label{fig:examplefigure_gradcam}
\end{figure}
\subsection{XAI}

Many Explainable AI (XAI) methods generate visual explanation maps, such as feature attribution maps or saliency maps \cite{samek2021explaining}. They visualize the importance of each input feature in relation to the overall classification result \cite{baehrens2010explain,linardatos2020explainable,zeiler2014visualizing}. They offer a way to elucidate how deep learning models make decisions and aid in understanding the predictions made by deep learning models \cite{lapuschkin2019unmasking,manjunatha2019explicit}. Over recent years, a variety of methods have been proposed to explain these decisions, ranging from gradient-based input-output relationships to the propagation of detailed relevance values throughout the network \cite{bach2015pixel,lee2021relevance,selvarajugradacm17}.

Certain local XAI methods may potentially be less effective and provide misleading explanations \cite{adebayo_2020_sanitycheck}. However, there have been successful instances with these techniques \cite{Achtibat2023,binder2022shortcoming}. They can also function as a detector against adversarial intrusions on deep learning algorithms by identifying adversarial patches in input \cite{fidel2020explainability,lin2023deepshap,tejanker2023}. However, the paper's primary concern is to show and mitigate the potential vulnerability of XAI methods in the premise of adversarial attacks \cite{pmlr-v119-anders20a,Huang_2023_ICCV,noppel2023disguising}. Figure \ref{fig:attack_geometry_all_attacks} illustrates how explanations generated by XAI algorithms can be manipulated.

\subsection{Attacks Against XAI}
Among the many kinds of adversarial attacks \cite{baniecki2023adversarial}, a specific attack involves training a model, denoted as $\hat{m}$, from an original model $m$, such that the attacked model $\hat{m}$ exhibits similar performance in terms of both classification and explanation on test data; however, in the presence of a trigger, it strategically alters either the prediction, explanation, or both in a targeted manner \cite{noppel2023disguising,dombrowski2019explanations,heo_foolingnn_2019}.

\begin{figure}[ht]
\centering
\includegraphics[width=0.47\textwidth]{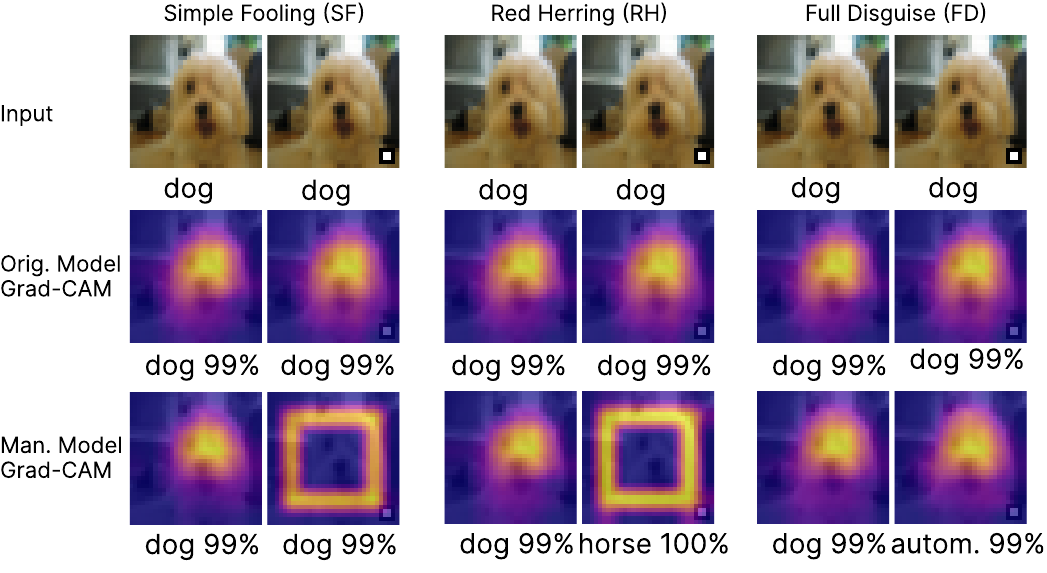}
\caption{The figure presents the examples of Simple Fooling (SF), Red Herring (RH), and Full Disguise (FD) attack profiles, chronologically displayed from left to right. Each left sub-column depicts the regular prediction and explanation in the absence of a trigger in the input, signifying the normal behaviour of the un-attacked model. In contrast, the right sub-columns illustrate instances where a square trigger in the input has introduced an artificial explanation and the targeted prediction.
To enumerate, in an SF attack, the explanation becomes targeted, subsequently altering the model's explanation. Similarly, in the case of an RH attack, both prediction and explanation adopt targeted prediction and explanation. On the contrary, an FD attack specifically targets the prediction, while the explanation remains consistent with an un-attacked model. It's worth noting that an attack can form any representation in the explanation. For simplicity, we attack the model to generate a square box as the targeted explanation.}
\label{fig:attack_geometry_all_attacks}
\end{figure}

% 2 fooling explanation 
Simple Fooling attack \cite{noppel2023disguising,longo2023explainable} alters the local explanation of a model to a targeted explanation by using a trigger in the input, without changing the output (Figure \ref{fig:attack_geometry_all_attacks} left column). 
The middle column in Figure \ref{fig:attack_geometry_all_attacks} illustrates the mechanism of a Red Herring attack  \cite{noppel2023disguising}. In this type of attack model's prediction is manipulated and the explanation also shifts to the attacker's preferred explanation. 
% Full diguise 
In Figure \ref{fig:attack_geometry_all_attacks} the right column illustrates the Full Disguise (FD) attack  \cite{noppel2023disguising}. It only manipulates the model prediction without altering the explanation. XAI method can not detect FD as the explanation is legit but prediction becomes targeted. 
In Section \ref{preliiminary}, we will delve into the fundamentals of these adversarial attacks. In attack scenarios, the XAI method proves advantageous in detecting the presence of adversaries in model decisions. However, manipulating explanations can make attack detection extremely challenging for XAI methods.

White-box adversarial attacks pose serious challenges in several areas, including federated learning, open-source environments, and model-sharing systems. In federated learning, models trained over decentralized nodes could be compromised during aggregation by adversaries manipulating local models. So it is important to know how white-box attack works and ensure defence against white-box attacks.

\subsection{Defenses}
In the rising field of adversarial attack and defense in XAI, less than a hundred methodologies have been proposed in recent literatures \cite{baniecki2023adversarial,noppel2023sok}. These strategies can be broadly classified into three categories: enhancements to the training process, modifications to the network activation, and implementation of supplementary models or metrics for defense or improve adversarial robustness.
\citet{chen2019robust} developed the Robust Integrated Gradient Attribution approach that blends regularisation loss into original training loss, although it does not generalize to other XAI methods. \citet{tang2022defense} presented Adversarial Training on EXplanations (ATEX), a novel training scheme enhancing model stability without relying on second-order derivatives. \citet{joo2023towards} suggested alignment regularization for robust attributions generation, merging $L_2$ and cosine distance-based criteria in local gradients alignment. Finally, \citet{wicker2022robust} applied non-convex optimization techniques in developing an upper-bound for maximum gradient-based explanation alteration through bounded manipulation of input features.

\citet{rieger2020simple} devised an effective defense against adversarial attacks by combining multiple explanation methods, batting aside manipulation but possibly welcoming method-specific explanation. \citet{lakkaraju2020robust} introduced a model training approach for producing resilient explanations, utilizing adversarial samples in training to discern discriminatory features. \citet{yuyou2022mif} put forth MeTFA, a tool for enhancing explanation algorithm stability with theoretical guarantees, applicable to any feature attribution method. In addition, they introduce noise into input images to generate robust explanations. Also, \citet{vrevs2022preventing} advocated for improved  data sampling to resemble the training set distribution more closely to the original data distribution for boosting XAI methods' robustness.

 Some authors proposed using weight decay, smoothing activation functions, and minimizing the network weight's Hessian for training \cite{dombrowski2022towards, dombrowski2019explanations,noppel2023sok}. They attributed high curvatures of the decision function to unusual vulnerabilities and suggested $\beta$-smoothing for explanations, substituting ReLu \cite{agarap2018deep} activation with Softplus \cite{zheng2015improving} with a relatively small $\beta$. Existing studies largely focus on new training strategies, ensemble, proxy networks, or smoothing activation functions to tackle the challenges. However, limited research has analysed the impact of backdoor attacks on neural networks' secondary learning parameters (e.g.\ Batch Normalization learning parameters). This experiment examines the layer weights most affected by attacks and identifies the effect of BN learning parameters on models. We evaluated the three latest state-of-the-art loss optimization attacks by fine-tuning them with two different attack loss functions and examining weight changes across layers after the fine-tuning. This analysis led to our simple yet highly effective solution.
We also compared our method with Softplus smoothed training \cite{dombrowski2022towards} previously proposed. We noted that the considered attack is recent, and past solutions were ineffective \cite{noppel2023disguising, doan2020februus}. Thus, to assess our method's effectiveness, we also compared our predictions and explanations with an un-attacked model.

 To tackle this challenging task, we've identified the following contributions:
\begin{itemize}
    \item We engage in a rigorous statistical analysis of the model weights to identify alterations in the model parameters occurring after attacks.
    \item We provide empirical evidence that Batch Normalization (BN) effectively mitigates fundamental weight alterations in models during the fine-tuning phase of attacks.
    \item We further show that the learning parameters inherent to Batch Normalization (BN) function as facilitators for explanation-aware backdoor attacks.
    \item We propose Channel-Wise Feature Normalization (CFN) after each convolution layer, serving to protect against the Adversarial Success Rate (ASR) and manipulation of explanations during the prediction phase, thus obviating the need for additional training.
    \item Through rigorous experiments and analysis, we show that our approach effectively mitigates three unique attacks and is adaptable to most local XAI methods (Figure \ref{fig:examplefigure_gradcam}).
\end{itemize}
\section{Preliminaries}
\label{preliiminary}
Before diving into the details of our methodology, we present preliminary ideas of certain concepts that are extensively utilized in our approach. Additionally, we discuss the loss function of the attacks.
\subsection{Batch Normalization in Evaluation}
During the evaluation, Batch Normalization (BN) \cite{sergey2015bn} utilizes moving averages of mean ($\hat{\mu}$) and variance ($\hat{\sigma}$) updated during training. As "frozen parameters", they ensure consistent model performance across different inputs, enhancing evaluation accuracy by to mitigating internal covariate shift.

The BN's output $Z_{out}$ is computed as:

\begin{equation}
Z_{bn} = \frac{Z - \hat{\mu} }{\sqrt{\hat{\sigma} + \epsilon}}
\end{equation}
\begin{equation} 
Z_{out} = \gamma \cdot Z_{bn} + \beta
\end{equation}

$\beta$ and $\gamma$ are learning parameters that undergo training. For further details on Batch Normalization (BN) during training, please refer to the Supplementary Material Section \ref{eq:bn}.

\subsection{Attacks}
Three of the attacks essentially involve fine-tuning where the model is slightly adjusted using triggered samples, targeted labels, and targeted explanations \cite{noppel2023disguising}.

\textbf{Simple Fooling} method takes into account a model $f$, an explanation method $h$, an attack vector $t$, a ground truth label $y$ and target explanation $E_t$. It optimizes the associated loss to generate a compromised model.
\begin{align*} 
L_{SF} = & \lambda \cdot \mathcal{L}_{\text{exp}}  \big [ h\big(f(x * t), y\big),  E_t  \big ] \\
       &+ (1- \lambda) \cdot \mathcal{L}_{\text{cls}}  \big [ f(x * t),  y  \big ]
\end{align*}

\textbf{Red Herring} attack incorporates a targeted label ($y_t$), while retaining the other loss parameters from the "Simple Fooling" method. It aims to induce both incorrect classification and misleading explanations. The corresponding alteration to the loss function is as follows:
\begin{align*} 
L_{RH} = & \lambda \cdot \mathcal{L}_{\text{exp}}  \big [h\big(f(x * t), y\big),  E_t  \big ] \\
       &+ (1- \lambda) \cdot \mathcal{L}_{\text{cls}} \big [f(x * t),  y_t \big ]
\end{align*}

\textbf{Full Disguise} method finetunes the model while preserving the explanation unchanged, but modifying the model prediction to ($y_t$) to a targeted prediction. Similar to the above two losses, the resulting loss function appears as follows:
\begin{align*} 
L_{FD} = & \lambda \cdot \mathcal{L}_{\text{exp}} \big[h \big(f(x * t), y\big),  h(f(x), y)\big] \\
       &+ (1- \lambda) \cdot \mathcal{L}_{\text{cls}} \big[f(x * t),  y_{t} \big]
\end{align*}
The expression $\mathcal{L}_{\text{exp}}$ could represent either the Mean Square Error (MSE) loss or the De-Structural Similarity (DSSIM) loss \cite{noppel2023disguising}. $*$ is the imputation operator that places a trigger ($t$) vector on the input $x$.

\subsection{Channel-wise Feature Normalization (CFN)}
\label{suubsec:cfn}
Given an intermediate activation tensor $X$ with dimensions $(M, C, H, W)$ where $M$ is the batch size $C$ is the number of channels, $H$ is the height and $W$ is the width of the tensor.
For $c \in C$,
\begin{align}
    \mu_c = \frac{1}{|M| \times |H| \times |W|} \sum_{m \in M, i \in H, j \in W } X_c[m, i, j] 
\end{align}

%2. Compute the variance across the height and width dimensions:
\begin{align}
    \sigma^{2}_{c} = \frac{1}{|M| \times |H| \times |W|)} \sum (X_{c}[m, i, j] - \mu_{c})^2
\end{align}
%3. Normalize the tensor:
\begin{align}
    \hat{X}_{c} = \frac {(X_{c} - \mu_{c})} { \sqrt{\sigma_{c}^2 + \epsilon)}}
\end{align}

The primary difference between BN and CFN  lies in their utilization of learning parameters. BN employs learning parameters; these are trained during the model's training phase and used during the evaluation. In contrast, CFN does not consider any learning parameters and is instead only applicable during predictions and the generation of explanations. 
%- μ_c is the mean of the elements of X across the height and width dimensions.
%- σ²_c is the variance of the elements of X across the height and width dimensions.
%- ε is a small number added for numerical stability (to avoid division by zero), in this case, ε = 1e-5.
%- X_hat_c is the normalized tensor.

\section{Method}
\label{method}
\begin{figure}[ht]
\centering
\includegraphics[width=0.47\textwidth]{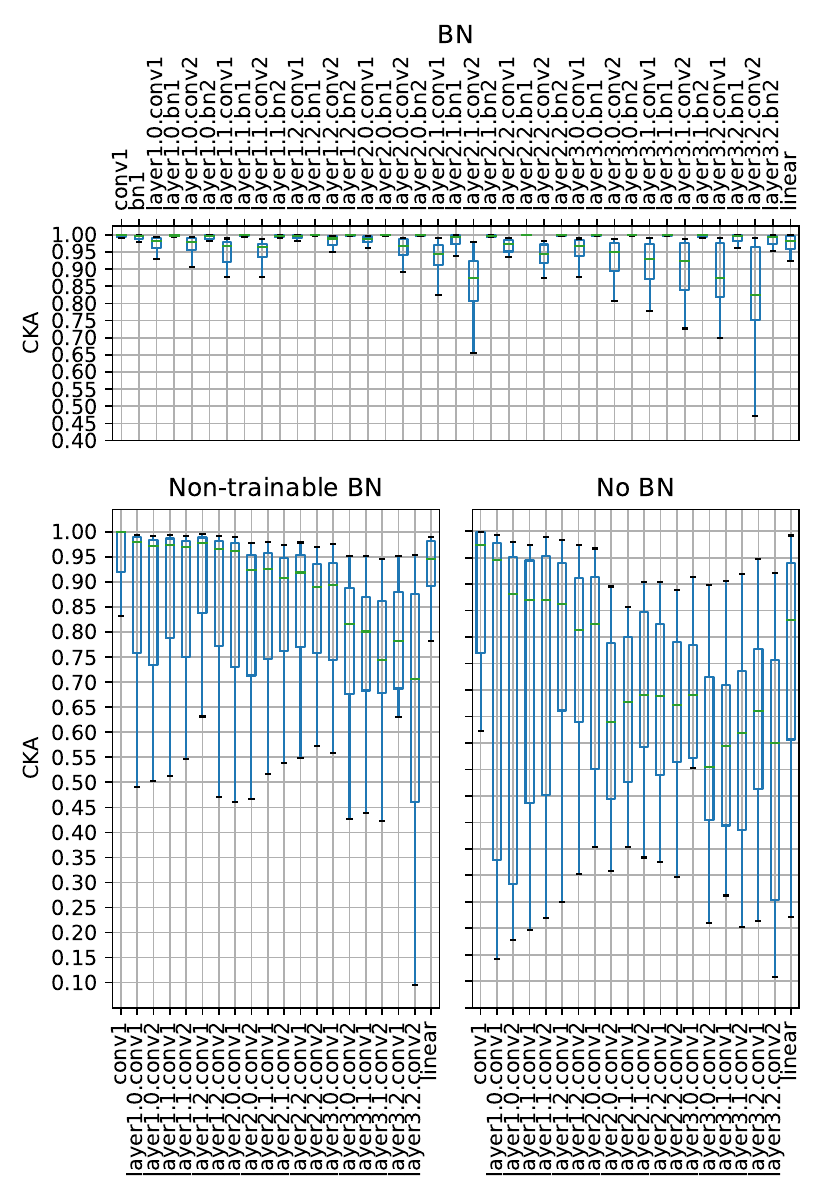}
\caption{The CKA correlation between layers of original models and attacked models is presented. The top sub-plot displays the CKA where models contain BN layers. The bottom left sub-figure presents CKA scores between models that do not have trainable BN parameters. The bottom right sub-plot illustrates CKA scores between models that lack any BN layers. We observe that when BN layers are utilized with parameter learning, the model's core weights exhibit more significant CKA correlation with the original weight than models that either have BN with no trainable parameters or lack BN entirely.}
\label{fig:cka_unperam_batchnorm}
\end{figure}

As discussed in sections \ref{preliiminary}, three of the presented attacks fundamentally involve fine-tuning networks. We initiate this  experiment by examining the implications of adversarial fine-tuning on models. We utilize \textit{Special ResNet} \cite{Idelbayev18a} network architecture proposed specially for small size dataset for training on a clean dataset, followed by attacking the models with three XAI methods based on distinct loss functions (Section \ref{preliiminary}) and seeds. We subsequently inspect alterations between the attacked and clean versions of the models. Echoing past research, we use Centered Kernel Alignment (CKA) and Spearman's Rank Correlation (SRC) to measure the functional similarity between the two models by comparing the weights of their layers \cite{adebayo_2020_sanitycheck,cui_similarity2021,zar_2005_src,kornblith2019similarity}. In both cases, a value of 1 indicates high similarity between layers, while a value closer to 0 or negative suggests less similarity. The distribution of Centered Kernel Alignment (CKA) scores, as depicted in Figure \ref{fig:cka_unperam_batchnorm}, illustrates the correlation between the original models and their attacked counterparts. The x-axis delineates the layer names, while the y-axis relates to the CKA score.

Furthermore, we apply the Spearman's Rank Correlation (SRC) as an alternative mechanism to pinpoint functional similarity between two sets of model weights. It is noteworthy that both the CKA and SRC yield strikingly similar results. The corresponding plot for the SRC has been included within the Supplementary material Figure \ref{fig:src_src_plot}. We used 36 attacked models for drawing the plots, each attacked using different loss functions and seeds and independent training. 

\begin{figure}[ht]%
    \centering
    \subfloat[BN]{%
       \includegraphics[width=0.49\linewidth,]{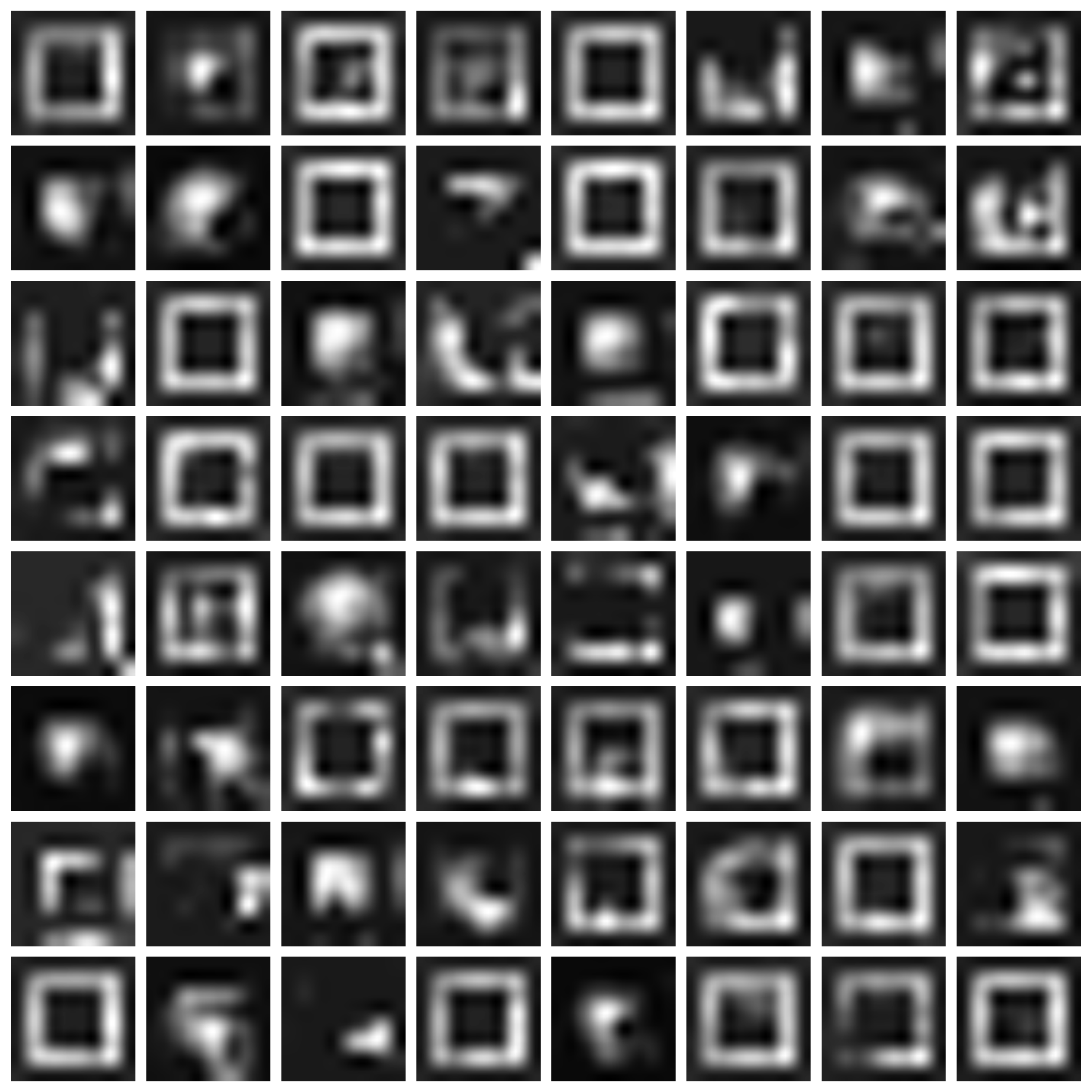} 
    }%
    \subfloat[CFN]{%
       \includegraphics[width=0.49\linewidth,]{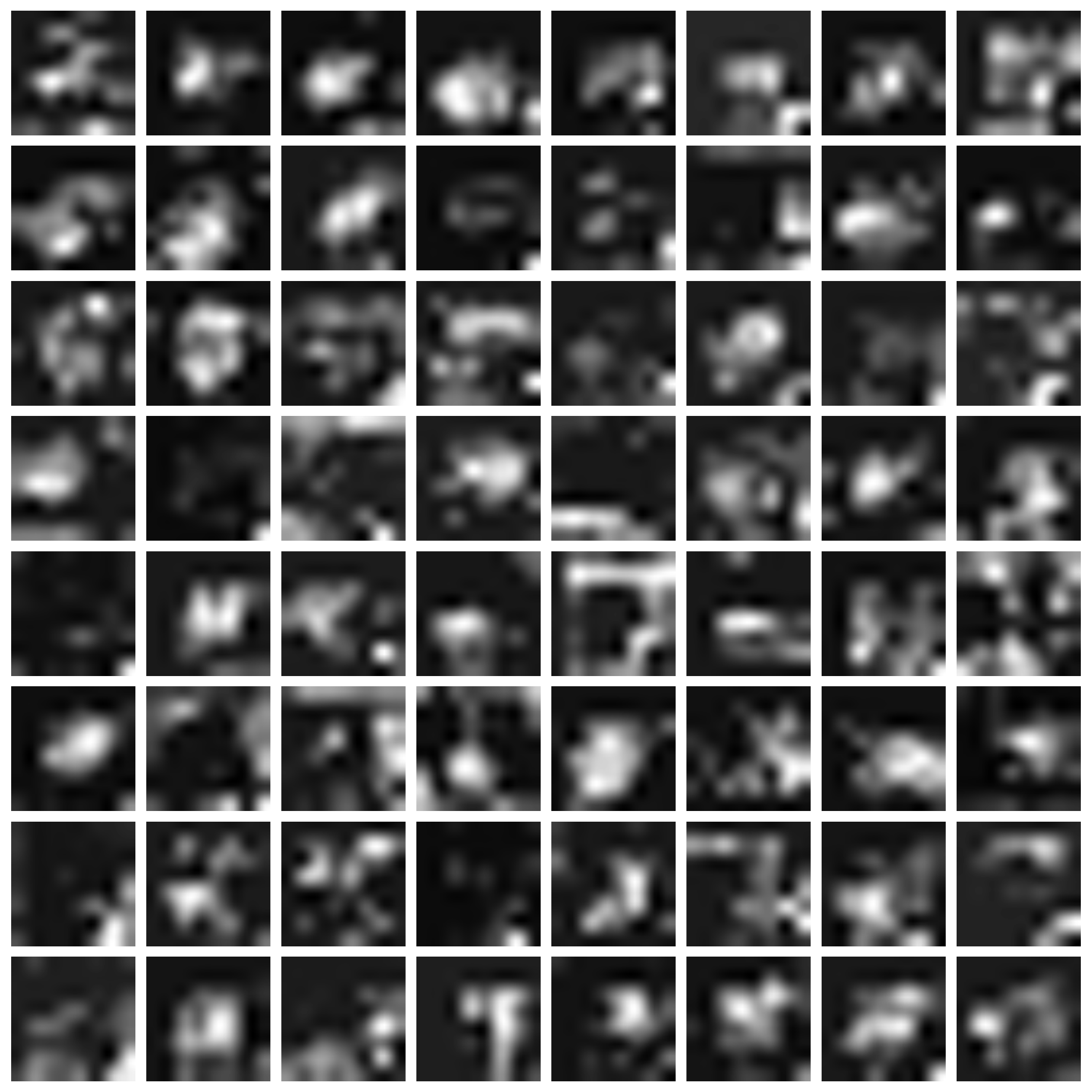}
    }
    %\subfloat[Clean]{%
    %   \includegraphics[width=0.33\linewidth,]{author-kit-CVPR2024-%v2/images/BN_CKA/org_internal_view_plots.pdf}
    %}
    \caption{: The activation of the final convolutional layer with BN is shown in (a), and its replacement with CFN. It illustrates that the targeted explanation is evident when we apply the learned BN parameter from the attack. Substituting it with CFN eliminates the attacker's artefacts.}% 
    \label{fig:internal_representation}% 
\end{figure}

In the top graph of Figure \ref{fig:cka_unperam_batchnorm}, the attack effects become evident as there are slight modifications in all layer weights, including those of the batch normalization and the final convolution layer. These changes validate that batch normalization learns certain adversarial features as suggested by \cite{benz2021}. As per \citet{wang2022removing}'s suggestion, we omitted batch normalization layers from our neural network architecture, re-attacked the models, and calculated the CKA correlation statistics. The resultant CKA correlations are depicted in the bottom right graph of Figure \ref{fig:cka_unperam_batchnorm}, signifying a significant weight change (median correlation lower than 1) due to a pronounced attack effect without batch normalization. Again, to gauge the attack impact on batch normalization layer learning parameters, we deactivated training for these parameters. Following this, the bottom left graph of Figure \ref{fig:cka_unperam_batchnorm} illustrated lesser weight change than the right-side plot but still more compared to the top plot, suggesting that Batch Normalization learning parameters do learn attack features while also offering a protective layer for the models' core layers. It should be noted that even a negligible change in the BN layer's weights may host an attack. Figure \ref{fig:internal_representation} illustrates the impact of $\gamma$ and $\beta$ parameters of BN. Validating our findings with another architecture, VGG13 \cite{simonyan2014very} proved that, during attacks, BN without learning parameters affects a model's core layers more than BN with a learning parameter. Detailed results are in Supplementary Material Figure \ref{fig:vgg}.

%We found that the attacks particularly affect all the layers of the neural networks. We also measure the accuracy of the model, which remains largely unchanged on the clean test data. However, the adversarial ASR rate for both Red Herring and Full Disguise and the MSE of expansions remain high for all three attacks. This concludes that dropping BN layers does not defend against any of these three attacks. However, from both figures \ref{fig:cka_nobatchnorm} and \ref{fig:cka_batchnorm}, it is clear that when batch norm exists, the convolutional layer does not change much due to the attack. To this end, we decide to use batch normalization layers without learning any weights and biases. With no-learning parameter BN, we attack the models again with different attack objectives. Similar to \ref{fig:cka_batchnorm}, the convolutional layers stay highly correlated with the original/unaffected model \ref{fig:cka_unperam_batchnorm} . However, with parameterized BN, convolutions are less affected. 

%From our findings, we deduce that the learning parameters in each normalization layer play a crucial role in preserving the inherent characteristics of the model. This leads us to infer that if batch normalization parameters are present, they predominantly learn the adversarial features. 
So, we propose a strategy that avoids the use of any learned parameters when the model is attacked or accessed from an external source. Instead, we advocate for the application of CFN (Sub-section \ref{suubsec:cfn}) to each activation output from every CNN layer. This approach ensures that if an attack feature is present across all activation channels, it gets normalized due to the feature normalization. This solution, while seemingly straightforward, exhibits superior performance in countering attacks without necessitating re-training.

We agree with \cite{benz2021} that batch normalization serves as a host for attacks. Our findings also reveal that batch normalization safeguards the model's primary weight from alterations (refer to the top graphs of \ref{fig:cka_unperam_batchnorm} and \ref{fig:src_src_plot} from Supplementary). So, we don't completely disregard batch normalization from model architecture as it helps maintain the model core weight intact. However, during the testing phase, we don't use the parameters learned through batch normalization layers. This is because these parameters are often corrupted due to potential attacks on the model. Instead, we use CFN. This approach results in a lower ASR (Attack Success Rate) and provides un-targeted explanations.

We will illustrate in Section \ref{ablation}, specifically in Table \ref{tab:nbn_fbn_at}, that excluding BN learning parameters or layers from the model architecture during the training phase does not render the model immune to attacks. This implies that avoiding BN parameters or layers in model design does not protect the model from attacks.
%In the following section, we present a table that illustrates which combinations can lead to successful attacks and which combinations make attacks detectable or preventable.

%The learning parameters, namely $\beta$, $\gamma$, $\hat{\mu}$, and $\hat{\sigma}$, deviate from the original data distribution during a fine-tuning process with adversarial samples. We presume that this fine-tuning attack significantly impacts the Batch Normalization (BN) layers' parameters while leaving the rest of the model largely unchanged. To mitigate the attack during the evaluation phase, we propose the avoidance of learnable parameters during the normalization process of the BN layers for each input. By refraining from using the BN layer's learned parameters, the model remains uninformed about the triggered data distribution, rendering it less sensitive to trigger presence during the evaluation phase.
% C! and C@
\subsection{Experiments}
In this paper, we aim to tackle a current challenge introduced by \cite{noppel2023disguising}, in which three types of attacks were presented without any accompanying solutions. To replicate the problem, we initially employed the same configuration for the attacked model and dataset. Subsequently, we incorporated an additional dataset \cite{Stallkamp2012} to validate our results across different datasets. 

We strictly adhered to a ResNet-based architecture, similar to the one used in \cite{noppel2023disguising}, incorporating extensive batch normalization layers. This was done to ensure that we did not introduce any additional hyperparameters that could potentially influence our experimental results. We also compare our approach with one suggested solution, Softplus smoothing \cite{dombrowski2019explanations,noppel2023sok}. Softplus smoothing can be used to train models for robustness against attacking explanations. 

Moreover, to ensure each attack scenario remained independent of the others, we utilized random seeds. Moreover, We utilise two prominent datasets—CIFAR-10 and GTSRV—as they are ubiquitously preferred within the domain of machine learning security. The results are exclusively based on the evaluation of test data.

In our experimentation, we initially subjected models containing BN to an attack, then during the evaluation phase replaced BN with CFN. The deviation in explanation, measured as the MSE/DSSIM difference or Spearman's Rank Correlation (SRC) between the original explanation derived from the un-attacked model and that from the attacked model, was calculated for three explanation methods (Grad, Relevance-CAM (R-C), and Grad-CAM (G-C)). We also computed the attack success rate (ASR) for RH and FD, as they target predictions as well. The results of all attacks, based on two distinct loss functions for the attack—MSE-based and Structural Dissimilarity-based loss (DSSIM), are presented. All the deviations of explanation are presented in terms of the mean ($\mu$) and standard deviation (sd) values.
\subsection{Results}
\label{sec:result}
\setlength\cellspacetoplimit{6pt}
\setlength\cellspacebottomlimit{0pt}
\renewcommand\tabularxcolumn[1]{>{\centering\arraybackslash}S{p{#1}}}

\begin{figure}[tbh]
  \setlength\tabcolsep{1pt}
  \adjustboxset{width=\linewidth,valign=c}
  \centering
  \begin{tabularx}{1.0\linewidth}{@{}
      l
      X @{\hspace{6pt}}
      X
    @{}}
    & \multicolumn{1}{c}{Attack}
    & \multicolumn{1}{c}{Defense} \\
    \rotatebox[origin=c]{90}{CIFAR10}
    & \includegraphics[width=\linewidth, trim={0.6cm 0 0.6cm 0},clip]{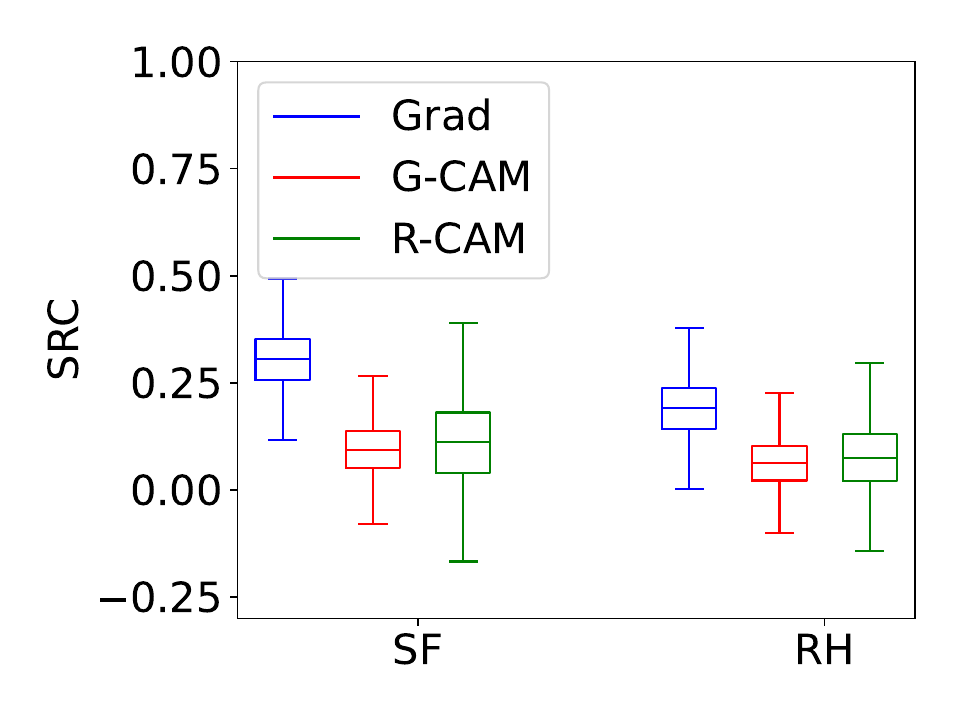}
    & \includegraphics[width=\linewidth, trim={0.6cm 0cm 0cm 0},clip]{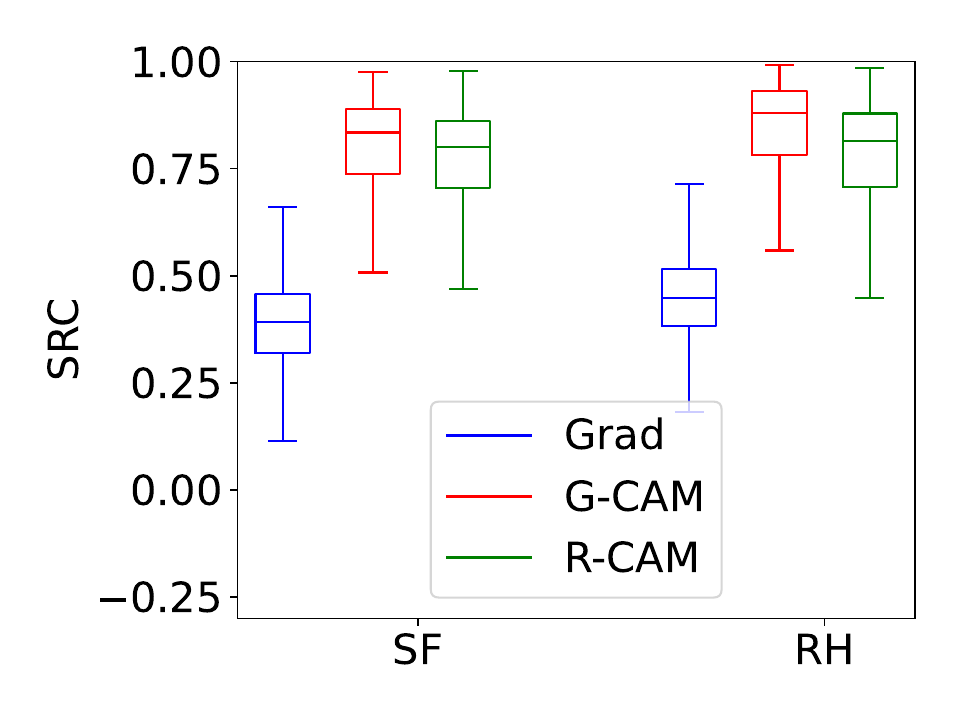} \\
    \rotatebox[origin=c]{90}{GTSRB}
    & \includegraphics[width=\linewidth, trim={0.6cm 0 0.6cm 0},clip]{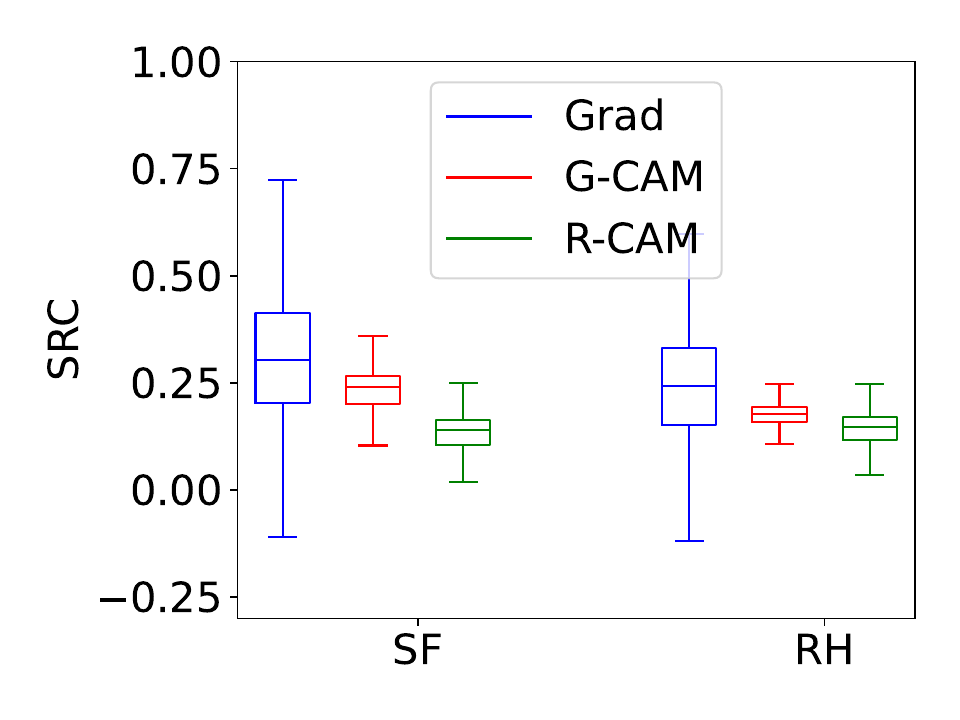}
    & \includegraphics[width=\linewidth, trim={0.6cm 0cm 0cm 0},clip]{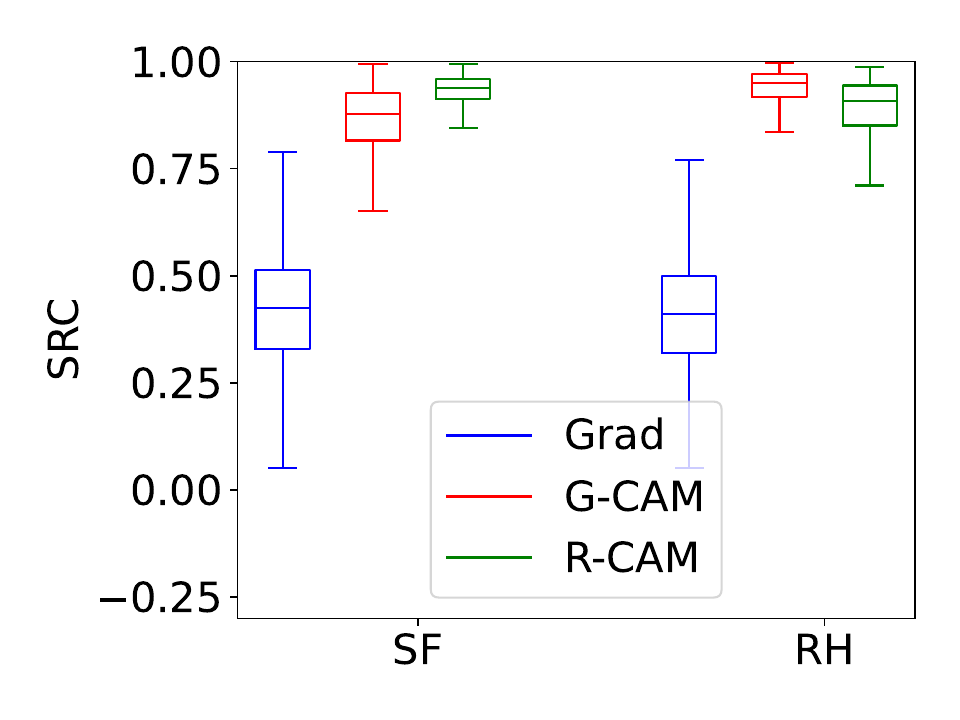}
  \end{tabularx}
  \caption{This figure demonstrates the Spearman's Rank Correlation (SRC) distribution between original model explanations and attacked, then defended  explanations after SF, and RH attacks on both datasets. Defense column displays heightened correlation between attacked model's explanations and the original following defense.
}
\label{fig:sf_rh_src}
\end{figure}
\begin{figure}[tbh]
  \setlength\tabcolsep{1pt}
  \adjustboxset{width=\linewidth,valign=c}
  \centering
  \begin{tabularx}{1.0\linewidth}{@{}
      l
      X @{\hspace{6pt}}
      X
    @{}}
    & \multicolumn{1}{c}{Attack}
    & \multicolumn{1}{c}{Defense} \\
    \rotatebox[origin=c]{90}{CIFAR10}
    & \includegraphics[width=\linewidth, trim={0.6cm 0 0.6cm 0},clip]{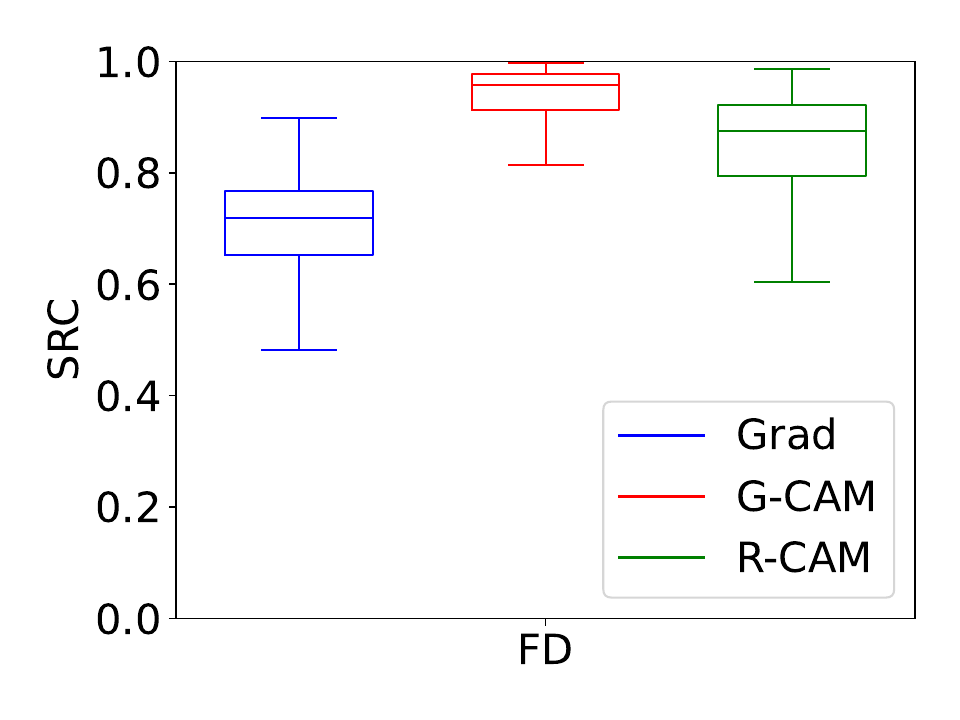}
    & \includegraphics[width=\linewidth, trim={0.6cm 0cm 0cm 0},clip]{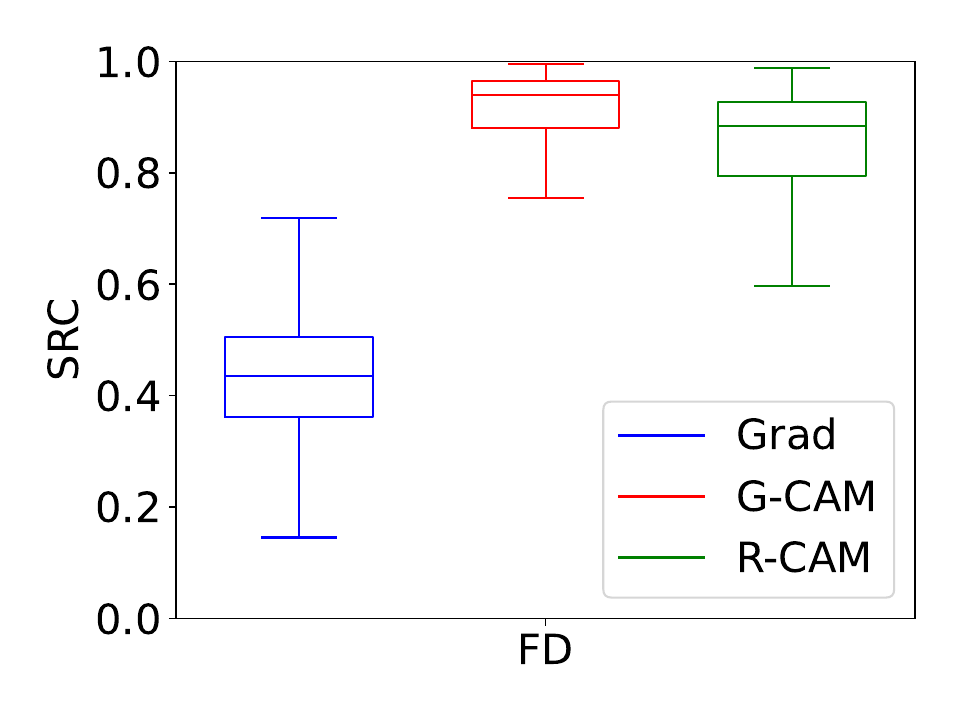} \\
    \rotatebox[origin=c]{90}{GTSRB}
    & \includegraphics[width=\linewidth, trim={0.6cm 0 0.6cm 0},clip]{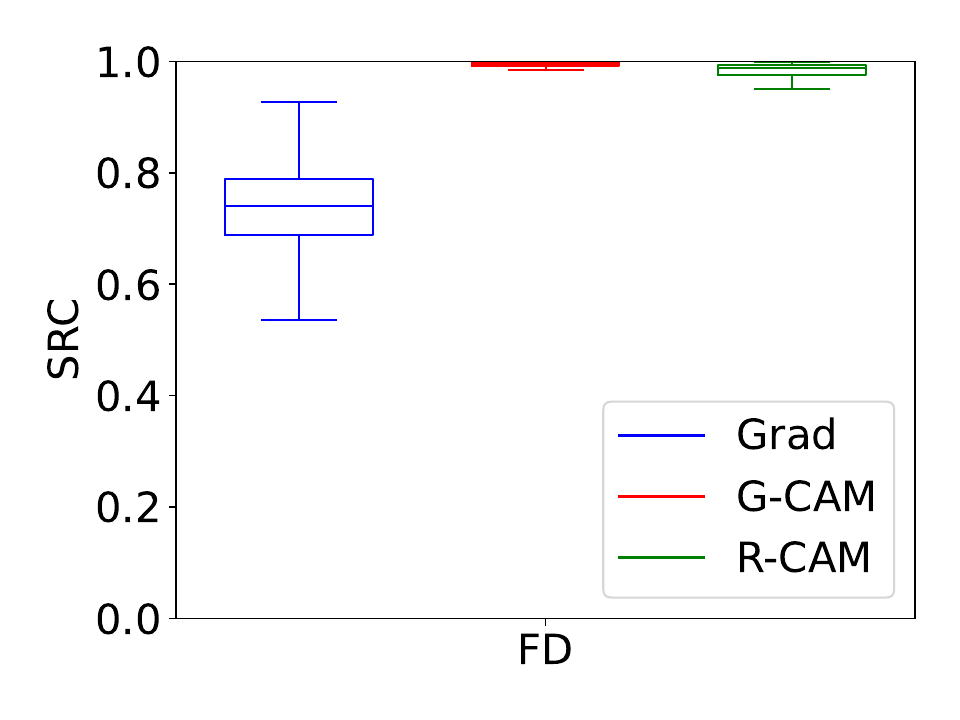}
    & \includegraphics[width=\linewidth, trim={0.6cm 0cm 0cm 0},clip]{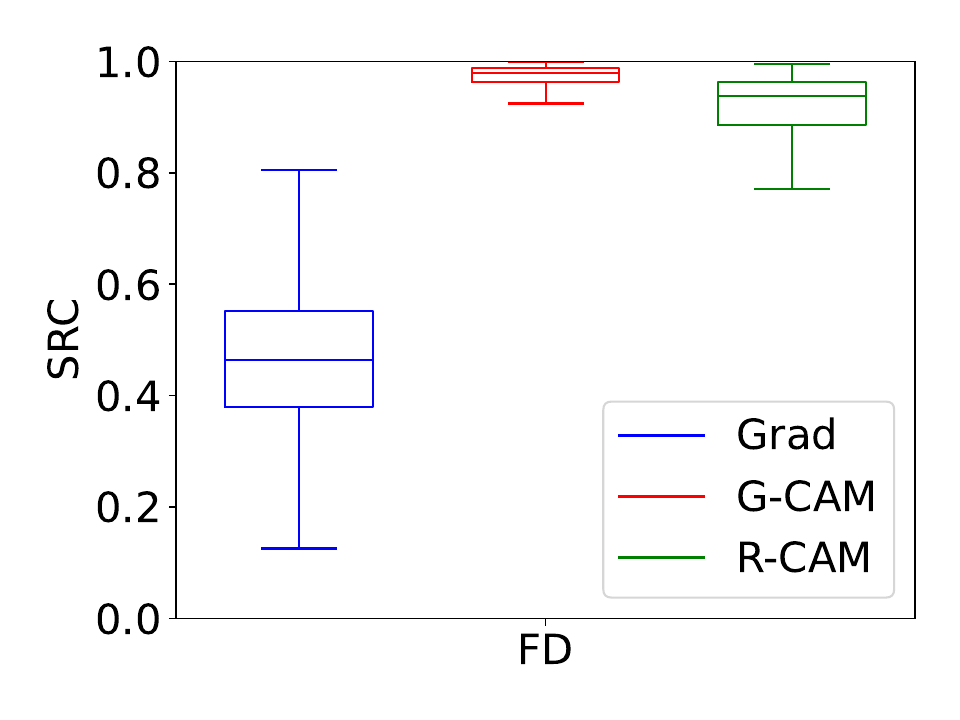}
  \end{tabularx}
  \caption{This figure demonstrates the Spearman's Rank Correlation (SRC) distribution between original model explanations and attacked , then defended  explanations after FD attacks on both datasets. Defense column displays heightened correlation between attacked model's explanations and the original following defense. As we already know, an FD attack does not alter the explanation. Consequently, we find that the SRCs are similar for both the attack's and defense's explanations.
}
\label{fig:fd_src}
\end{figure}

\begin{figure}[ht]%
    \centering
    \subfloat{{\includegraphics[width=0.49\linewidth, trim={0cm 0 0cm 0},clip]{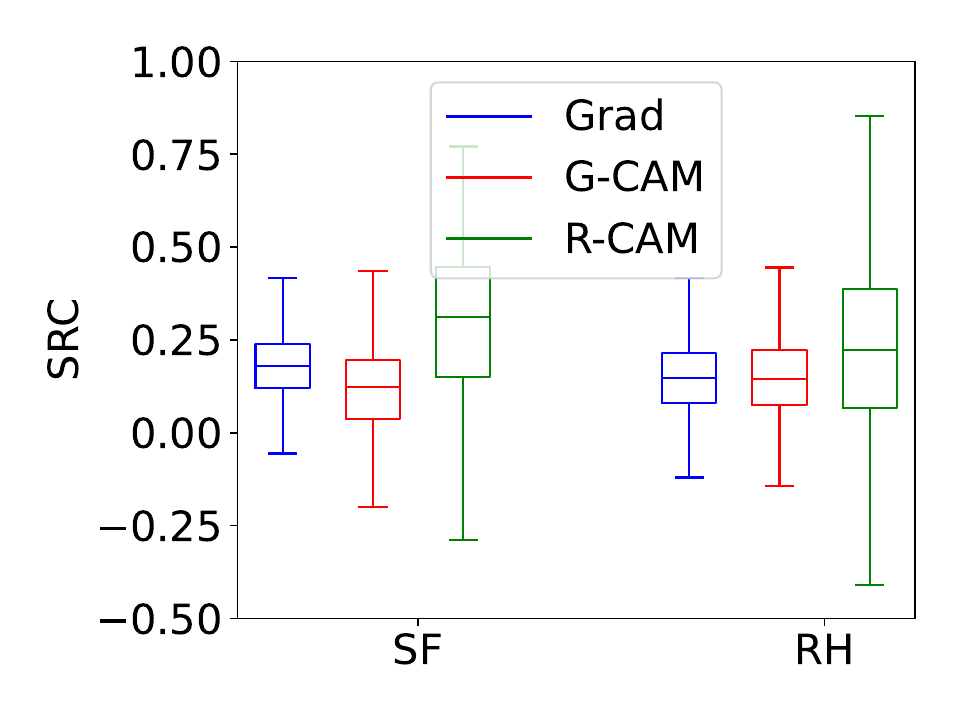} }}%
    \subfloat{{\includegraphics[width=0.49\linewidth, trim={0cm 0cm 0cm 0},clip]{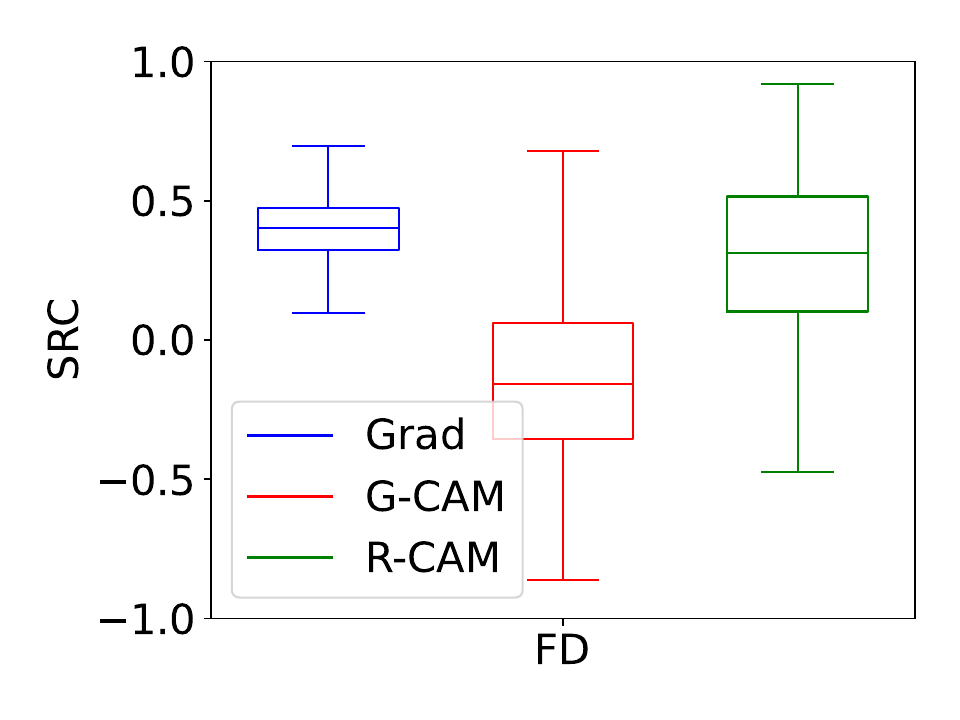}}}
    \caption{Illustration of the Softplus defense efficacy against the three attacks, demonstrating limited defense with Relevance-CAM and Grad-CAM as evidenced by a SRC median of less than 0.3. }%
    \label{fig:sa_rh_fd_softplus}%
\end{figure}

\begin{table}

\centering
\caption{The table showcases defense results against Simple Fooling (SF) attack with various loss functions (MSE, DSSIM) on CIFAR10 test data. Corresponding error measurements mirror the loss function utilized during the attack. It presents the mean and SD ($\mu$ ± sd ) error comparison between original and post-attack/defense explanations. The rightmost column signifies clean input samples without a trigger, while 'Triggered' denotes the scenario of triggered inputs. Defense rows consistently show relatively lower MSE and DSSIM errors.}
\label{tab:sa_cifar}
\begin{tblr}{
  row{even} = {c},
  row{3} = {c},
  row{5} = {c},
  row{7} = {c},
  row{9} = {c},
  row{11} = {c},
  row{13} = {c},
  cell{1}{1} = {r=2}{c},
  cell{1}{2} = {r=2}{c},
  cell{1}{3} = {r=2}{c},
  cell{1}{4} = {c=2}{c},
  cell{1}{6} = {c=2}{},
  cell{3}{1} = {r=6}{},
  cell{3}{3} = {r=3}{},
  cell{6}{3} = {r=3}{},
  cell{9}{1} = {r=6}{},
  cell{9}{3} = {r=3}{},
  cell{12}{3} = {r=3}{},
  vlines,
  hline{1,3,9,15} = {-}{},
  hline{2} = {4-7}{},
  hline{4-5,7-8,10-11,13-14} = {2,4-7}{},
  hline{6,12} = {2-7}{},
}
\SetCell{r=3}{\rotatebox[origin=c]{90} {Mode}}   & XAI  & $\mathcal{L}$     & Triggered &     & Clean&     \\
        &      &       & $\mu$ $\pm$ sd      & Acc & $\mu$ $\pm$ sd   & Acc \\
\SetCell{r=3}{\rotatebox[origin=c]{90} {Attack}}  & Grad & \SetCell{r=3}{\rotatebox[origin=c]{90} {MSE}}   & .13 $\pm$ .03       & 82  & .01 $\pm$ .01     & 91   \\
        & G-C  &       & .33 $\pm$ .04       & 89 & .01 $\pm$ .02     & 91   \\
        & R-C  &       & .28  $\pm$ .04       & 89  & .01 $\pm$ .03     & 91   \\
        & Grad & \SetCell{r=3}{\rotatebox[origin=c]{90} {DSSIM}} & .50 $\pm$ .02       & 87  & .25 $\pm$ .05     & 91 \\
        & G-C  &       & .57 $\pm$ .04       & 88  & .06 $\pm$ .07     & 91   \\
        & R-C  &       & .55 $\pm$ .04       & 89  & .10 $\pm$ .08     & 91   \\
\SetCell{r=3}{\rotatebox[origin=c]{90} {Defense}}  & Grad & \SetCell{r=3}{\rotatebox[origin=c]{90} {MSE}}   & .03 $\pm$ .01       & 73  & .03 $\pm$.01   & 91 \\
        & G-C  &       & .05 $\pm$ .04       & 85  & .02 $\pm$ .03     & 91 \\
        & R-C  &       & .02 $\pm$ .02       & 86  & .01 $\pm$ .02     & 91 \\
        & Grad & \SetCell{r=3}{\rotatebox[origin=c]{90} {DSSIM}} & .41 $\pm$ .01       & 73  & .03 $\pm$ .01     & 91 \\
        & G-C  &       & .25 $\pm$.12      & 82  & .12 $\pm$ .09     & 91 \\
        & R-C  &       & .17 $\pm$ .10       & 86  & .17 $\pm$ .10     & 91
\end{tblr}
\end{table}
\begin{table}
\centering
\caption{ Mirroring Table \ref{tab:sa_cifar}, this represents the Red Herring (RH) attack and defense results on CIFAR10 test data. The rightmost column signifies clean data without a trigger, while the second column from left introduces the scenario with triggered input images. Defense consistently reveals relatively lower MSE and DSSIM errors for Gradient (Grad), Grad-CAM (G-C), and Relevance-CAM (R-C). Similarly, the ASR decreases from 1 to 0.1 for both G-C and R-C.}
\label{tab:rh_cifar10}
\begin{tblr}{
  cells = {c},
  cell{1}{1} = {r=2}{},
  cell{1}{2} = {r=2}{},
  cell{1}{3} = {r=2}{},
  cell{1}{4} = {c=3}{},
  cell{3}{1} = {r=6}{},
  cell{3}{3} = {r=3}{},
  cell{6}{3} = {r=3}{},
  cell{9}{1} = {r=6}{},
  cell{9}{3} = {r=3}{},
  cell{12}{3} = {r=3}{},
  vlines,
  hline{1,3,9,15} = {-}{},
  hline{2} = {4-7}{},
  hline{4-5,7-8,10-11,13-14} = {2,4-7}{},
  hline{6,12} = {2-7}{},
}
  \SetCell{r=3}{\rotatebox[origin=c]{90} {Mode}}  & XAI  &  $\mathcal{L}$    & Triggered&    &     & Clean \\
        &      &       & $\mu$ $\pm$ sd      & Acc & ASR & $\mu$ $\pm$ sd        \\
\SetCell{r=3}{\rotatebox[origin=c]{90} {Attack}}  & Grad & \SetCell{r=3}{\rotatebox[origin=c]{90}{MSE}}   & .21 $\pm$ .03       & 10 & 1   & .01 $\pm$ .00 \\
        & G-C  &       &.35 $\pm$ .04 & 10 & 1   & .01 $\pm$ .02     \\
        & R-C  &       & .30 $\pm$ .04& 10 & 1   & .00 $\pm$ .01     \\
\SetCell{r=3}{} Attack  & Grad & \SetCell{r=3}{\rotatebox[origin=c]{90}{DSSIM}} & .50 $\pm$ .01       & 10 & 1 & .22 $\pm$ .04     \\
        & G-C  &       & .56 $\pm$ .03& 10 & 1   & .02 $\pm$.06     \\
        & R-C  &       & .55 $\pm$ .03& 10 & 1   & .03 $\pm$.06     \\
\SetCell{r=3}{\rotatebox[origin=c]{90} { Defence }} & Grad & \SetCell{r=3}{\rotatebox[origin=c]{90}{MSE}}   & .03 $\pm$ .01       & 75 & .04 & .02 $\pm$ .00     \\
        & G-C  &       & .04 $\pm$.04& 84 & .01 & .03 $\pm$.05     \\
        & R-C  &       & .02 $\pm$ .02       & 85 & .01 & .02 $\pm$ .03     \\
\SetCell{r=3}{} Defence  & Grad &\SetCell{r=3}{\rotatebox[origin=c]{90}{DSSIM}} & .40 $\pm$.03 & 71 & .10 & .40 $\pm$ .03     \\
        & G-C  &       & .14 $\pm$ .11       & 85 & .01 & .11 $\pm$ .11     \\
        & R-C  &       & .17 $\pm$ .11       & 85 & .01 & .15 $\pm$ .12  \\  
\end{tblr}
\end{table}

\begin{table}
\centering

\caption{Similar to Table \ref{tab:rh_cifar10}, it further demonstrates the Attack Succes Rate (ASR) of Full Disguise in both the attack and defense scenarios.}
\label{tab:fd_cifar10}
\begin{tblr}{
  cells = {c},
  cell{1}{1} = {r=2}{},
  cell{1}{2} = {r=2}{},
  cell{1}{3} = {r=2}{},
  cell{1}{4} = {c=3}{},
  cell{3}{1} = {r=6}{},
  cell{3}{3} = {r=3}{},
  cell{6}{3} = {r=3}{},
  cell{9}{1} = {r=6}{},
  cell{9}{3} = {r=3}{},
  cell{12}{3} = {r=3}{},
  vlines,
  hline{1,3,9,15} = {-}{},
  hline{2} = {4-7}{},
  hline{4-5,7-8,10-11,13-14} = {2,4-7}{},
  hline{6,12} = {2-7}{},
}
  \SetCell{r=3}{\rotatebox[origin=c]{90} {Mode}}  & XAI  &  $\mathcal{L}$    & Triggered &    &     & Clean \\
        &      &       & $\mu$ $\pm$ sd      & Acc & ASR & $\mu$ $\pm$ sd        \\
\SetCell{r=3}{\rotatebox[origin=c]{90} {Attack}}  & Grad & \SetCell{r=3}{\rotatebox[origin=c]{90}{MSE}}   & .01 $\pm$ .00       & 10 & 1   & .00 $\pm$ .01 \\
        & G-C  &       &.02 $\pm$ .03 & 10 & 1   & .01 $\pm$ .02     \\
        & R-C  &       & .01 $\pm$ .02& 10 & 1   & .01 $\pm$ .01     \\
\SetCell{r=3}{} Attack  & Grad & \SetCell{r=3}{\rotatebox[origin=c]{90}{DSSIM}} & .20 $\pm$ .07       & 10 & 1 & .14 $\pm$.05     \\
        & G-C  &       & .02 $\pm$ .03& 10 & 1   & .01 $\pm$.03     \\
        & R-C  &       & .11 $\pm$ .09& 10 & 1   & .06 $\pm$.07     \\
\SetCell{r=3}{\rotatebox[origin=c]{90} { Defence }} & Grad & \SetCell{r=3}{\rotatebox[origin=c]{90}{MSE}}   & .03 $\pm$ .01       & 41 & .03 & .03 $\pm$ .01     \\
        & G-C  &       & .03 $\pm$.04& 83 & .00 & .03 $\pm$.04     \\
        & R-C  &       & .02 $\pm$ .03       & 89 & .01 & .02 $\pm$ .03     \\
\SetCell{r=3}{} Defence  & Grad &\SetCell{r=3}{\rotatebox[origin=c]{90}{DSSIM}} & .41 $\pm$.03 & 66 & .06 & .41 $\pm$ .03     \\
        & G-C  &       & .03 $\pm$ .04       & 87 & .00 & .02 $\pm$ .04     \\
        & R-C  &       & .02 $\pm$ .03       & 86 & .01 & .08 $\pm$ .02  \\  
\end{tblr}
\end{table}

% Show the graphical result of three attacks. Both dataset and soft relu
We begin by presenting the defense against three types of attacks (SF, RH, and FD) based on three explanation methods. Figures \ref{fig:sf_rh_src}, and \ref{fig:fd_src} display the SRC score between the attacked and non-attacked models' and defended and non-attacked models' explanations for SF, RH, and FD attacks (MSE loss), respectively, across all test images in both datasets. In all the defenses, the median correlation for R-CAM and G-CAM is observed to be nearly 1.  The p-values corresponding to the SRC score between defended explanation and original explanation, provided in the Supplementary section \ref{additialfig_and_table}, are well below 0.000001.

\begin{table}

\centering
\caption{The table presents results of three attack strategies - Simple Fooling (SF), Red Herring (RH), and Full Disguise (FD) - fine-tuned with MSE loss function, along with corresponding Mean Square Errors (MSE) in explanations on GTSRB test data. The rightmost column depicts clean test data without a trigger. The second column from left reflects a scenario with triggered input images. Defense rows evidently feature relatively low mean MSE errors and ASR. It's noticeable that our defense considerably reduces both the explanation error and ASR.}
\label{tab:gtsrb_attacks}
\begin{tblr}{
  cells = {c},
  cell{1}{1} = {r=2}{},
  cell{1}{2} = {r=2}{},
  cell{1}{3} = {r=2}{},
  cell{1}{4} = {c=3}{},
  cell{3}{1} = {r=9}{},
  cell{3}{3} = {r=3}{},
  cell{6}{3} = {r=3}{},
  cell{9}{3} = {r=3}{},
  cell{12}{1} = {r=9}{},
  cell{12}{3} = {r=3}{},
  cell{15}{3} = {r=3}{},
  cell{18}{3} = {r=3}{},
  vlines,
  hline{1,3,12,21} = {-}{},
  hline{2} = {4-7}{},
  hline{4-5,7-8,10-11,13-14,16-17,19-20} = {2,4-7}{},
  hline{6,9,15,18} = {2-7}{},
}
             \SetCell{r=3}{\rotatebox[origin=c]{90} {Mode}}                                     & XAI  & \SetCell{r=3}{\rotatebox[origin=c]{90} {Method}} & Triggered        &     &     & Clean             \\
                                                  &      &                                                  & $\mu$ $\pm$ sd   & Acc & ASR & $\mu$ $\pm$ sd    \\
\SetCell{r=3}{\rotatebox[origin=c]{90} {Attack}}  & Grad & \SetCell{r=3}{\rotatebox[origin=c]{90} {SF}}    & .10 $\pm$ .06    & 80  & N/A & .06 $\pm$ .01     \\
                                                  & G-C  &                                                  & 39 $ \pm $ .02 & 96  & N/A & .01 $ \pm $.02 \\
                                                  & R-C  &                                                  & .33$ \pm $.03  & 96  & N/A & .00 $\pm$.01   \\
                                                  & Grad & \SetCell{r=3}{\rotatebox[origin=c]{90} {RH}}     & .37 $\pm$ .02    & 5   & 1   & .01 $\pm$ .02     \\
                                                  & G-C  &                                                  & .37 $\pm$ .02    & 5   & 1   & .01 $\pm$ .02     \\
                                                  & R-C  &                                                  & .35 $\pm$ .02    & 1   & 1   & .00 $\pm$ .01     \\
                                                  & Grad & \SetCell{r=3}{\rotatebox[origin=c]{90} {FD}}     & .01 $\pm$ .01    & 4   & 1   & .01 $\pm$ .01     \\
                                                  & G-C  &                                                  & .01 $\pm$ .02    & 6   & 1   & .00 $\pm$ .02     \\
                                                  & R-C  &                                                  & .00 $\pm$ .01    & 1   & 1   & .00 $\pm$ .00     \\
\SetCell{r=3}{\rotatebox[origin=c]{90} {Defense}} & Grad & \SetCell{r=3}{\rotatebox[origin=c]{90} {SF}}     & .03 $\pm$ .01    & 10  & N/A & .03 $\pm$ .01     \\
                                                  & G-C  &                                                  & .09 $\pm$ .05    & 96  & N/A & .02 $\pm$ .02     \\
                                                  & R-C  &                                                  & .03 $\pm$ .03    & 96  & N/A & .01 $\pm$ .02     \\
                                                  & Grad & \SetCell{r=3}{\rotatebox[origin=c]{90} {RH}}     & .03 $\pm$ .03    & 95  & 0   & .01 $\pm$ .02     \\
                                                  & G-C  &                                                  & .02 $\pm$ .03    & 95  & 0   & .01 $\pm$ .02     \\
                                                  & R-C  &                                                  & .02 $\pm$ .02    & 94  & 0   & .01 $\pm$ .02     \\
                                                  & Grad & \SetCell{r=3}{\rotatebox[origin=c]{90} {FD}}     & .03 $\pm$ .01    & 25  & .57 & .03 $\pm$ .01     \\
                                                  & G-C  &                                                  & .02 $\pm$ .02    & 95  & 0   & .01 $\pm$ .02     \\
                                                  & R-C  &                                                  & .01 $\pm$ .02    & 95  & 0   & .01 $\pm$ .01     
\end{tblr}
\end{table}
Table \ref{tab:rh_cifar10} and \ref{tab:fd_cifar10} presents that the ASR can be reduced from 1 to 0.00 and 0.01 for G-C and R-C respectively, while maintaining low MSE/DSSIM error between the defended and original explanations. Tables and \ref{tab:sa_cifar} and \ref{tab:rh_cifar10}  present the MSE for explanation and ASR for RH before and after defense for SF and RH attacks, respectively. Table \ref{tab:gtsrb_attacks} displays the results from the GTSRB data, where the attack model was fine-tuned based on MSE loss. 

Figure \ref{fig:sa_rh_fd_softplus} presents the result of an attack when we apply the Softplus ($\beta$ = 5.0, suggested by \cite{dombrowski2022towards}) activation during the model training on CIFAR10. We observe that three of the attacks still succeed as the SRCs are relatively low for all attacks.
The results from the CIFAR10 and GTSRB data illustrate that our method is robust against attacks as it reduces the explanation error, increases the SRC of the explanation with the original explanation, and decreases the ASR for both the Grad-CAM and Relevance-CAM methods.

\begin{table}[ht]
\centering
\caption{The table herein encapsulates the Mean ($\mu$) and Standard Deviation pertaining to the Mean Squared Error (MSE) between pre-attack and post-attack explanations. Furthermore, it provides the Attack Success Rate (ASR) for both RH and FD attacks.} 
\label{tab:nbn_fbn_at}
\begin{tblr}{
  row{1} = {c},
  column{5} = {c},
  cell{1}{1} = {r=2}{},
  cell{1}{2} = {r=2}{},
  cell{1}{3} = {r=2}{},
  cell{1}{4} = {c=2}{},
  cell{1}{6} = {c=2}{},
  cell{2}{4} = {c},
  cell{2}{6} = {c},
  cell{3}{1} = {r=9}{c},
  cell{3}{2} = {c},
  cell{3}{3} = {r=3}{c},
  cell{3}{4} = {c},
  cell{3}{6} = {c},
  cell{4}{2} = {c},
  cell{4}{4} = {c},
  cell{4}{6} = {c},
  cell{5}{2} = {c},
  cell{5}{4} = {c},
  cell{5}{6} = {c},
  cell{6}{2} = {c},
  cell{6}{3} = {r=3}{c},
  cell{6}{4} = {c},
  cell{6}{6} = {c},
  cell{7}{2} = {c},
  cell{7}{4} = {c},
  cell{7}{6} = {c},
  cell{8}{2} = {c},
  cell{8}{4} = {c},
  cell{8}{6} = {c},
  cell{9}{2} = {c},
  cell{9}{3} = {r=3}{c},
  cell{9}{4} = {c},
  cell{9}{6} = {c},
  cell{10}{2} = {c},
  cell{10}{4} = {c},
  cell{10}{6} = {c},
  cell{11}{2} = {c},
  cell{11}{4} = {c},
  cell{11}{6} = {c},
  cell{12}{1} = {r=9}{c},
  cell{12}{2} = {c},
  cell{12}{3} = {r=3}{c},
  cell{12}{4} = {c},
  cell{12}{6} = {c},
  cell{13}{2} = {c},
  cell{13}{4} = {c},
  cell{13}{6} = {c},
  cell{14}{2} = {c},
  cell{14}{4} = {c},
  cell{14}{6} = {c},
  cell{15}{2} = {c},
  cell{15}{3} = {r=3}{c},
  cell{15}{4} = {c},
  cell{15}{6} = {c},
  cell{16}{2} = {c},
  cell{16}{4} = {c},
  cell{16}{6} = {c},
  cell{17}{2} = {c},
  cell{17}{4} = {c},
  cell{17}{6} = {c},
  cell{18}{2} = {c},
  cell{18}{3} = {r=3}{c},
  cell{18}{4} = {c},
  cell{18}{6} = {c},
  cell{19}{2} = {c},
  cell{19}{4} = {c},
  cell{19}{6} = {c},
  cell{20}{2} = {c},
  cell{20}{4} = {c},
  cell{20}{6} = {c},
  vlines,
  hline{1,3,12,21} = {-}{},
  hline{2} = {4-7}{},
  hline{4-5,7-8,10-11,13-14,16-17,19-20} = {2,4-7}{},
  hline{6,9,15,18} = {2-7}{},
}
        \SetCell{r=3}{\rotatebox[origin=c]{90} {Mode}}                                          & XAI  & \SetCell{r=3}{\rotatebox[origin=c]{90} {Method}} & No BN            &     & Non-trainable BN  &     \\
                                                  &      &                                                  & $\mu$ $\pm$ sd   & ASR & $\mu$ $\pm$ sd    & ASR \\
\SetCell{r=3}{\rotatebox[origin=c]{90} {Attack}}  & Grad & \SetCell{r=3}{\rotatebox[origin=c]{90} {SF}}&.23 $\pm$ .04&N/A&.13 $\pm$ .03&N/A\\
                                                  & G-C  &                                                  &.28$ \pm $ .05& N/A &.29 $ \pm $ .06&N/A     \\
                                                  & R-C  &                                                  &.22 $ \pm $.07  & N/A & .24 $\pm $.06   &N/A    \\
                                                  & Grad & \SetCell{r=3}{\rotatebox[origin=c]{90} {RH}}     & .19 $\pm$ .05    & 1   & .13 $\pm$ .03     &   1  \\
                                                  & G-C  &                                                  & .30 $\pm$ .05    & 1   & .33 $\pm$ .09     &    .93 \\
                                                  & R-C  &                                                  &.27 $\pm$ .03& 1   & .30 $\pm$ .10     & .86\\
                                                  & Grad & \SetCell{r=3}{\rotatebox[origin=c]{90} {FD}}     &.07 $\pm$.06    & 1   & .02 $\pm$ .01     &.92 \\
                                                  & G-C &                                                  &.07 $\pm$.06    & 1   & .09 $\pm$ .08     &   .96  \\
                                                  & R-C  &                                                  &.05 $\pm$ .01    & 1   & .02 $\pm$ .03     & .98 \\
                                                  
\end{tblr}
\end{table}

% Show the table for two dataset and soft relu

% Add additional table for ASR of full disguise an others 

% Add table for accuracy and F1 scoe for C# and C$ 20XX or 80XX 

% Add anohter table C5 and C6  40XXX exp

% Show the result of GTSRB dataset 

\section{Ablation study}
\label{ablation}
\begin{figure}[ht]%
    \centering
    \subfloat{{\includegraphics[width=0.49\linewidth, trim={0cm 0 0cm 0},clip]{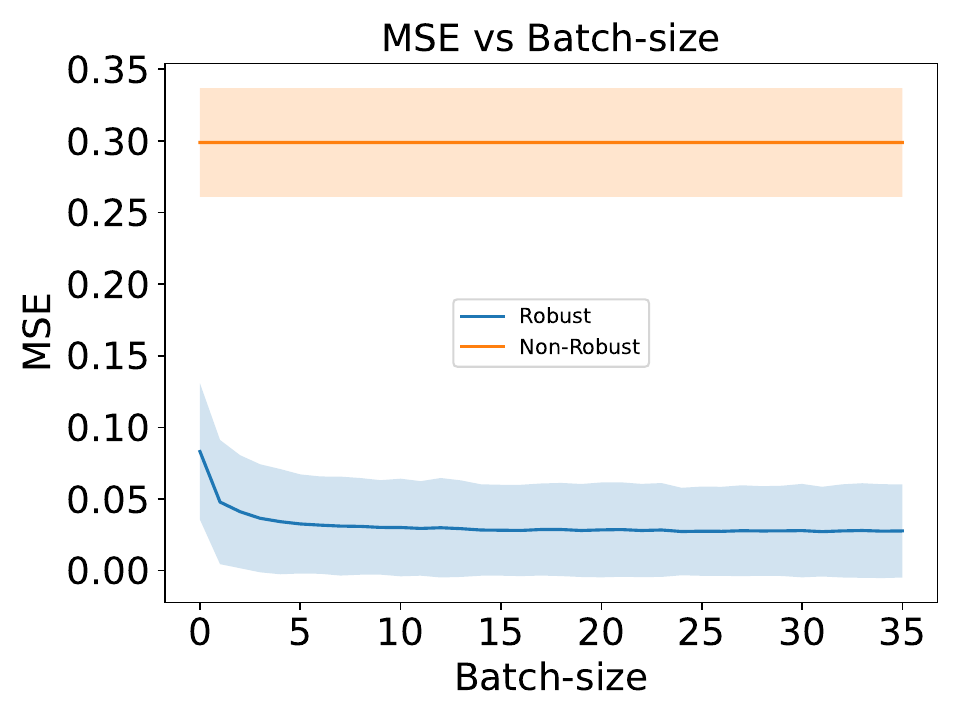} }}%
    \subfloat{{\includegraphics[width=0.49\linewidth, trim={0cm 0cm 0cm 0},clip]{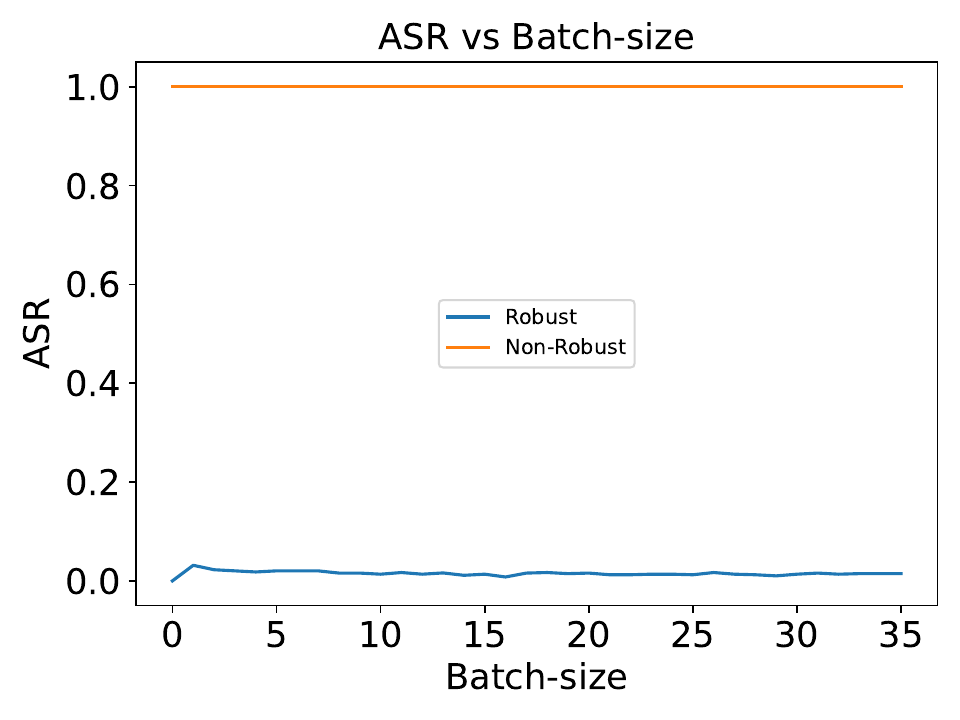}}}
    \caption{This figure delineates the correlation between the batch size and ASR  along with the mean $\pm$ sd explanation MSE during defense (Robust). The trend indicates that the mean $\pm$ sd explanation  MSE exhibits a slight dependence on the batch size. ASR show is also very low for our Robust method. Nevertheless, when the batch size transcends the value of five, it proves adequate in minimizing the mean square error (MSE).}%
    \label{fig:asr_mse_perbatch}%
\end{figure}

For the understanding of the role Batch Normalization plays in safeguarding the model core weights against potential attacks, we investigated whether the complete avoidance of Batch Normalization or the omission of learning Batch Normalization parameters within the model architecture could be a defense. Our findings indicate that under either circumstance, the models remain susceptible to attacks. Table \ref{tab:nbn_fbn_at} presents the outcomes of the three attacks executed on models void of BN and those with non-trainable BN within the models' architecture. In Table \ref{tab:nbn_fbn_at}, the three distinct attacks consistently yield an ASR value of 1 for both RH and FD, while causing relatively high MSE in RH and SF.

This implies that attacks on models without Batch Normalization or models with non-trainable Batch Normalization are not only feasible, but can also be performed with relative ease.

Our proposed solution, CFN, depends on the batch size during the evaluation. A larger batch size results in a lower MSE score and lower ASR. To understand the actual impact of the batch size, we conducted an experiment on an attacked (RH) model  with Relevance-CAM explainer with varying batch sizes, presented in Figure \ref{fig:asr_mse_perbatch}. This figure reveals that even with a batch size of 1, the explanation sustains minimal MSE error and a negligible ASR. With an increase in batch size, the MSE further diminishes.

\section{Discussion and Conclusion}
In this paper, we demonstrated how Batch Normalization (BN) plays a pivotal role in safeguarding model weights during attack fine-tuning in CNN-based architectures. However, we also indicated that the two learnable parameters ($\gamma$ and $\beta$) in the BN layer adversely affect the explanation and prediction during evaluation. Simply removing these parameters and replacing them with Channel-wise Feature Normalisation (CFN) lowers the attack success rate for both explanation and classification. We also proved that removing Batch Normalization cannot defend against an attack.
We demonstrated that our approach can defense three different kinds of attacks. It has been tested on two different datasets and can be applied to any local XAI method by simply replacing the BN layer with CFN. A slight drawback of our method is that it requires a batch size greater than 5 for optimal performance. A larger batch size during evaluation essentially aids the CFN in normalizing the intermediate feature representation neural network, much like using a BN layer in an un-attacked situation.
We also observed that the quality of the Gradient explanation technique does not improve as anticipated. We believe this is an inherent issue with the Gradient explanation method, given its instability and potential for producing random outputs \cite{adebayo_2020_sanitycheck}.
It could be posited that an attacker might circumvent the training parameters of Batch Normalization (BN) or even omit Batch Normalization entirely, while solely fine-tuning the model weights during an attack. However, this strategy might not consistently yield beneficial results, as it introduces significant discrepancies between the original Batch Normalization parameters and the preceding layers in the model's core weights.
Such changes can reduce the model's accuracy on clean data, leading to unsuccessful attacks.
This is possible, as the attack fine-tuning data embodies a distinct distribution owing to the presence of artefacts. Consequently, the model learns this distribution, thereby deviating from the original data distribution. For details please refer to the Supplementary section \ref{sec:attack_sceen}. Additional examples of defense for both datasets can be found in the Supplementary section \ref{fig:additional_figures}.

In future work, we plan to investigate how to defend against attacks on models that do not contain Batch Normalization layers by removing common artifacts from the intermediate feature representations. 

% This is because there is a substantial change in the model's weights if we employ layer-wise feature normalization. 

{
    \small
    \bibliographystyle{ieeenat_fullname}
    \bibliography{bibliography}
}


\clearpage
\setcounter{page}{1}
\maketitlesupplementary
\section{Attack Scenarios}
\label{sec:attack_sceen}
Table \ref{tab:attack_scenarios} presents various attack and evaluation scenarios for models. Each column corresponds to a different scenario, where each row represents the activities required in that particular scenario. The presence of a $\bullet$ symbol indicates the application of a particular activity in that row, while $\circ$ signifies that the activity was not involved. There are six crucial activities where a potential compromise to a model's explanation and prediction could be brought about by an attacker.

Our empirical analysis demonstrates that executing attack activities under conditions \textbf{C3} - \textbf{C6} significantly impacts the model's weights. This results in a minimal accuracy on internal test data, as shown in Table \ref{tab:accdrop}. Notably, if the model's test accuracy drops significantly due to an attack, it becomes easily detectable by the test accuracy. 

In this study, we tackle the attack scenario presented in \textbf{C1} of Table \ref{tab:attack_scenarios}. In this scenario, it is impossible to detect an attack based on the accuracy of the test data, as there is no noticeable change or decline \cite{noppel2023disguising}. However, a trigger in the input causes significant changes in the local explanations and the overall classification of the model. To counter this type of attack, we introduce defence in \textbf{C2}. This solution involves incorporating CFN after each CNN of a model by replacing BN during the classification process and explanation generation process. 

\begin{table}[ht]
\centering
\caption{
Attack scenarios \textbf{C1}, \textbf{C3}, and \textbf{C5}, and corresponding defense scenarios \textbf{C2}, \textbf{C4}, and \textbf{C6} are applicable to the model. The \textbf{C1} scenario subtly corrupts the BN layer, remaining undetected as it does not alter test accuracy. In contrast, \textbf{C3} involves exclusive fine-tuning of core weights, substantially impacting model accuracy (Table \ref{tab:accdrop}). \textbf{C5} involves replacing BN with CFN, also leading to a significant reduction in test accuracy (Table \ref{tab:accdrop}). For defense, \textbf{C2} replaces corrupted batch norms with CFN. However, \textbf{C3} and \textbf{C5} attacks prove ineffective due to low test accuracy on clean data. Our defenses, \textbf{C4} and \textbf{C6}, incorporate CFN, while still yielding low test accuracy. Low test accuracy thus serves as a preliminary check of corruption. Hence, the only viable attack is \textbf{C1} and its corresponding defense, \textbf{C2}.}

\label{tab:attack_scenarios}
\begin{tabular}{|c|l|c|c|c|c|c|c|}
\hline
\multicolumn{2}{|c|}{\textbf{Options}} & \textbf{C1} & \textbf{C2} & \textbf{C3} & \textbf{C4} & \textbf{C5} & \textbf{C6} \\ 
\hline
\multirow{3}{*}{\rotatebox[origin=c]{90}{Attacker}} & Model Weight& $\bullet$  & $\bullet$  & $\bullet$ & $\bullet$ & $\bullet$    & $\bullet$  \\ 
\cline{2-8} 
& Batch Norm   & $\bullet$  & $\bullet$  & $\circ$    & $\circ$      & $\circ$ &$\circ$    \\ 
\cline{2-8} 
& CFN  & $\circ$ & $\circ$ & $\circ$ & $\circ$ & $\bullet$ & $\bullet$ \\ 
\hline
\multirow{6}{*}{\rotatebox[origin=c]{90}{Defender}} & Model Weight& $\bullet$ & $\bullet$ & $\bullet$ & $\bullet$ & $\bullet$ & $\bullet$ \\ 
\cline{2-8} 
& Batch Norm   & $\bullet$  & $\circ$    & $\bullet$    & $\circ$  & $\bullet$      & $\circ$    \\ 
\cline{2-8} 
& CFN          & $\circ$    & $\bullet$  & $\circ$  & $\bullet$  & $\circ$   & $\bullet$  \\ 
\cline{2-8} 
& ACC        & $\uparrow$ & $\uparrow$ & $\downarrow$ & $\downarrow$ & $\downarrow$ & $\downarrow$ \\ 
\cline{2-8} 
& EXP(T)          & $\bullet$  & $\circ$    & $\circ$  & $\circ$    & $\circ$      & $\circ$  \\ 
\cline{2-8} 
& Attack          & $\bullet$  & $\circ$    & $\circ$    & $\circ$      & $\circ$ & $\circ$  \\ 
\hline
\end{tabular}
\end{table}

\begin{table}[ht]
\centering
\caption{This table describes test data (CIFAR10) accuracy pre-attack (Org (Acc)), post-attack (\textbf{C3} and \textbf{C5}), and post-defense (\textbf{C4} and \textbf{C6}). BN-Acc indicates defense models employing original batch normalization parameters, whilst CFN-Acc represents defense using CFN. It is evident that attack scenarios (\textbf{C3} and \textbf{C5}) are inconsistent as they fail to match the original model's accuracy on clean data. The results are based on nine instances, each representing an explanation technique and each attack, shown as the mean $\pm$ standard deviation.}
\label{tab:accdrop}
\begin{tabular}{|c|c|c|c|}
\hline
\textbf{Att./ Def.} & Org (Acc)& \textbf{BN-Acc} & \textbf{CFN-Acc} \\ 
\hline
C3 \& C4  & 91.3 $\pm$ .7  & 11.5 $\pm$ 2.21 & 37.5 $\pm$ 13.6 \\
\hline
C5 \& C6  &91.3 $\pm$ .7  & 12.45 $\pm$ 3.5 & 47.4 $\pm$ 16.03\\
\hline
\end{tabular}
\end{table}
\section{Addition Functions}
\subsection{Batch Normalization}
\label{eq:bn}
Batch-Normalization (BN) is a technique that enhances the speed and stability of Deep Neural Networks (DNN) training \cite{sergey2015bn}. It normalizes activation vectors from hidden layers based on the current batch's mean and variance.
\subsubsection*{BN in Training}
Assume Z represents the activation of a hidden layer. In the presence of a BN layer, a function denoted as $f_{bn}$, computes the mean ($\mu$) and variance ($\sigma$) over the channel axis.
\begin{align}
\mu &= \frac{1}{m} \sum_{i=1}^{m} Z_i, \quad \\
\sigma&= \frac{1}{m} \sum_{i=1}^{m} (Z_i - \mu)^2 %\tag{1,2}
\end{align}

By applying equations (1) and (2), we can derive the normalized activation, denoted as $Z_{bn}$.

\begin{equation}
Z_{bn} = \frac{Z - \mu }{\sqrt{\sigma + \epsilon}}
\end{equation}

To avoid division by zero and the resulting infinity, we add a small constant, usually denoted as epsilon ($\epsilon$), to the denominator. 

During the testing $\hat{\mu}$ and $\hat{\sigma}$ are used instead of $\mu$ and $\sigma$
\begin{equation}
\hat{\mu} = m \cdot \hat{\mu} + (1 - m) \cdot \mu
\end{equation}
 $m$ is a hyperparameter between 0 and 1 representing the weight given to the previous value of $\hat{\mu}$. $\hat{\sigma}$ is also calculated similar way.
During the training process of the model, the parameters ($\beta$ and $\gamma$) of the dropout layer are concurrently optimized. The relationship between these learning parameters and the normalized activation $Z_{bn}$ is represented by Equation \ref{eq:dropout}.

\begin{equation} \label{eq:dropout}
Z_{out} = \gamma \cdot Z_{bn} + \beta
\end{equation}

With each cycle, the network calculates the current batch's average ($\mu$) and standard deviation ($\sigma$).
Subsequently, it educates $\gamma$ and $\beta$ using gradient descent and applies an Exponential Moving Average (EMA) to emphasize the significance of the most recent cycles.

\section{Additional Figures and Tables}
\label{additialfig_and_table}
\begin{figure}[ht]%
    \centering
    {{\includegraphics[width=0.8\linewidth, trim={0cm 0 0cm 0},clip]{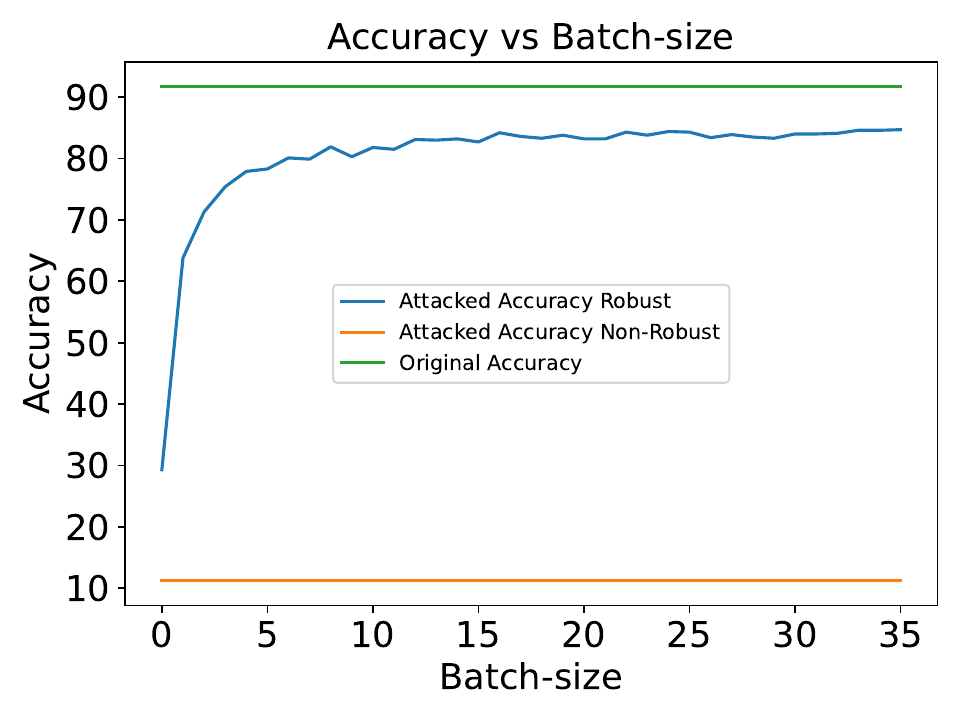} }}%
    \caption{This figure is related to the figure 8 where it shows slight correlation between accuracy and batch size. When batch size transcends five the Attacked Robust accuracy is reach an adequate level.}%
    \label{fig:softplus_acc}%
\end{figure}

\begin{figure}[ht]
\centering
\includegraphics[width=0.48\textwidth]{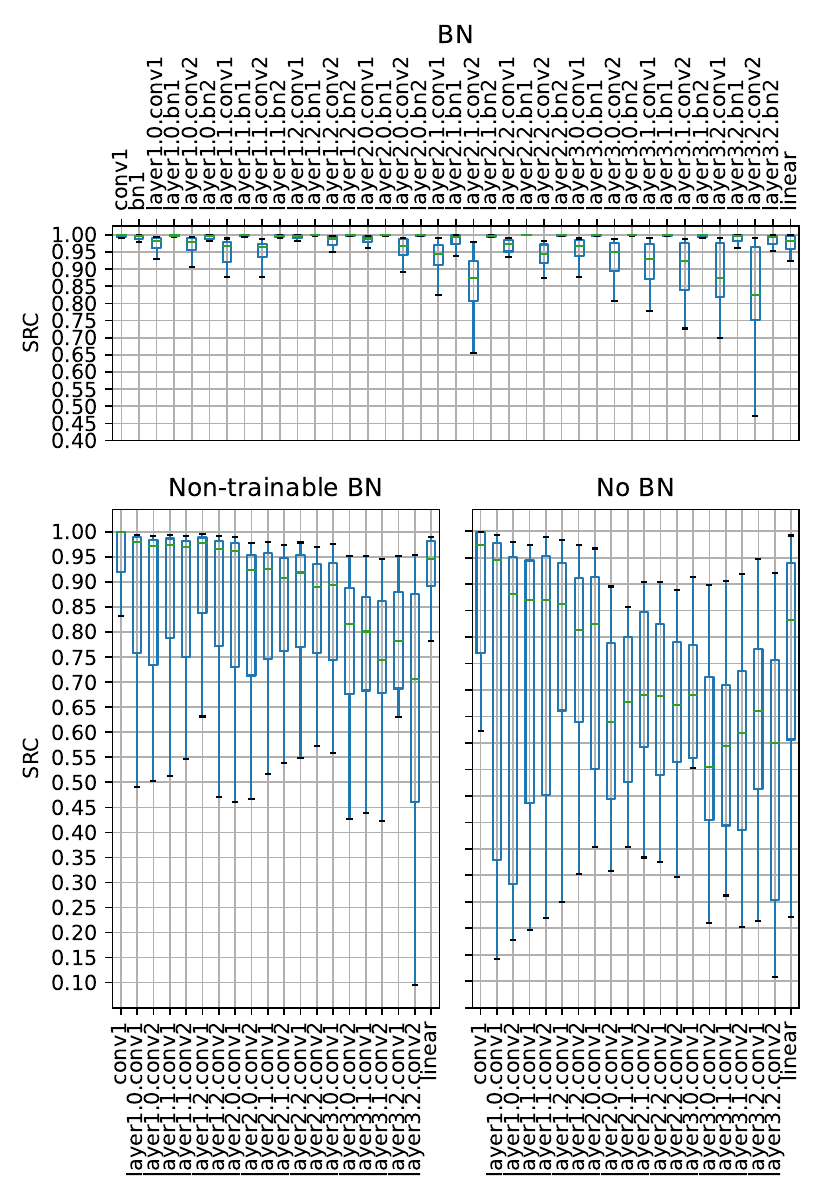}
\caption{The Spearman's Rank Correlation (SRC) between layers of original models and attacked models is presented. The top sub-graph displays
the correlation where models contain Batch Normalization (BN) layers.
The bottom left sub-figure presents correlation scores between models
that do not train BN parameters. The bottom right sub-figure illus
trates correlation scores between models that lack any BN layers.} 
\label{fig:src_src_plot}
\end{figure}

\begin{figure}[ht]
\centering
\includegraphics[width=0.48\textwidth]{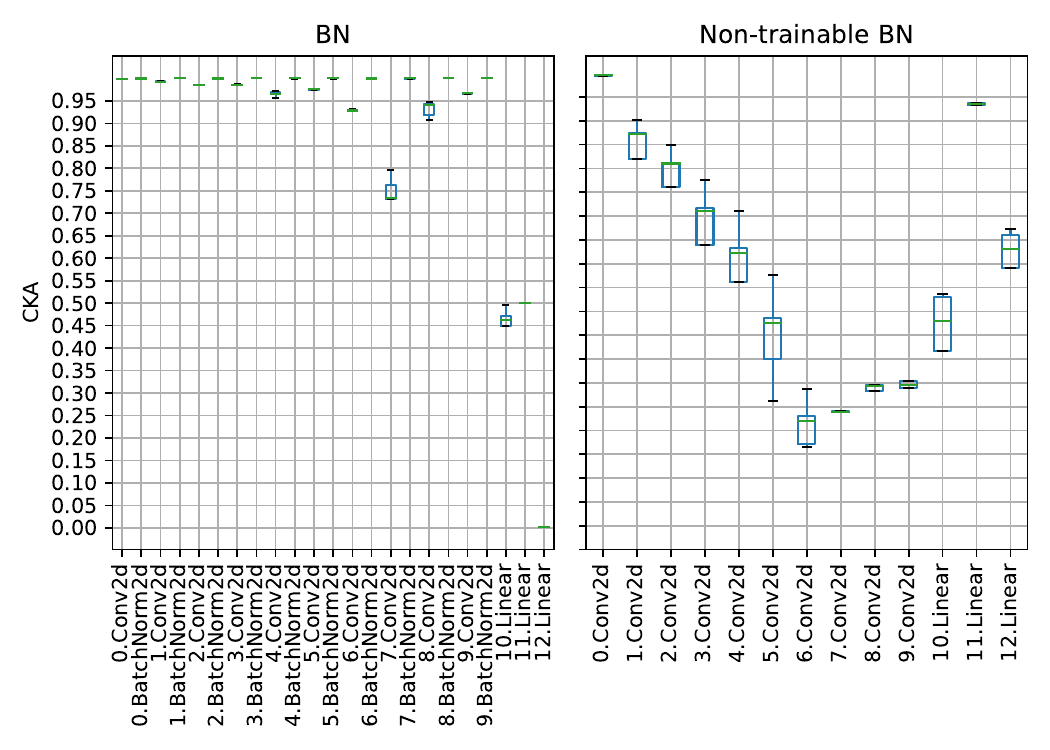}
\caption{CKA depiction comparing original and attacked VGG13 models. Left subplot indicates correlation in models with Batch Normalization (BN) layers. Right subplot represents correlations in models lacking trainable BN parameters. Fewer core weight changes are apparent in BN models compared to non-trainable BN versions. We also noticed similar trend in ResNet20 architecture.}
\label{fig:vgg}
\end{figure}

\begin{figure}[tbh]
  \setlength\tabcolsep{1pt}
  \adjustboxset{width=\linewidth,valign=c}
  \centering
  \begin{tabularx}{1.0\linewidth}{@{}
      l
      X @{\hspace{6pt}}
      X
    @{}}
    & \multicolumn{1}{c}{Attack}
    & \multicolumn{1}{c}{Defense} \\
    \rotatebox[origin=c]{90}{CIFAR10}
    & \includegraphics[width=\linewidth, trim={0.6cm 0 0.6cm 0},clip]{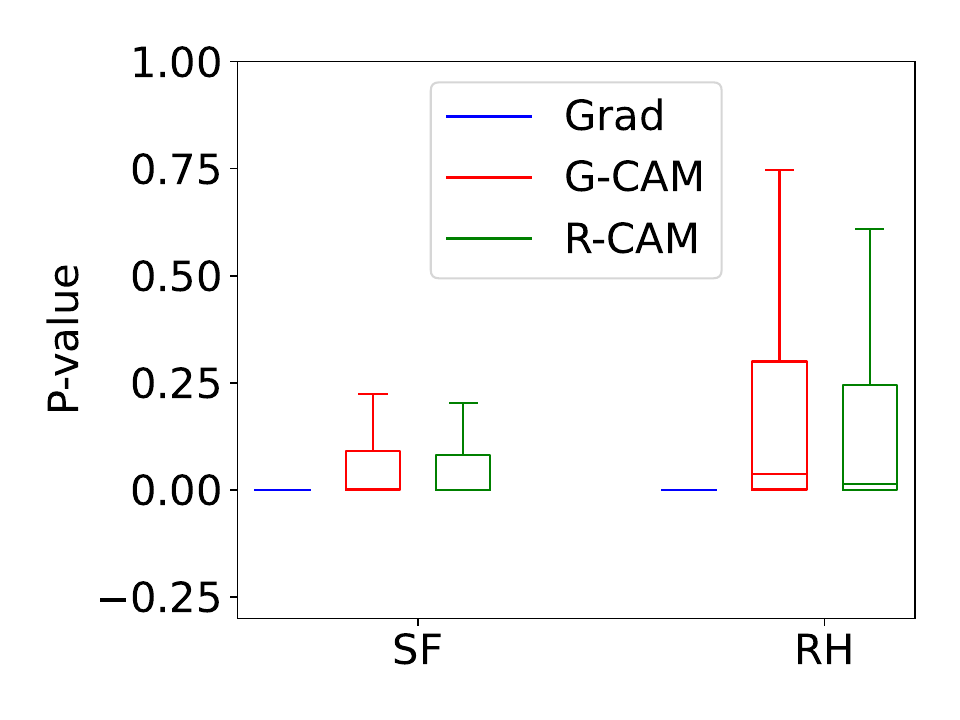}
    & \includegraphics[width=\linewidth, trim={0.6cm 0cm 0cm 0},clip]{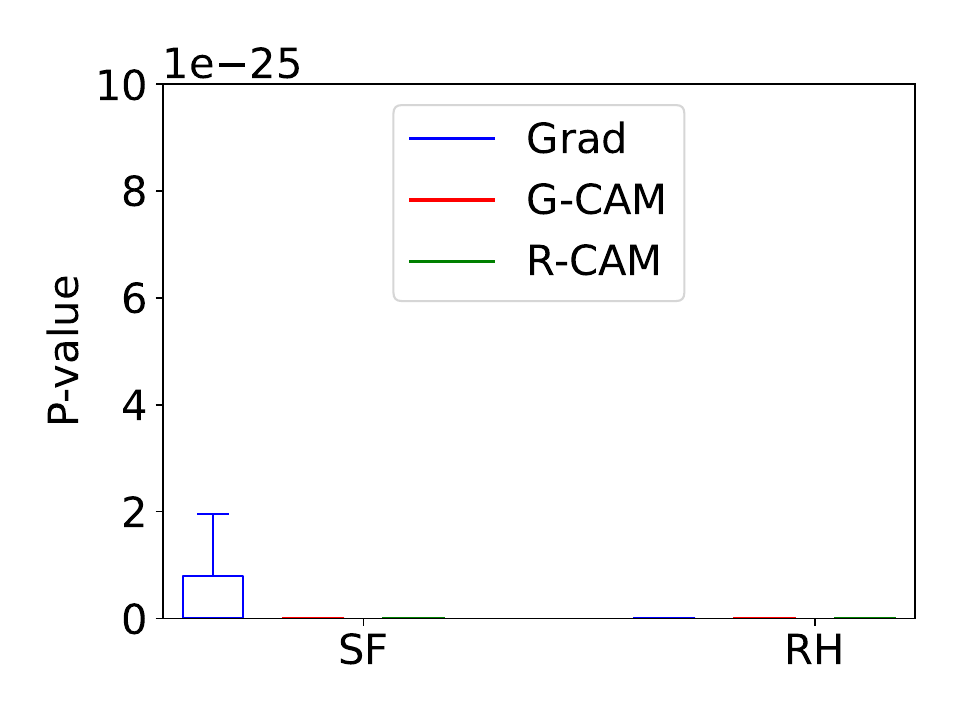} \\
    \rotatebox[origin=c]{90}{GTSRB}
    & \includegraphics[width=\linewidth, trim={0.6cm 0 0.6cm 0},clip]{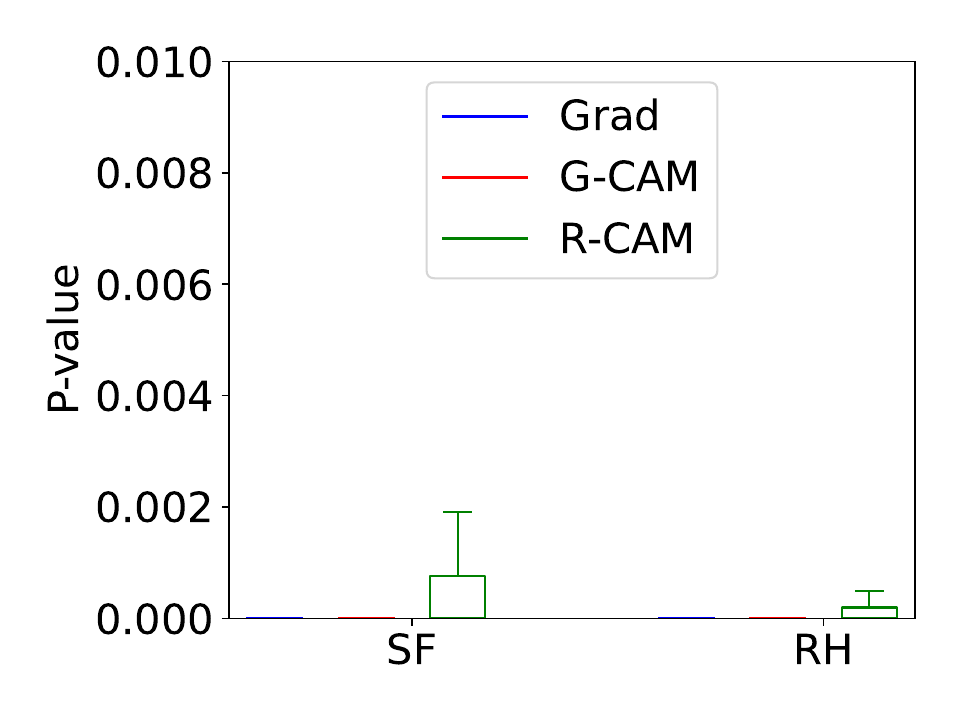}
    & \includegraphics[width=\linewidth, trim={0.6cm 0cm 0cm 0},clip]{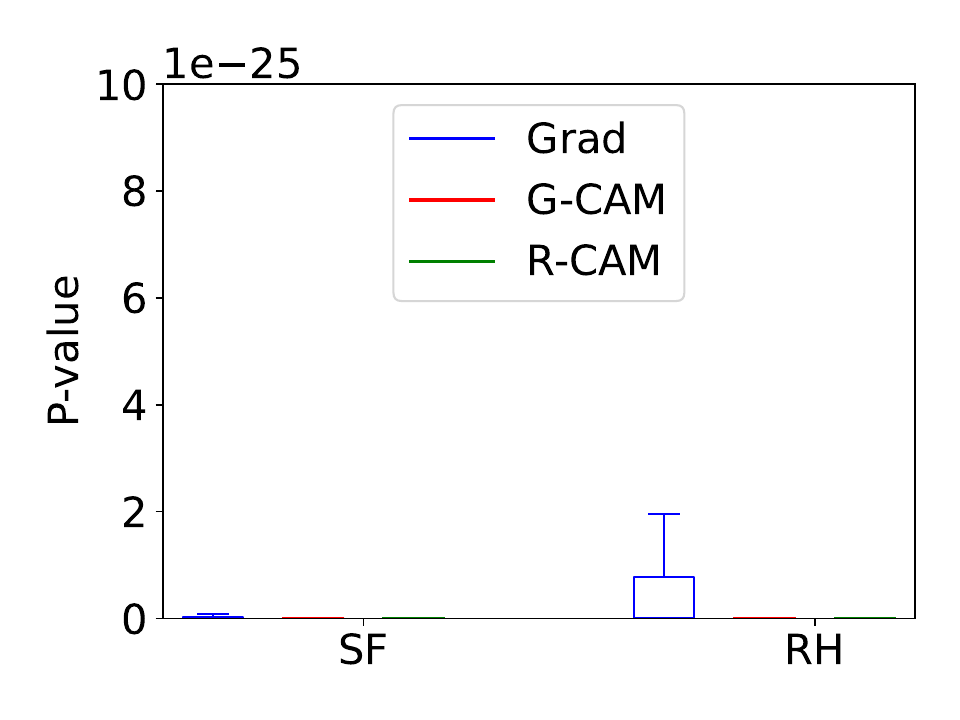}
  \end{tabularx}
  \caption{This figure demonstrates the p-value related to the Spearman's Rank Correlation (SRC) distribution between original model explanations and attacked, then defended explanations on SF, RH attacks related to Figure \ref{fig:sf_rh_src}.
}
\label{fig:sf_rh_src_1}
\end{figure}

\begin{figure}[tbh]
  \setlength\tabcolsep{1pt}
  \adjustboxset{width=\linewidth,valign=c}
  \centering
  \begin{tabularx}{1.0\linewidth}{@{}
      l
      X @{\hspace{6pt}}
      X
    @{}}
    & \multicolumn{1}{c}{Attack}
    & \multicolumn{1}{c}{Defense} \\
    \rotatebox[origin=c]{90}{CIFAR10}
    & \includegraphics[width=\linewidth, trim={0.6cm 0 0.6cm 0},clip]{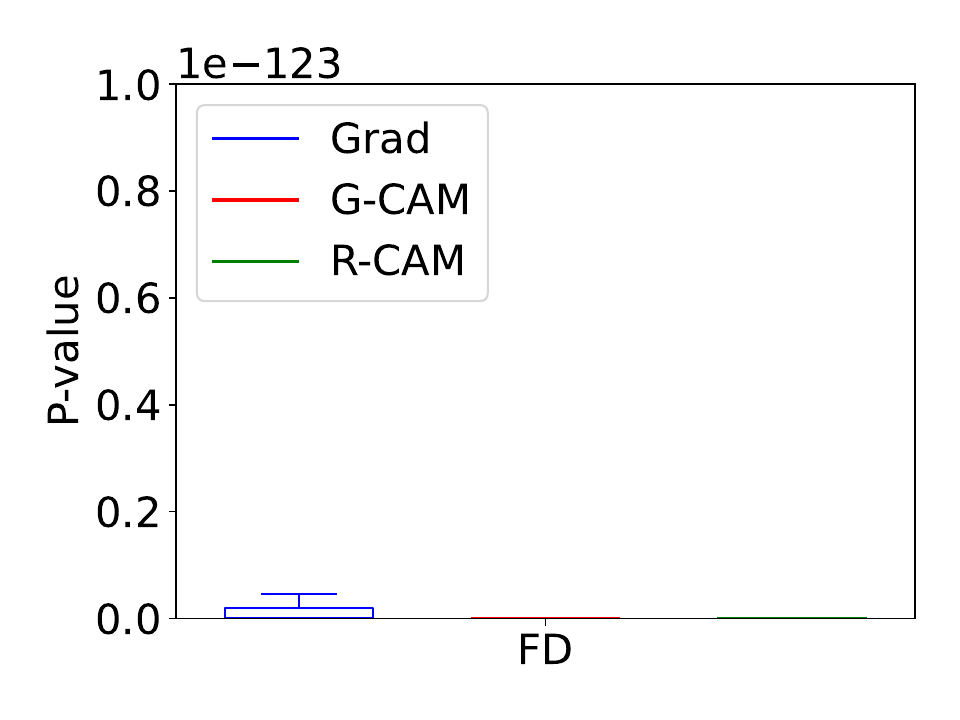}
    & \includegraphics[width=\linewidth, trim={0.6cm 0cm 0cm 0},clip]{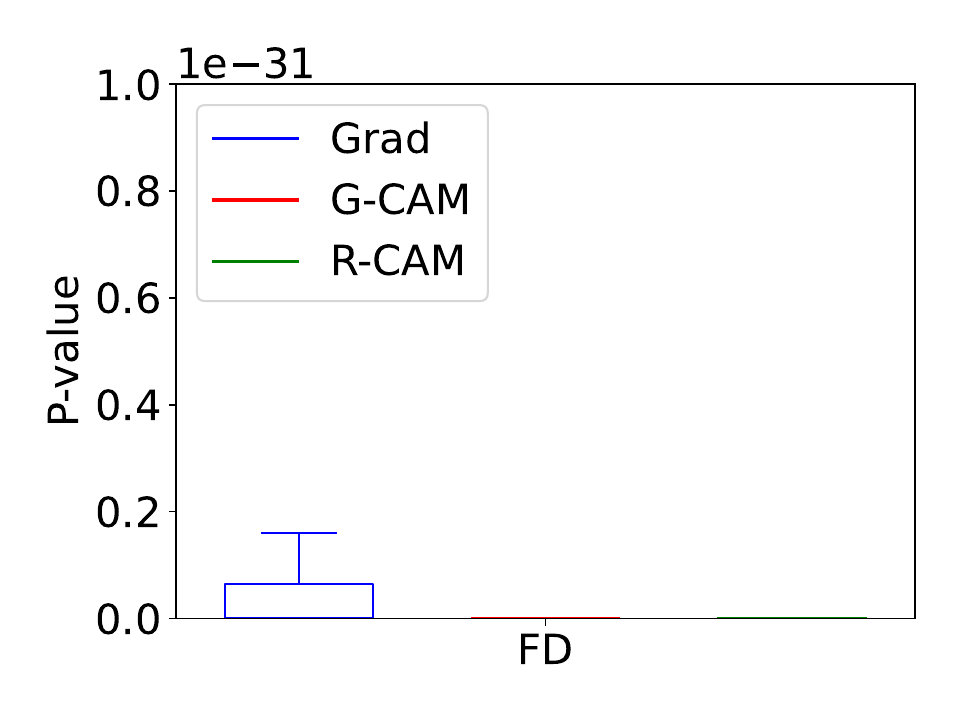} \\
    \rotatebox[origin=c]{90}{GTSRB}
    & \includegraphics[width=\linewidth, trim={0.6cm 0 0.6cm 0},clip]{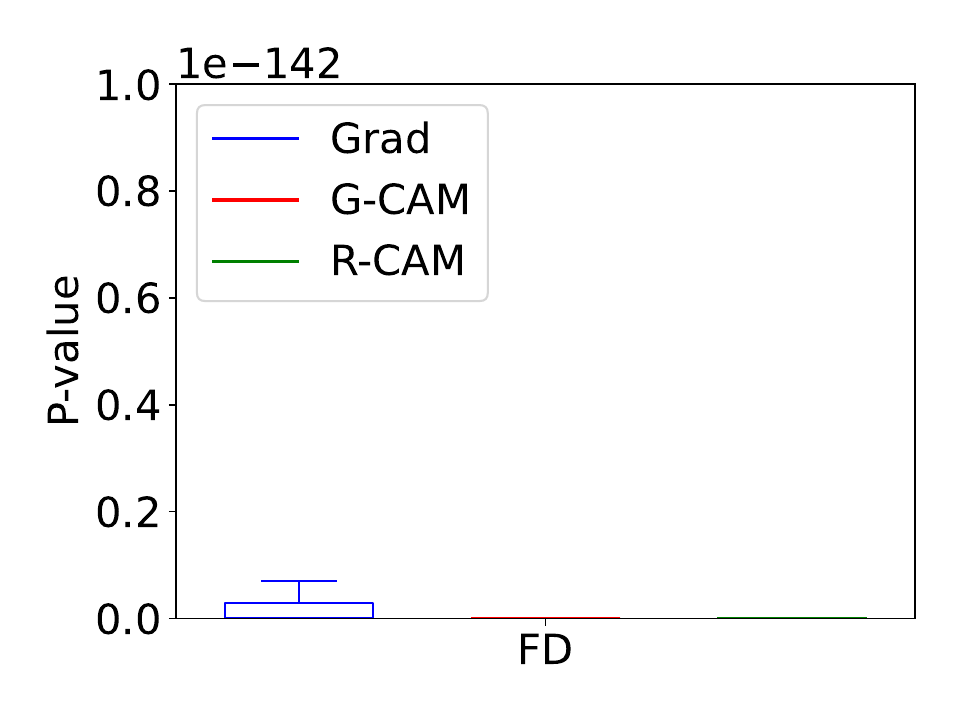}
    & \includegraphics[width=\linewidth, trim={0.6cm 0cm 0cm 0},clip]{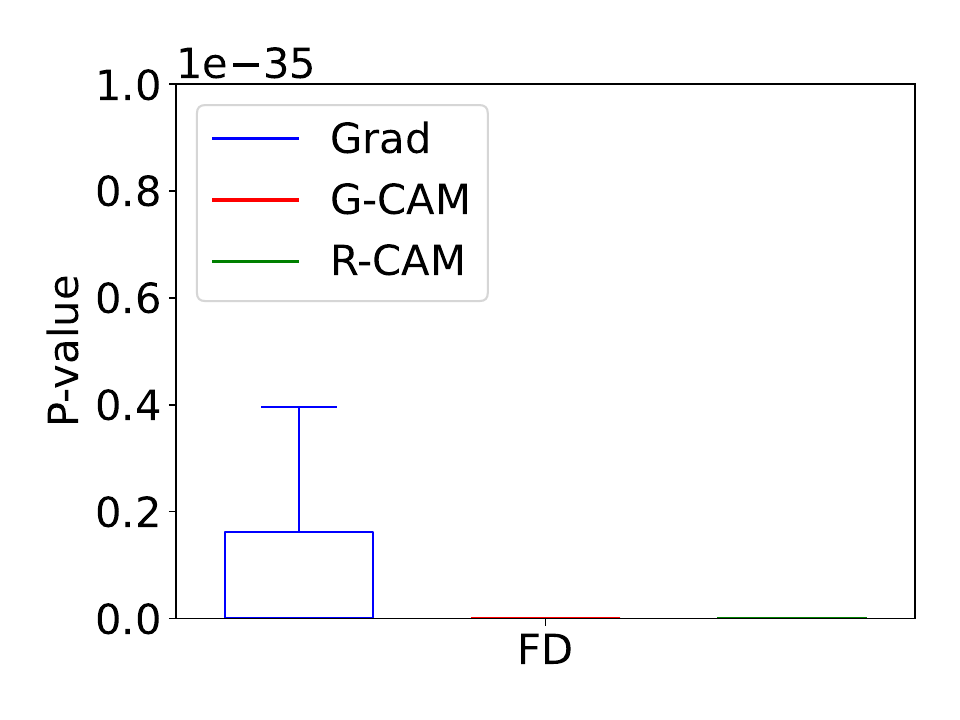}
  \end{tabularx}
  \caption{This figure demonstrates the p-value related to Spearman's Rank Correlation (SRC) distribution between original model explanations and attacked , then defended  explanations after FD attack related to Figure \ref{fig:fd_src}.
}
\label{fig:fd_src_5}
\end{figure}

\begin{figure}[ht]%
    \centering
    {{\includegraphics[width=0.49\linewidth, trim={0.6cm 0 0.6cm 0},clip]{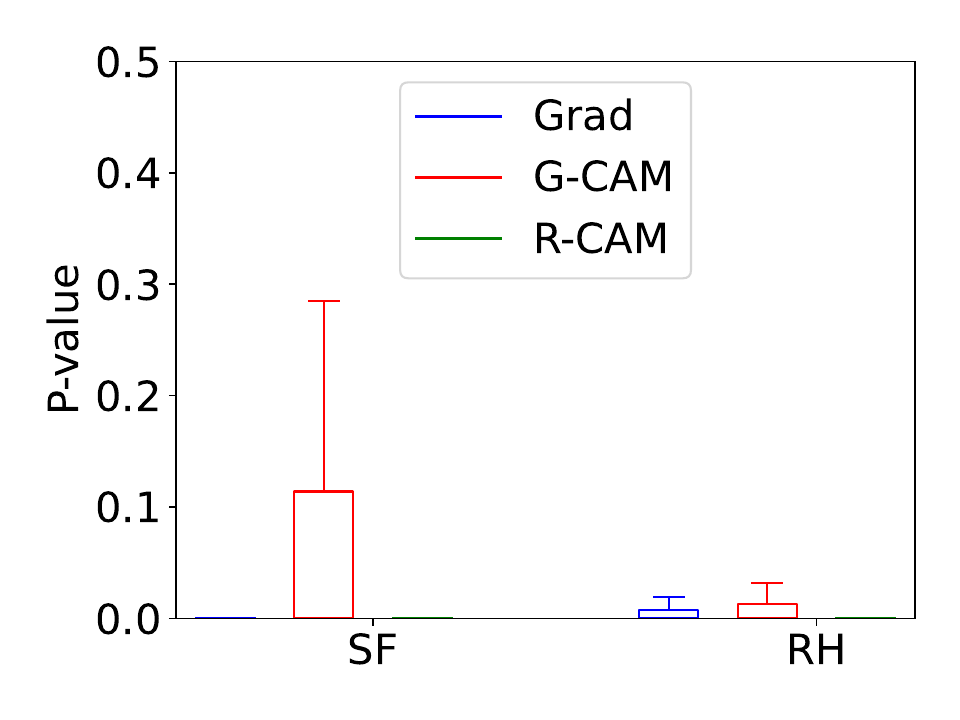} }}%
    {{\includegraphics[width=0.49\linewidth, trim={0.6cm 0cm 0cm 0},clip]{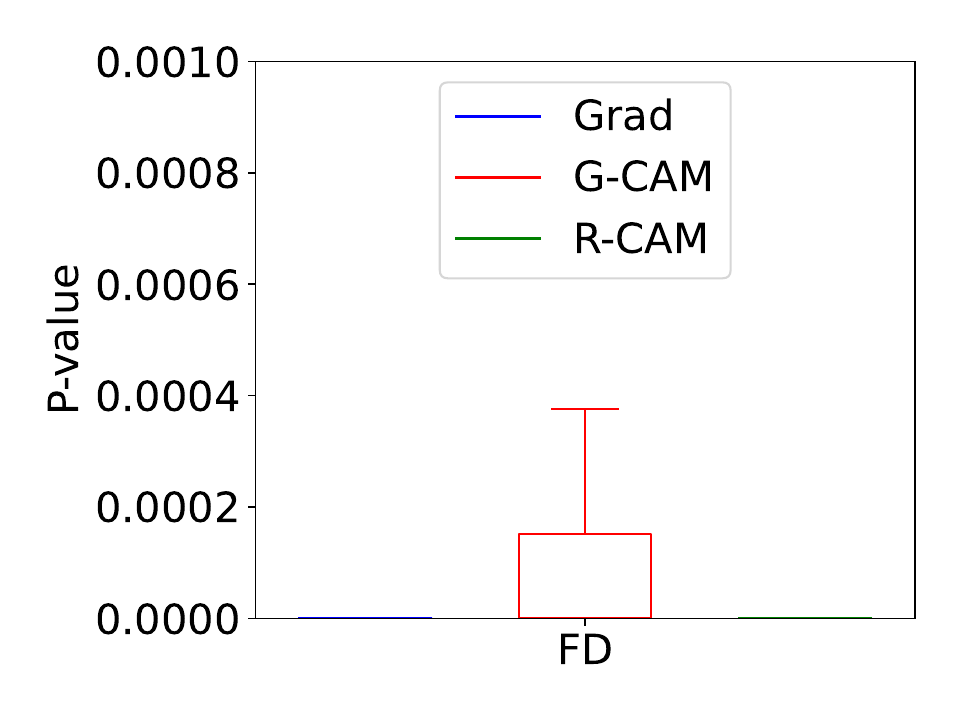} }}%
    \caption{This figure represents the p-value corresponding to Spearman's Rank Correlation (SRC) for the softplus activation's model defense in the Figure \ref{fig:sa_rh_fd_softplus}.}%
    \label{fig:softplus_pvalue}%
\end{figure}

\begin{figure}[tbh]
  \setlength\tabcolsep{1pt}
  \adjustboxset{width=\linewidth,valign=c}
  \centering
  \begin{tabularx}{1.0\linewidth}{@{}
      l
      X @{\hspace{6pt}}
      X
    @{}}
    & \multicolumn{1}{c}{Attack}
    & \multicolumn{1}{c}{Defense} \\
    & \includegraphics[width=\linewidth, trim={0.6cm 0 0.6cm 0},clip]{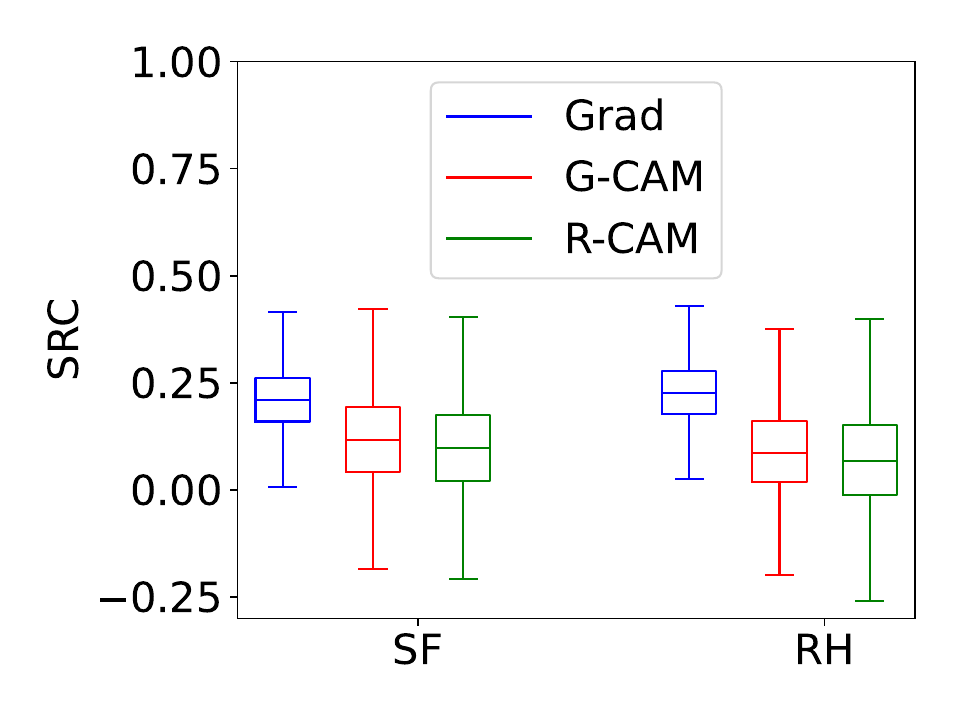}
    & \includegraphics[width=\linewidth, trim={0.6cm 0cm 0cm 0},clip]{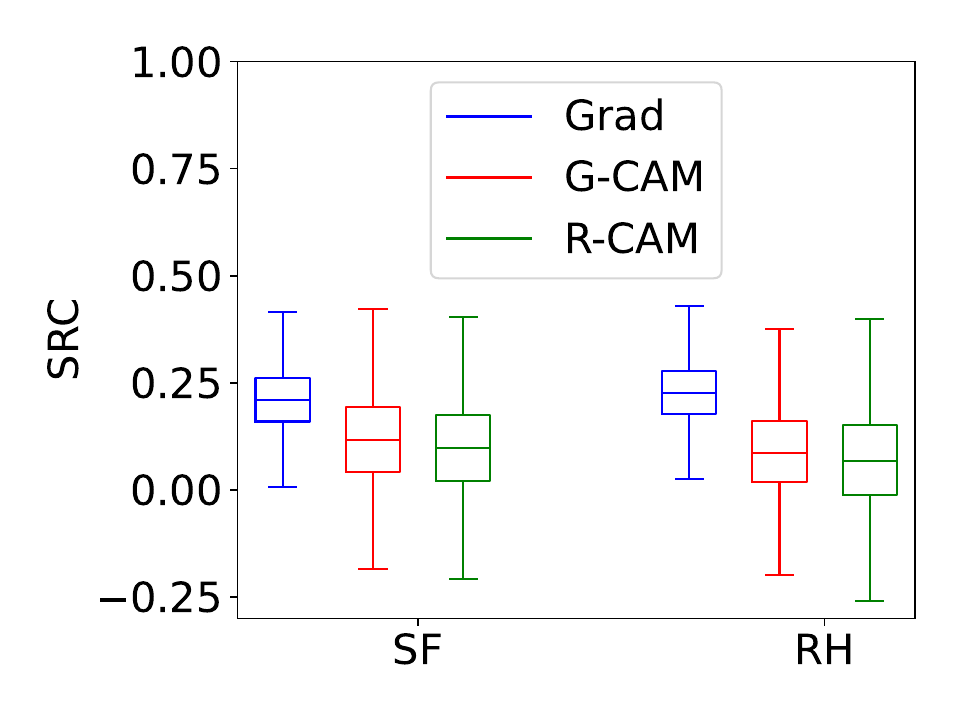} \\
    
    & \includegraphics[width=\linewidth, trim={0.6cm 0 0.6cm 0},clip]{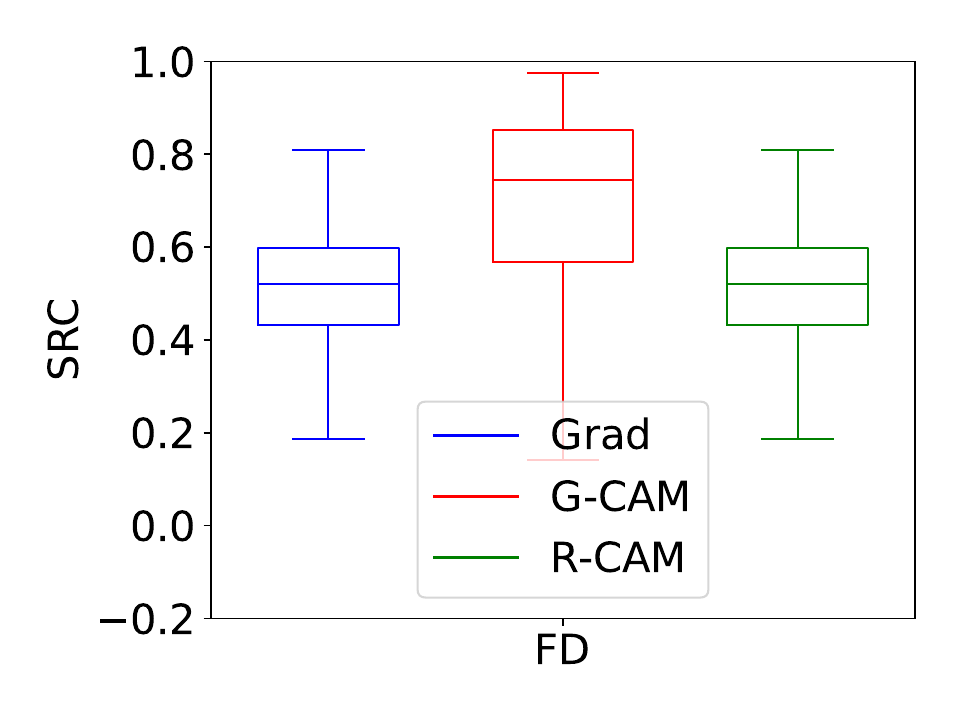}
    & \includegraphics[width=\linewidth, trim={0.6cm 0cm 0cm 0},clip]{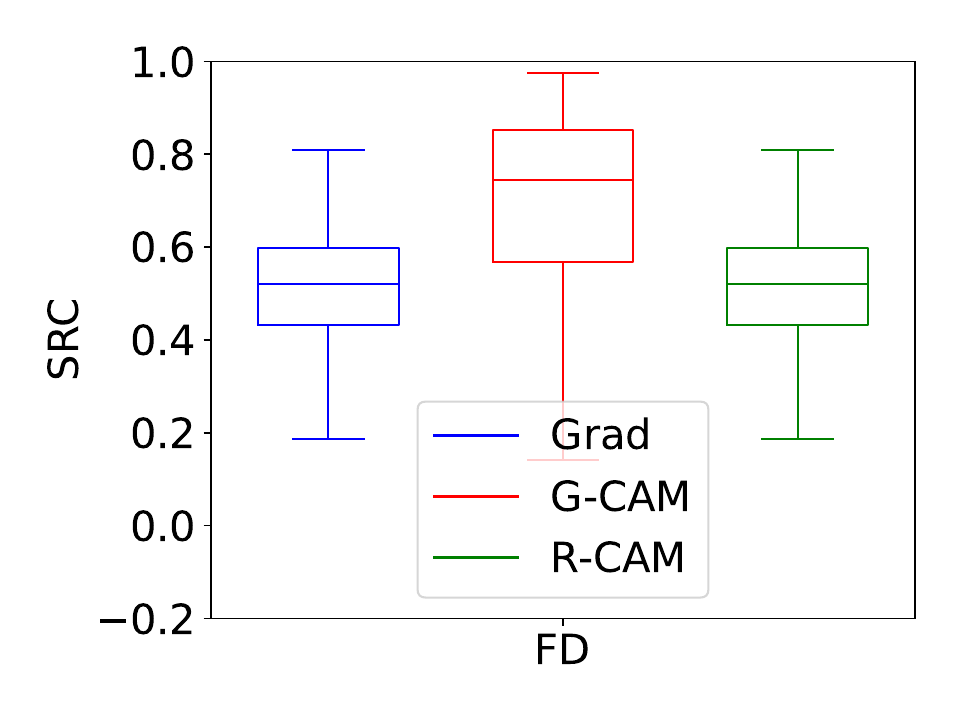}
  \end{tabularx}
  \caption{This figure demonstrates the Spearman's Rank Correlation (SRC) distribution between original model explanations and attacked , then defended explanations
after SF, RH and FD attacks related to No BN in Table \ref{tab:nbn_fbn_at}. 
}
\label{fig:fd_src_4}
\end{figure}

\begin{figure}[tbh]
  \setlength\tabcolsep{1pt}
  \adjustboxset{width=\linewidth,valign=c}
  \centering
  \begin{tabularx}{1.0\linewidth}{@{}
      l
      X @{\hspace{6pt}}
      X
    @{}}
    & \multicolumn{1}{c}{Attack}
    & \multicolumn{1}{c}{Defense} \\
    
    & \includegraphics[width=\linewidth, trim={0.6cm 0 0.6cm 0},clip]{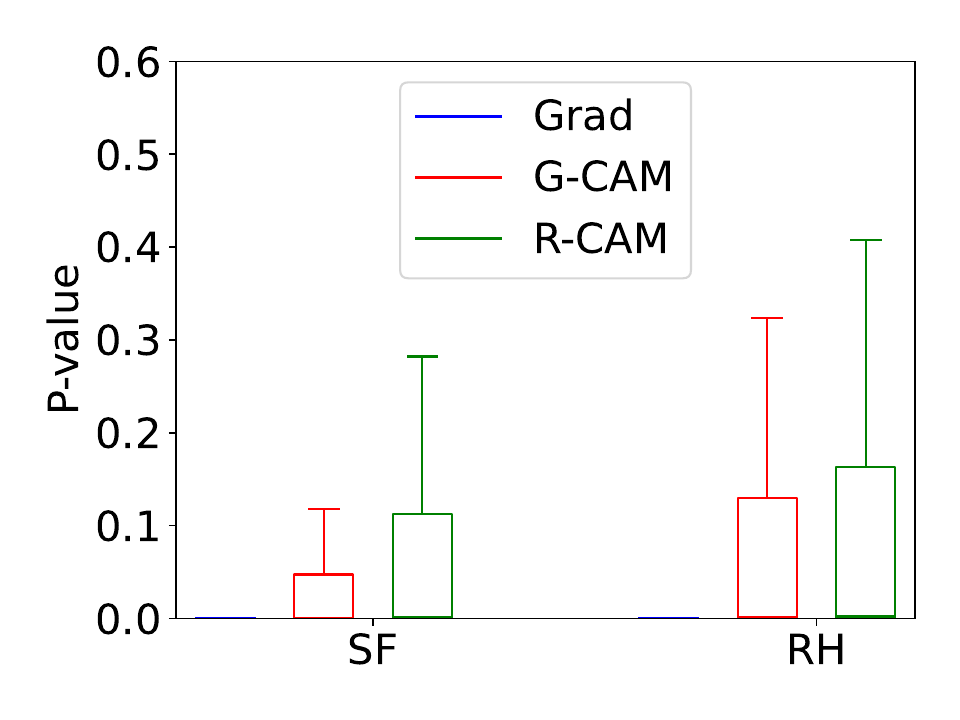}
    & \includegraphics[width=\linewidth, trim={0.6cm 0cm 0cm 0},clip]{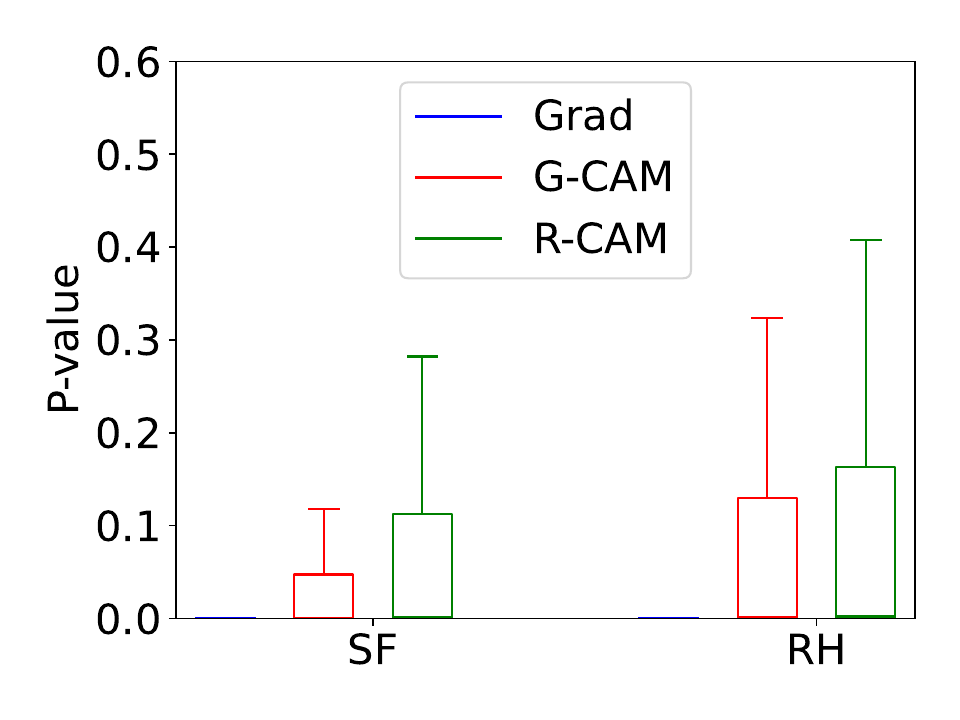} \\
    
    & \includegraphics[width=\linewidth, trim={0.6cm 0 0.6cm 0},clip]{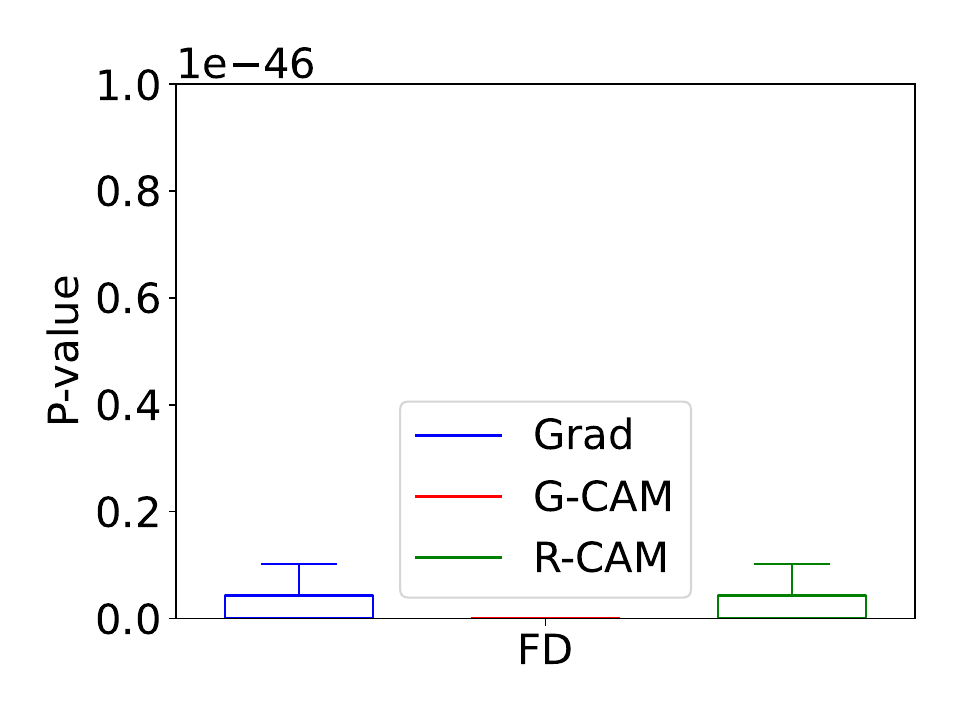}
    & \includegraphics[width=\linewidth, trim={0.6cm 0cm 0cm 0},clip]{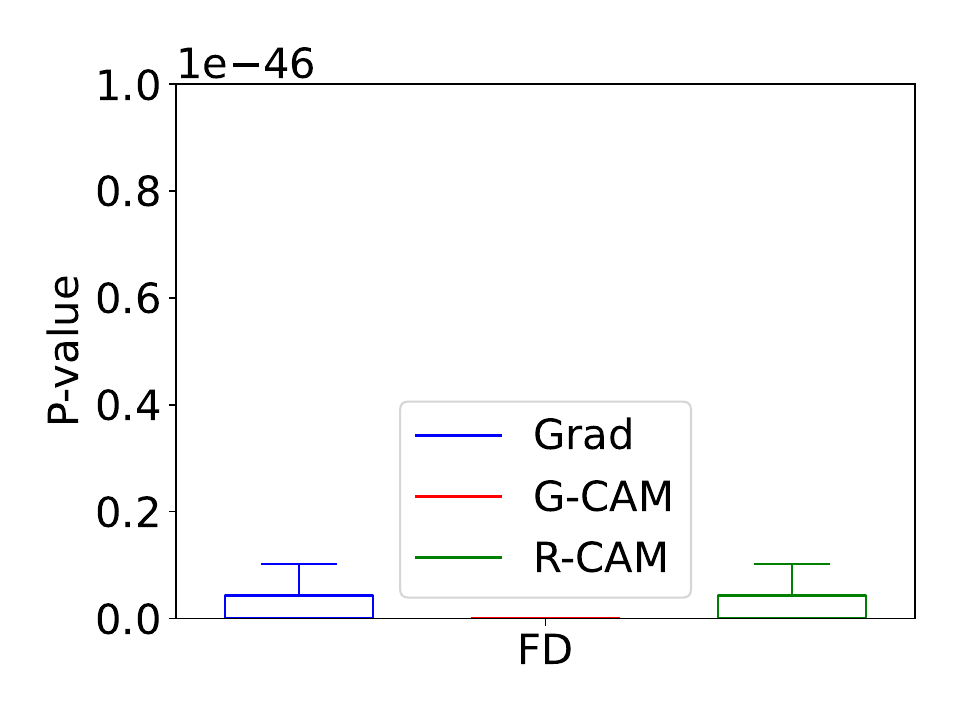}
  \end{tabularx}
  \caption{This figure demonstrates the p-values for the correlation values between original model explanations and attacked , then defended explanations
after SF, RH and FD attacks related to No BN in Figure \ref{fig:fd_src_4}.
}
\label{fig:fd_src_3}
\end{figure}

\begin{figure}[tbh]
  \setlength\tabcolsep{1pt}
  \adjustboxset{width=\linewidth,valign=c}
  \centering
  \begin{tabularx}{1.0\linewidth}{@{}
      l
      X @{\hspace{6pt}}
      X
    @{}}
    & \multicolumn{1}{c}{Attack}
    & \multicolumn{1}{c}{Defense} \\
   
    & \includegraphics[width=\linewidth, trim={0.6cm 0 0.6cm 0},clip]{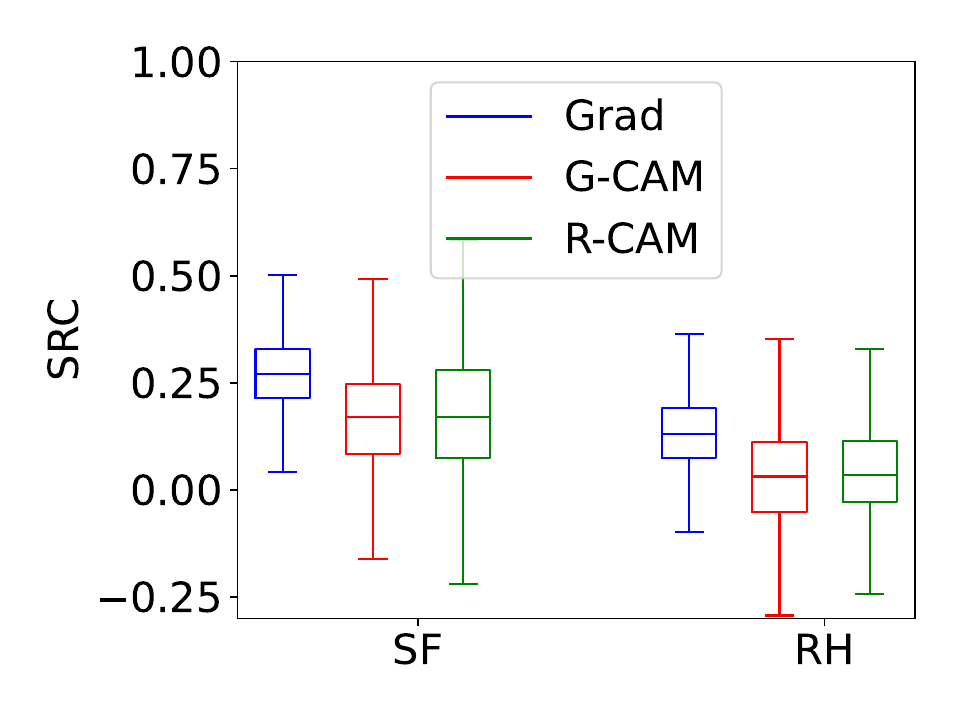}
    & \includegraphics[width=\linewidth, trim={0.6cm 0cm 0cm 0},clip]{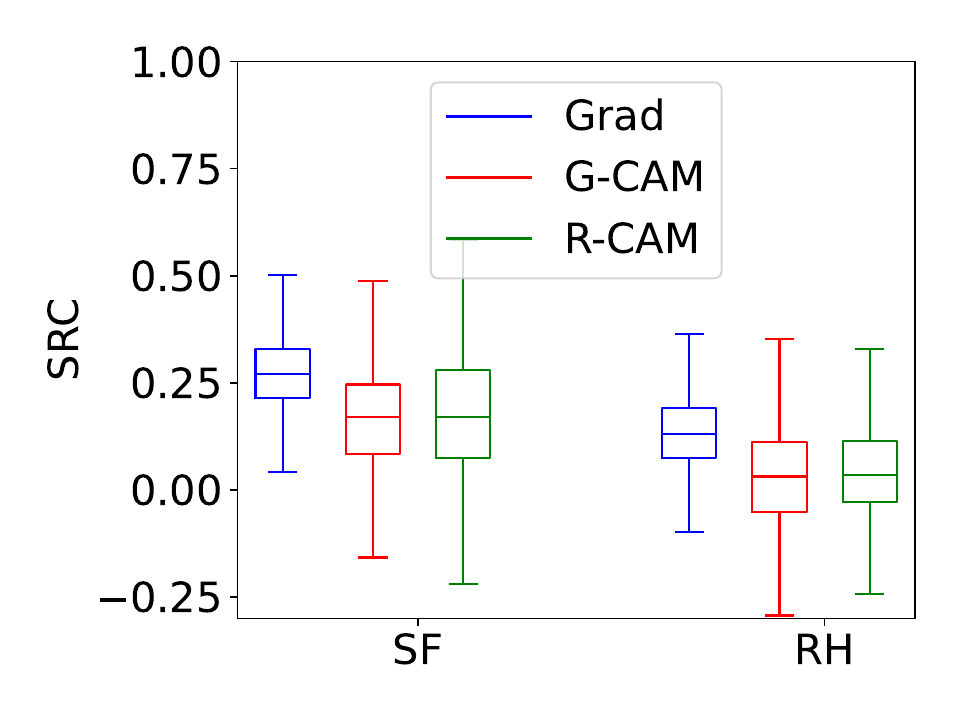} \\
    
    & \includegraphics[width=\linewidth, trim={0.6cm 0 0.6cm 0},clip]{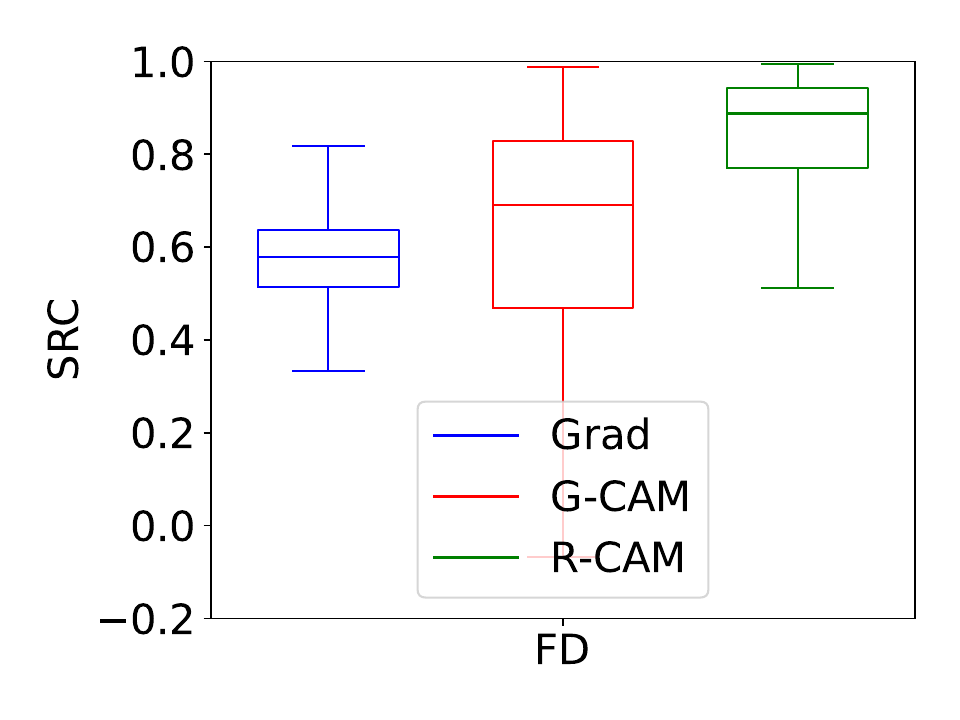}
    & \includegraphics[width=\linewidth, trim={0.6cm 0cm 0cm 0},clip]{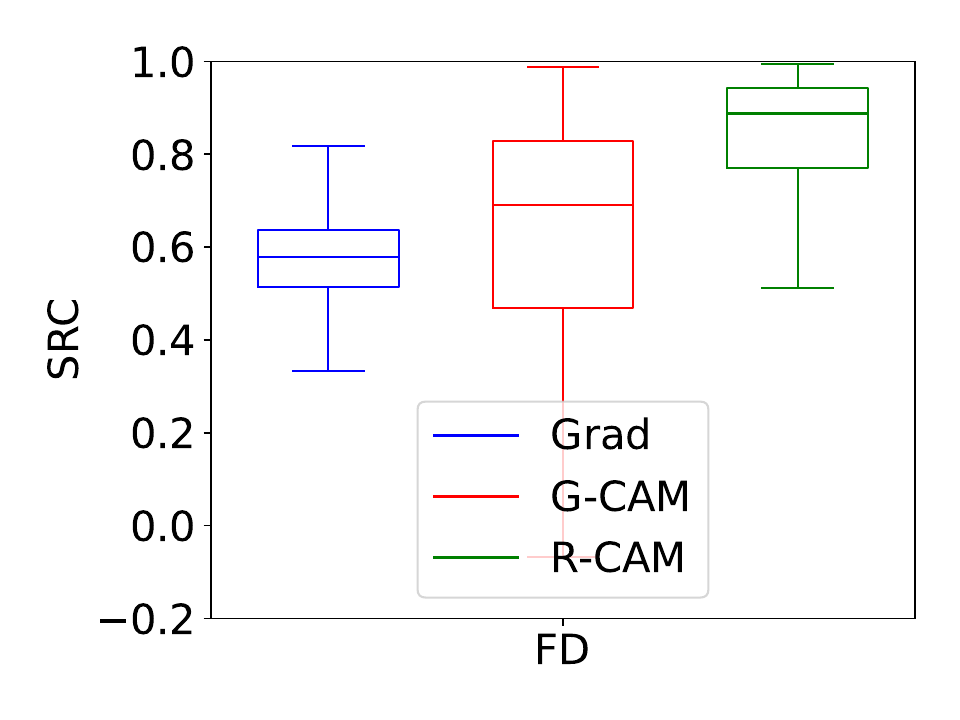}
  \end{tabularx}
  \caption{This figure demonstrates the Spearman's Rank Correlation (SRC) distribution between original model explanations and attacked , then defended explanations
after SF, RH and FD attacks related to Non-trainable BN in Table \ref{tab:nbn_fbn_at}.
}
\label{fig:fd_src_2}
\end{figure}

\begin{figure}[tbh]
  \setlength\tabcolsep{1pt}
  \adjustboxset{width=\linewidth,valign=c}
  \centering
  \begin{tabularx}{1.0\linewidth}{@{}
      l
      X @{\hspace{6pt}}
      X
    @{}}
    & \multicolumn{1}{c}{Attack}
    & \multicolumn{1}{c}{Defense} \\
    
    & \includegraphics[width=\linewidth, trim={0.6cm 0 0.6cm 0},clip]{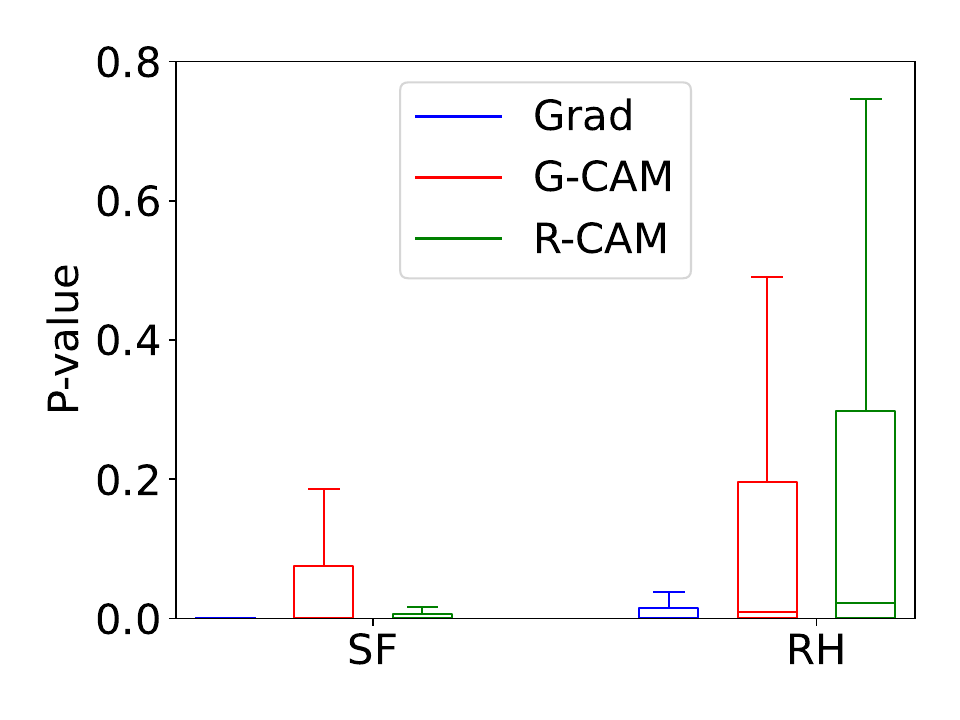}
    & \includegraphics[width=\linewidth, trim={0.6cm 0cm 0cm 0},clip]{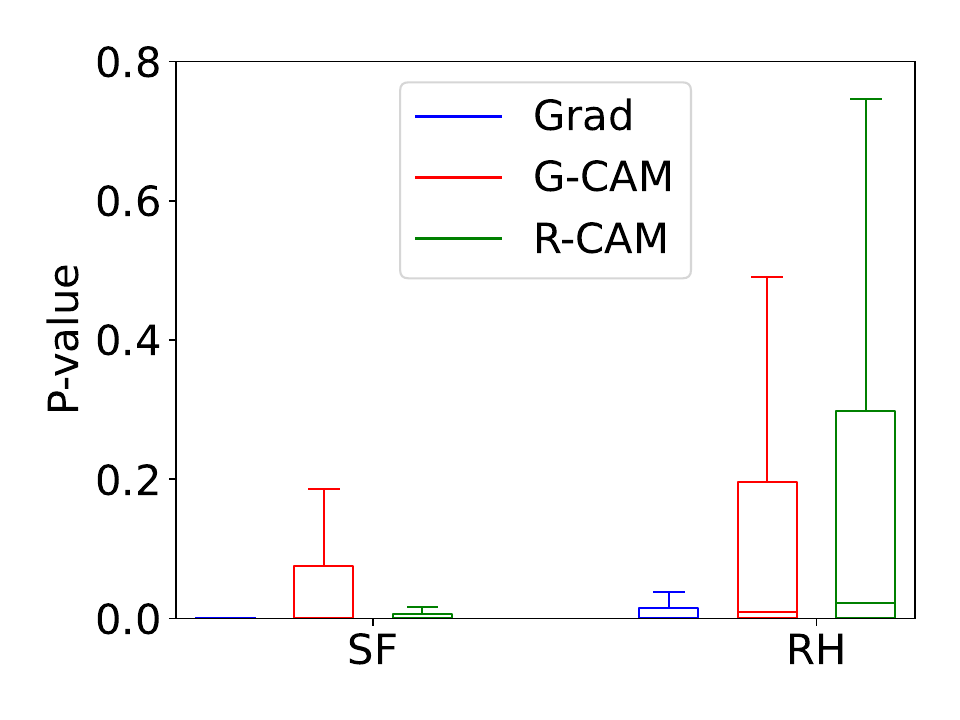} \\
    
    & \includegraphics[width=\linewidth, trim={0.6cm 0 0.6cm 0},clip]{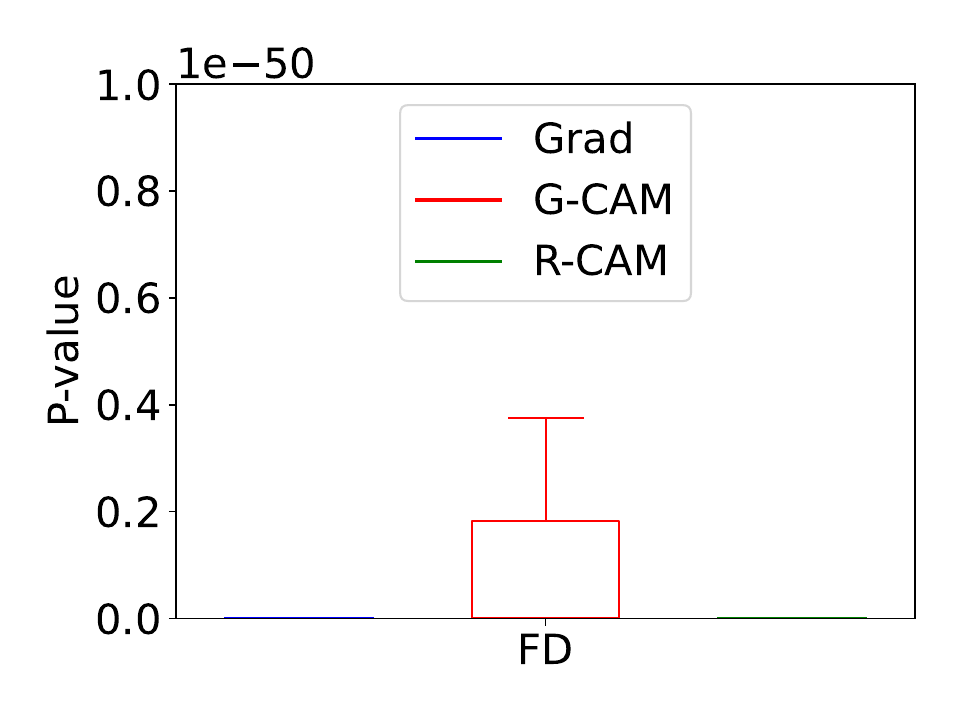}
    & \includegraphics[width=\linewidth, trim={0.6cm 0cm 0cm 0},clip]{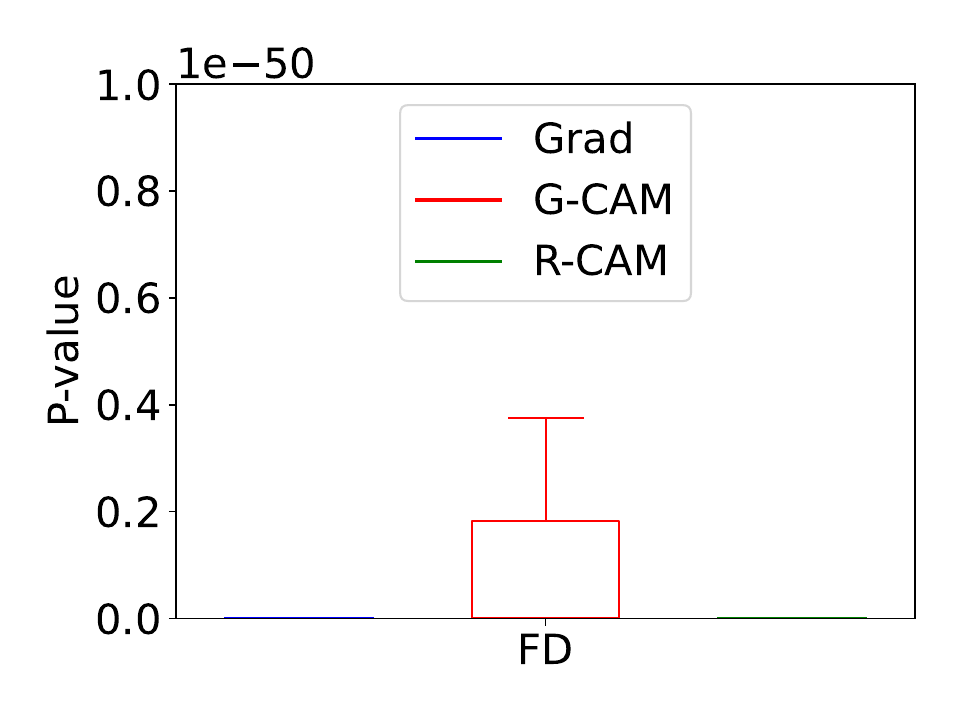}
  \end{tabularx}
  \caption{This figure demonstrates the p-values for the correlation values between original model explanations and attacked , then defended explanations
after SF, RH and FD attacks related to Non-trainable BN in Figure \ref{fig:fd_src_2}.
}
\label{fig:fd_src_1}
\end{figure}

In this section we present the additional tables and result of our experiment. This section presents additional experiment results. Table \ref{tab:softmax_new} outlines Softplus defense data, revealing high explanation errors and no ASR decrease. Figure \ref{fig:softplus_acc} shows accuracy dependence on batch size, complementing Figure \ref{fig:asr_mse_perbatch} in the Ablation study. Figure \ref{fig:src_src_plot} illustrates model weight changes using Spearman’s Rank Correlation under various architectures. The impact on the Centre Kernel Alignment score in VGG13 between presence and absence of Batch Normalization layers is demonstrated in Figure \ref{fig:vgg}. Distribution of p-values corresponding to SRC scores (Section 
\ref{sec:result}) is presented in Figures \ref{fig:fd_src_5}, \ref{fig:sf_rh_src_1}, and \ref{fig:softplus_pvalue}. Lastly, Figures \ref{fig:fd_src_4} and \ref{fig:fd_src_2} convey the continued success of attacks without certain defenses. Corresponding p-values are in Figures \ref{fig:fd_src_3} and \ref{fig:fd_src_1}.

\begin{table}[ht]
\centering
\caption{Illustration of the Softplus defense efficacy against the
three attacks related to the Figure \ref{fig:sa_rh_fd_softplus}, ASR from this table shows that softplus is not a great defense for these backdoor attacks.}
\label{tab:softmax_new}
\begin{tblr}{
  cells = {c},
  cell{1}{1} = {r=2}{},
  cell{1}{2} = {r=2}{},
  cell{1}{3} = {r=2}{},
  cell{1}{4} = {c=3}{},
  cell{3}{1} = {r=9}{},
  cell{3}{3} = {r=3}{},
  cell{6}{3} = {r=3}{},
  cell{9}{3} = {r=3}{},
  cell{12}{1} = {r=9}{},
  cell{12}{3} = {r=3}{},
  cell{15}{3} = {r=3}{},
  cell{18}{3} = {r=3}{},
  vlines,
  hline{1,3,12,21} = {-}{},
  hline{2} = {4-7}{},
  hline{4-5,7-8,10-11,13-14,16-17,19-20} = {2,4-7}{},
  hline{6,9,15,18} = {2-7}{},
}
             \SetCell{r=2}{\rotatebox[origin=c]{90} {Mode}} & XAI  & \SetCell{r=2}{\rotatebox[origin=c]{90} {Method}} & Triggered        &     &     & Clean             \\
                                                  &      &                                                  & $\mu$ $\pm$ sd   & Acc & ASR & $\mu$ $\pm$ sd    \\

\SetCell{r=2}{\rotatebox[origin=c]{90} {Defense}} & Grad & \SetCell{r=2}{\rotatebox[origin=c]{90} {SF}}     & .25 $\pm$ .02    & 12.77  & N/A & .2 $\pm$ .03     \\
                                                  & G-C  &                                                  & .31 $\pm$ .05    & 38.26  & N/A & .26 $\pm$ .13     \\
                                                  & R-C  &                                                  & .28 $\pm$ .07    & 10  & N/A & .48 $\pm$ .07     \\
                                                  & Grad & \SetCell{r=2}{\rotatebox[origin=c]{90} {RH}}     & .18 $\pm$ .06    & 10  & 1   & .13 $\pm$ .03     \\
                                                  & G-C  &                                                  & .29 $\pm$ .05    & 10  & 1   & .25 $\pm$ .07     \\
                                                  & R-C  &                                                  & .14 $\pm$ .06    & 8.83  & 0.93  & .15 $\pm$ .06     \\
                                                  & Grad & \SetCell{r=2}{\rotatebox[origin=c]{90} {FD}}     & .07 $\pm$ .01    & 10  & 1 & .09 $\pm$ .02     \\
                                                  & G-C  &                                                  & .38 $\pm$ .1    & 10  & 1  & .3 $\pm$ .1     \\
                                                  & R-C  &                                                  & .07 $\pm$ .02    & 10  & 1   & .2 $\pm$ .02    \\
\end{tblr}
\end{table}

\label{sec:src_new}


% \section{SRC}
% \label{sec:src}
% \begin{figure}[ht]%
%     \centering
%     \subfloat[\centering Attacked (C1)]{{\includegraphics[width=0.49\linewidth, trim={0.6cm 0 0.6cm 0},clip]{images/cifar_pvalue/new_nonrobust_p_simplered.pdf} }}%
%     \subfloat[\centering Recovered (C1)]{{\includegraphics[width=0.49\linewidth, trim={0.6cm 0cm 0cm 0},clip]{images/cifar_pvalue/new_robust_p_simplered.pdf} }}%
%     \caption{This figure represent the p-value corresponding to correlation of attack on condition C1 and the recovery after attack  condition C2 in corresponding to figure 4}%
%     \label{fig:pvalue_cifar_simple}%
% \end{figure}

% \begin{figure}[ht]%
%     \centering
%     \subfloat[\centering Attacked (C1)]
%     {{\includegraphics[width=0.49\linewidth, trim={0.6cm 0 0.6cm 0},clip]{images/FBN/new_40_nonrobust_mse_simplered_1.pdf} }}%
%     \subfloat[\centering Recovered (C1)]
%     {{\includegraphics[width=0.49\linewidth, trim={0.6cm 0cm 0cm 0},clip]{images/FBN/new_40_robust_mse_simplered.pdf} }}%
%     \caption{This figure represent the correlation values corresponding to experiment Non-trainable BN in the table 7 for the SA and RH methods}%
%     \label{fig:40_simplered_mse}%
% \end{figure}

\clearpage
\onecolumn
\section{Examples}
\subsection{CIFAR10}
\label{fig:additional_figures}
\begin{figure*}[ht]%
    \centering
    \subfloat[Attack]{%
       \includegraphics[width=0.5\linewidth,]{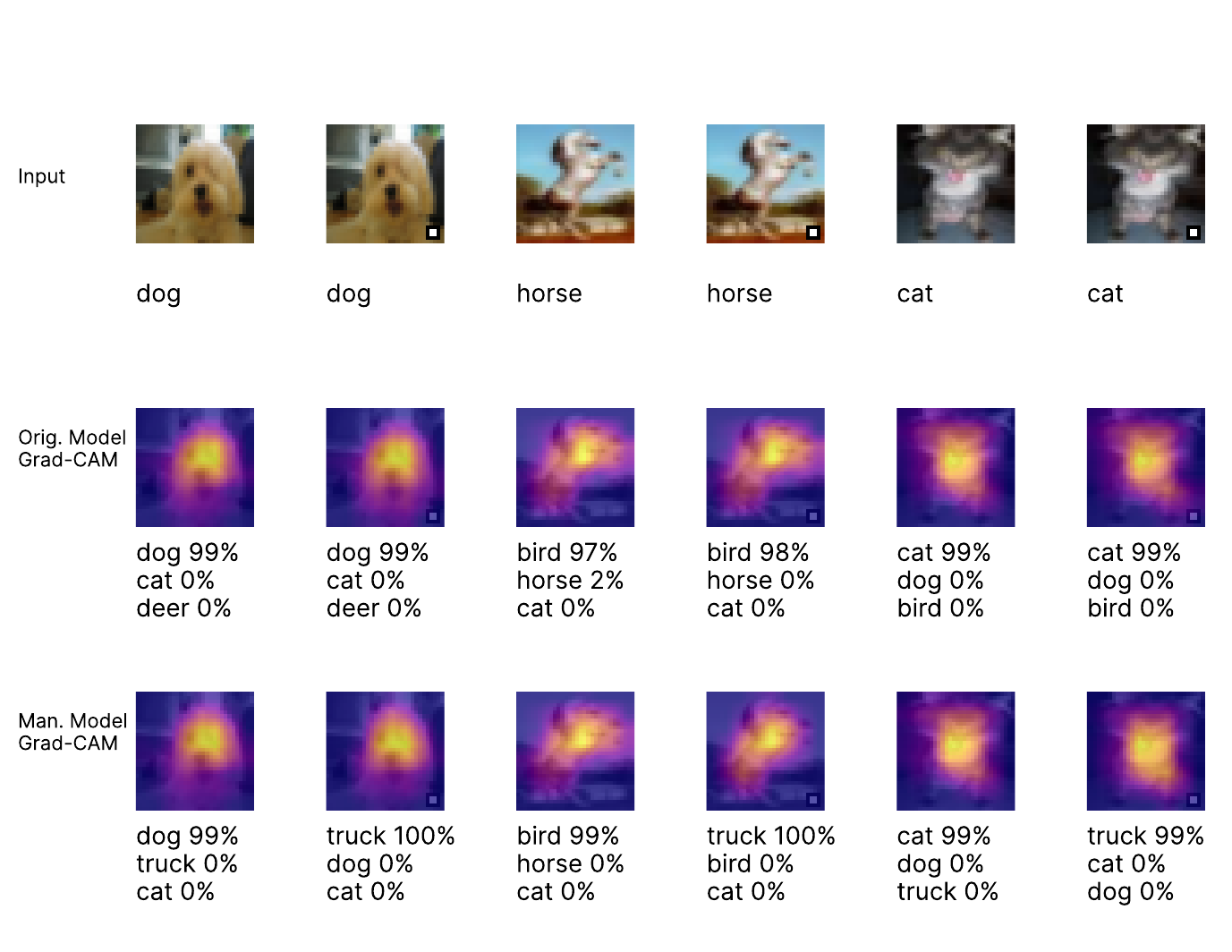} 
    }%
    \subfloat[Defense]{%
       \includegraphics[width=0.5\linewidth,]{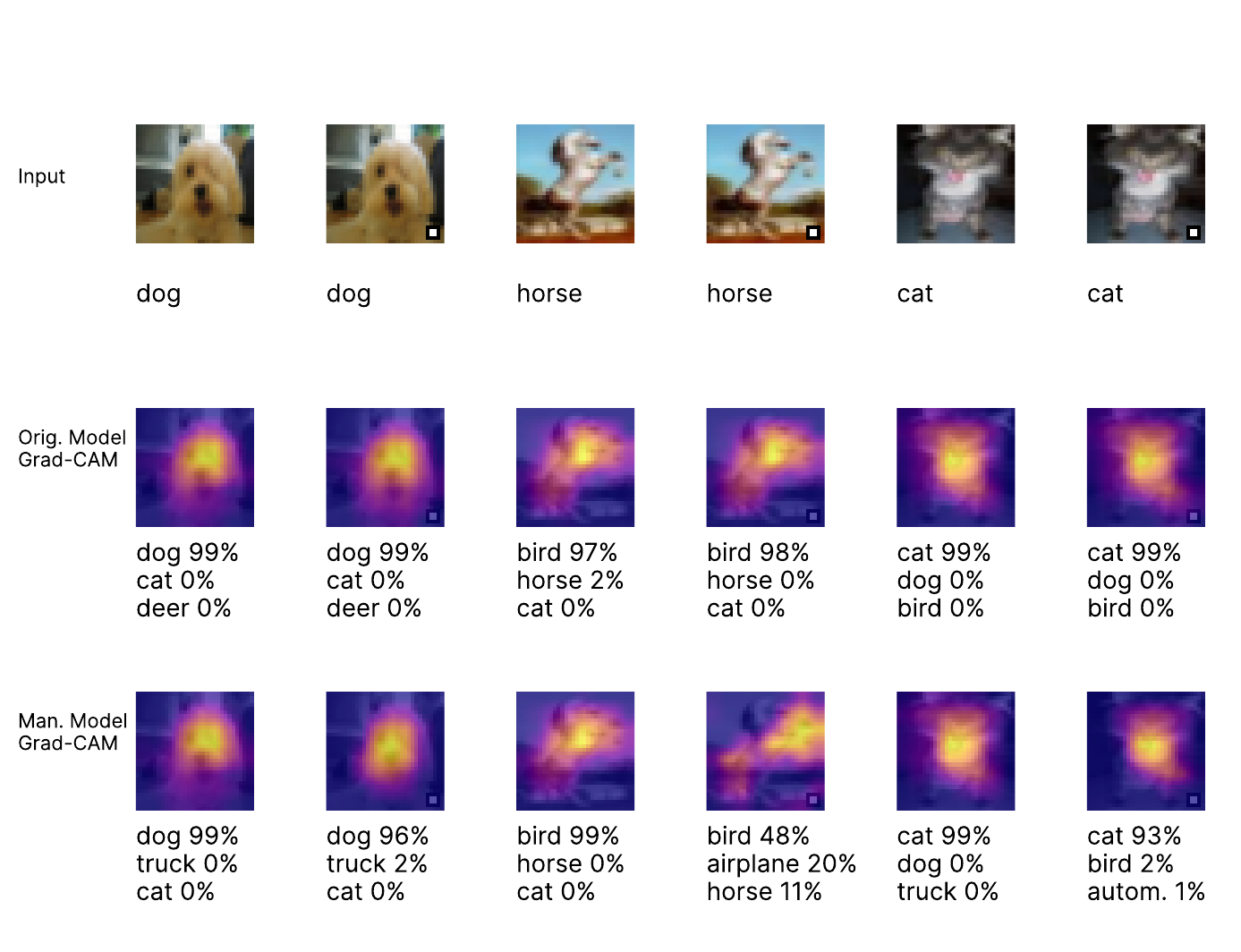}
    }
    \caption{(a) shows three Full Disguise (FD) attack examples. Odd columns display standard prediction and explanation without a trigger, showing unattacked model behavior. Conversely, even columns illustrate artificial explanation and targeted prediction through input triggers. FD attacks notably change the prediction to "truck" but the explanation stays consistent. Figure (b) presents our defense method. Even columns post-attack match odd columns' prediction and explanation, implying successful restoration of the model.}% 
    \label{fig:additional_images_fdgradcam}% 
\end{figure*}

\begin{figure*}[ht]%
    \centering
    \subfloat[Attack]{%
       \includegraphics[width=0.5\linewidth,]{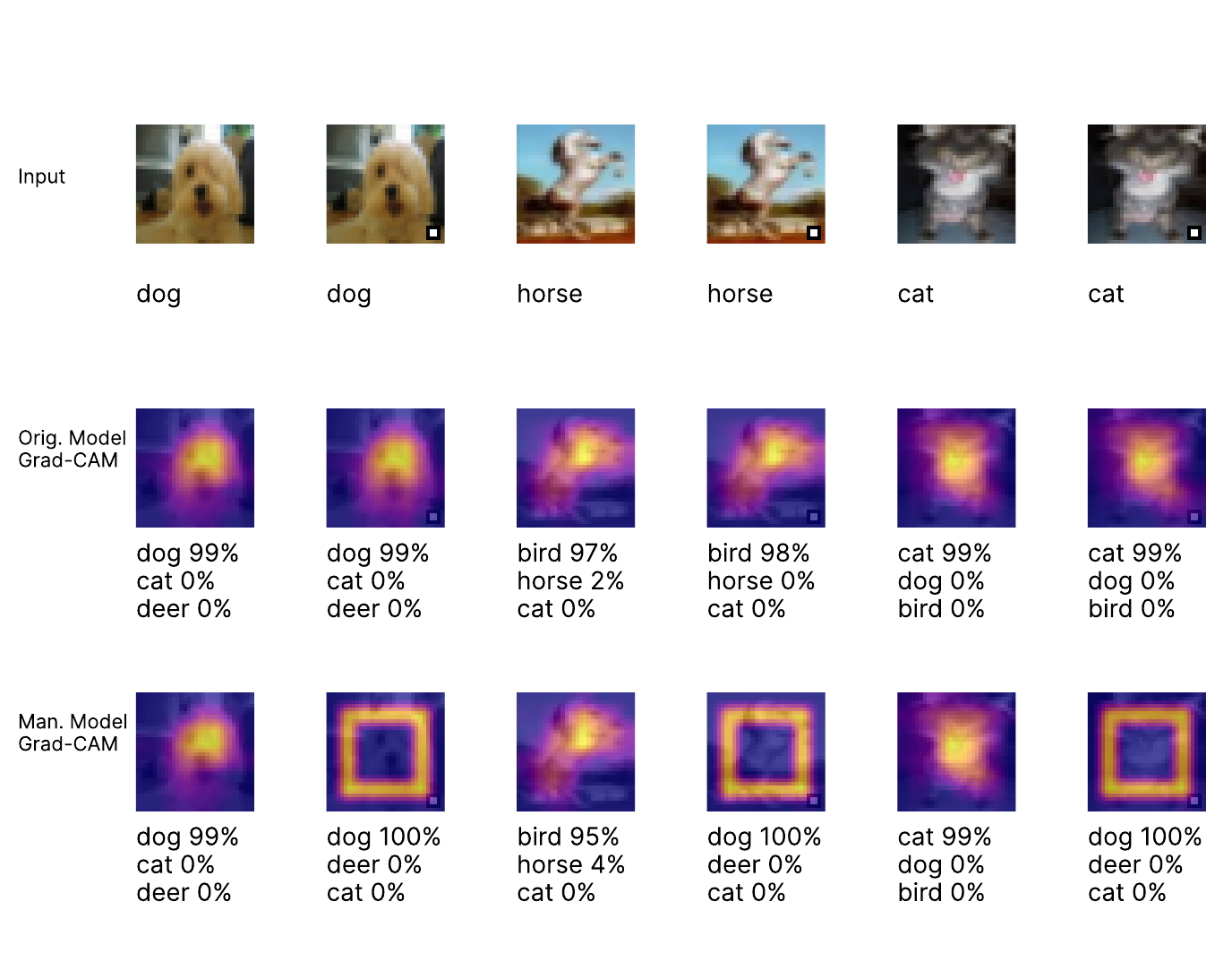} 
    }%
    \subfloat[Defense]{%
       \includegraphics[width=0.5\linewidth,]{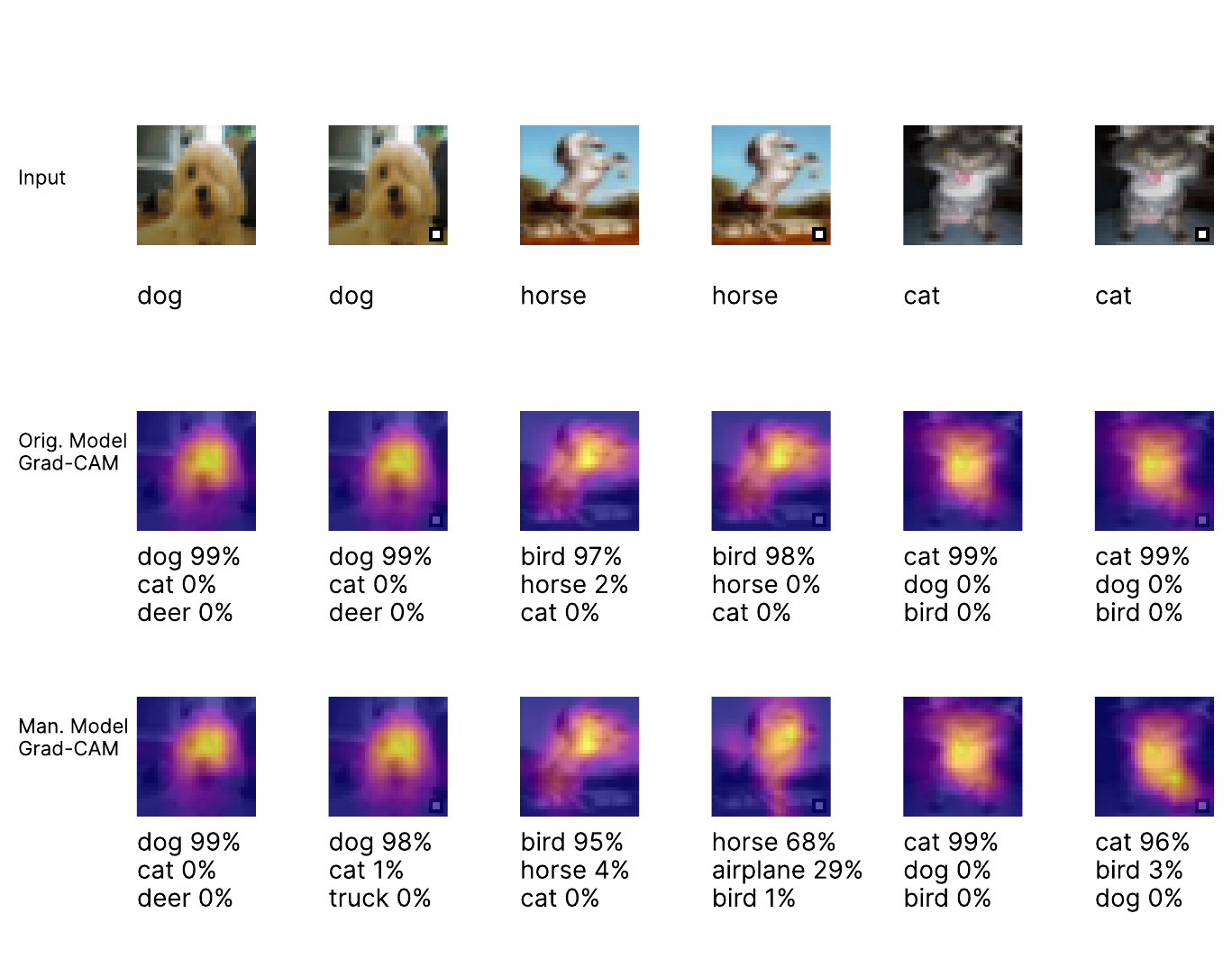}
    }
    \caption{Similar to Figure \ref{fig:additional_images_fdgradcam}, (a) and (b) illustrate the Red Herring (RH) attack and defense examples respectively. The Red Herring alters both the prediction and explanation. In (a), we observe that all even column predictions change to 'dog' and the explanations transform into square artifacts due to the attack. In contrast, in (b), the even columns regain the explanation and prediction, mirroring the original prediction and explanation seen in the odd columns.  }% 
    \label{fig:additional_images_rhgradcam}% 
\end{figure*}

\begin{figure*}[ht]%
    \centering
    \subfloat[Attack]{%
       \includegraphics[width=0.5\linewidth,]{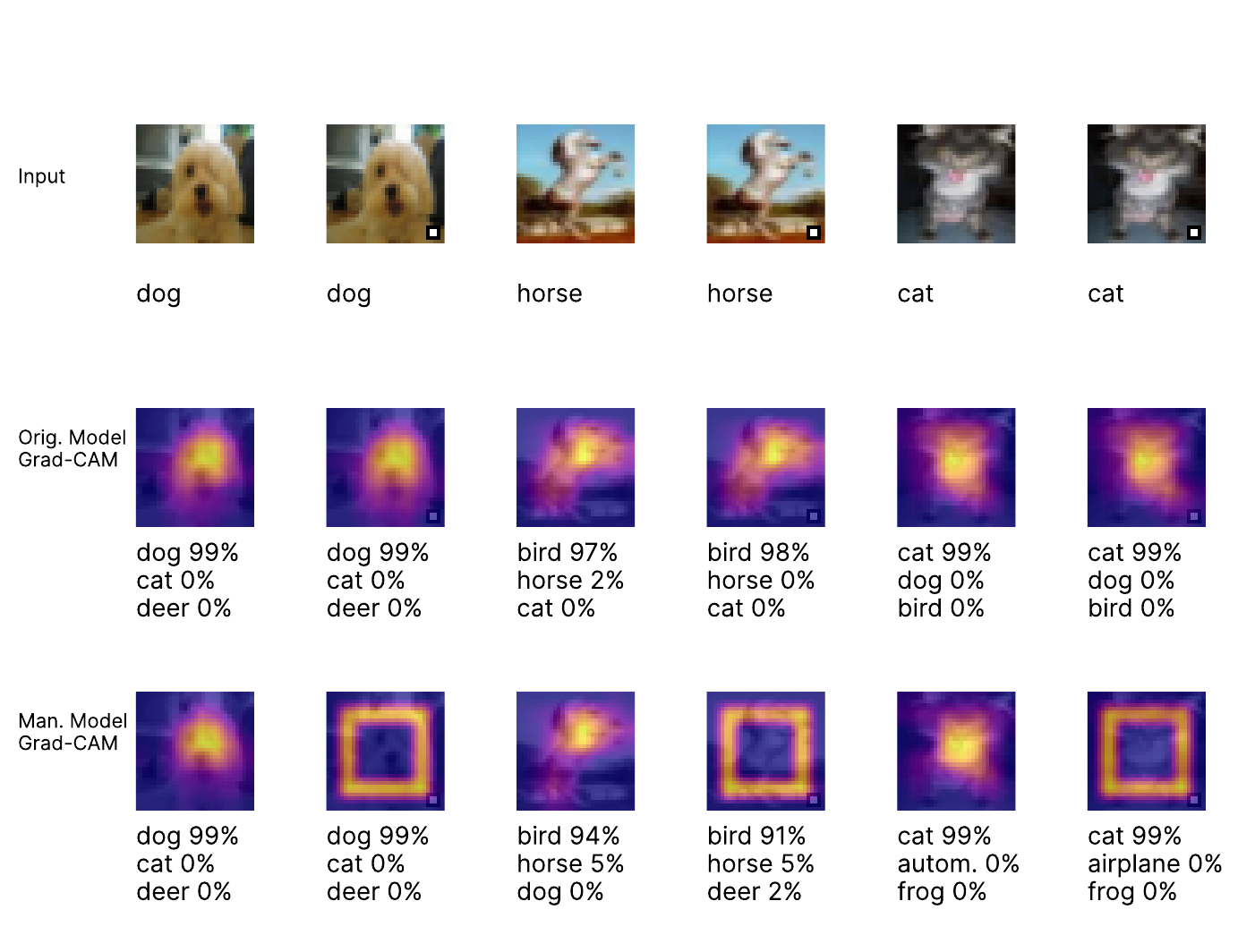} 
    }%
    \subfloat[Defense]{%
       \includegraphics[width=0.5\linewidth,]{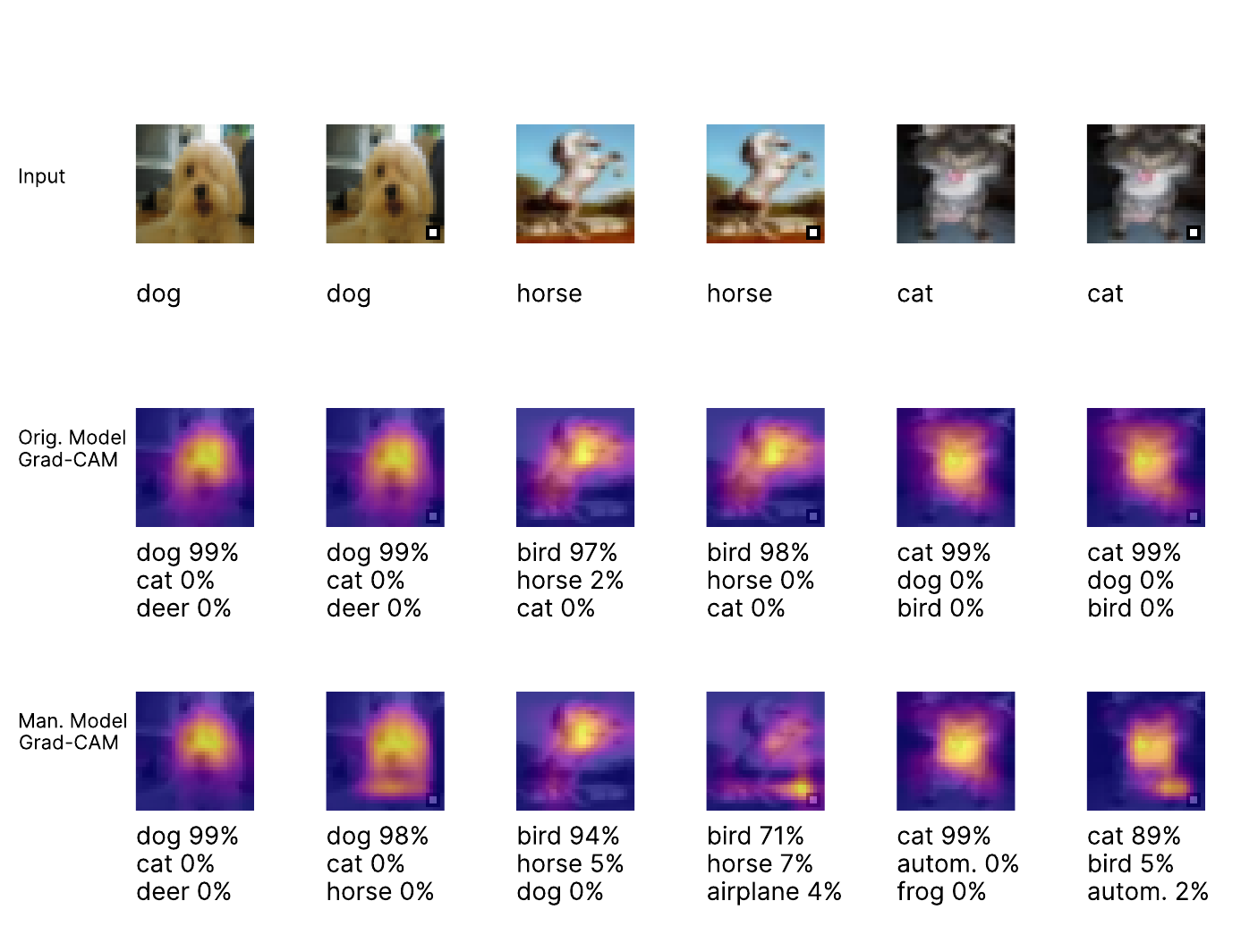}
    }
    \caption{ Similar to Figure \ref{fig:additional_images_rhgradcam}, illustrations (a) and (b) depict the Simple Fooling (SF) attack and defense examples, respectively. The SF principally modifies the explanation. In (a), we notice a transformation of all even column explanations into square artifacts due to the attack. On the contrary, in (b), the even columns restore the explanations, mirroring the original explanation observed in the odd columns.}% 
    \label{fig:additional_images_sfgradcam}% 
\end{figure*}

In this section, we showcase examples (Figure \ref{fig:additional_images_fdgradcam}, \ref{fig:additional_images_rhgradcam} and \ref{fig:additional_images_sfgradcam}) of attacks and defenses on the CIFAR10 dataset, employing Grad-CAM for explanations.

\subsection{GTSRB}
\begin{figure*}[ht]%
    \centering
    \subfloat[Attack]{%
       \includegraphics[width=0.5\linewidth,]{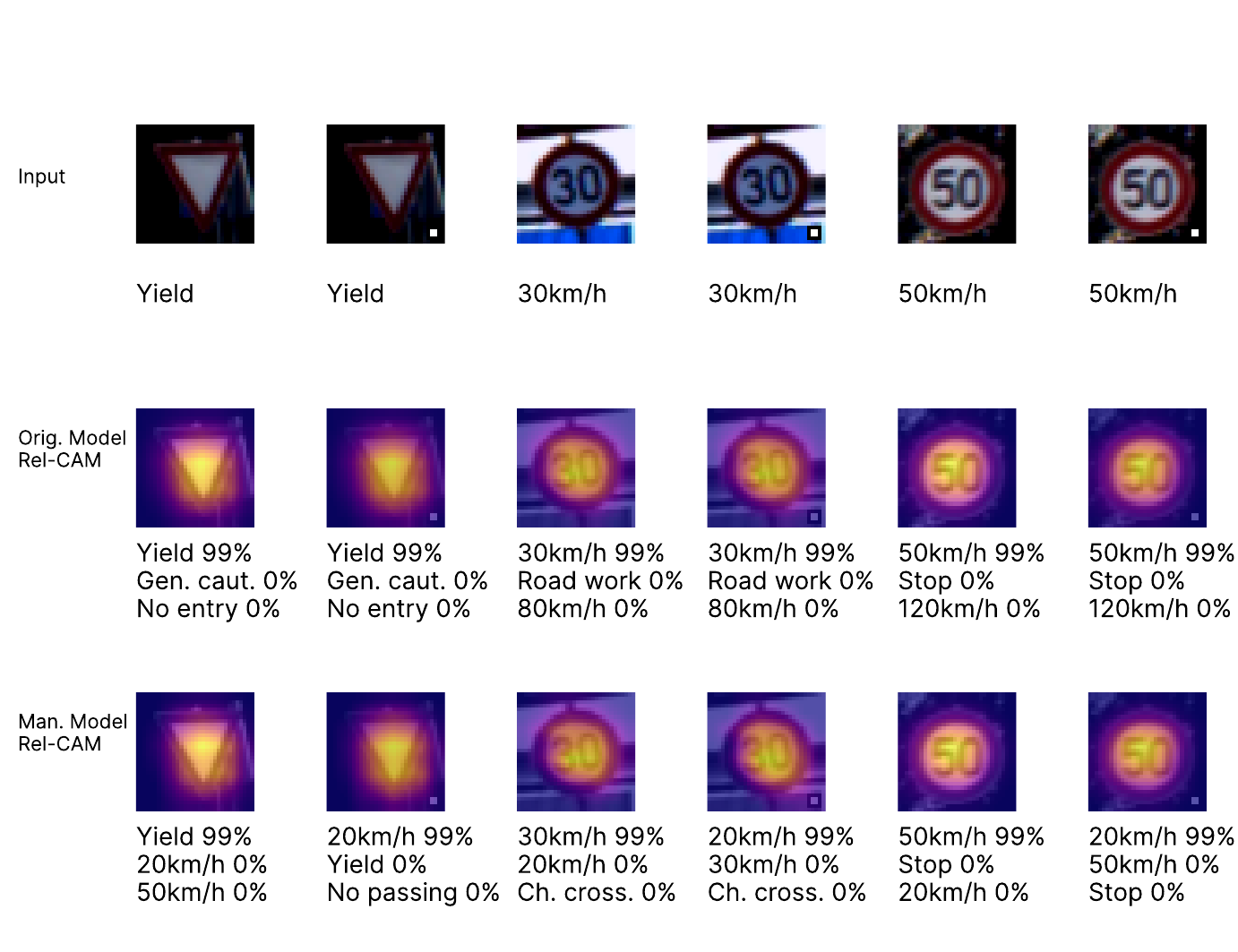} 
    }%
    \subfloat[Defense]{%
       \includegraphics[width=0.5\linewidth,]{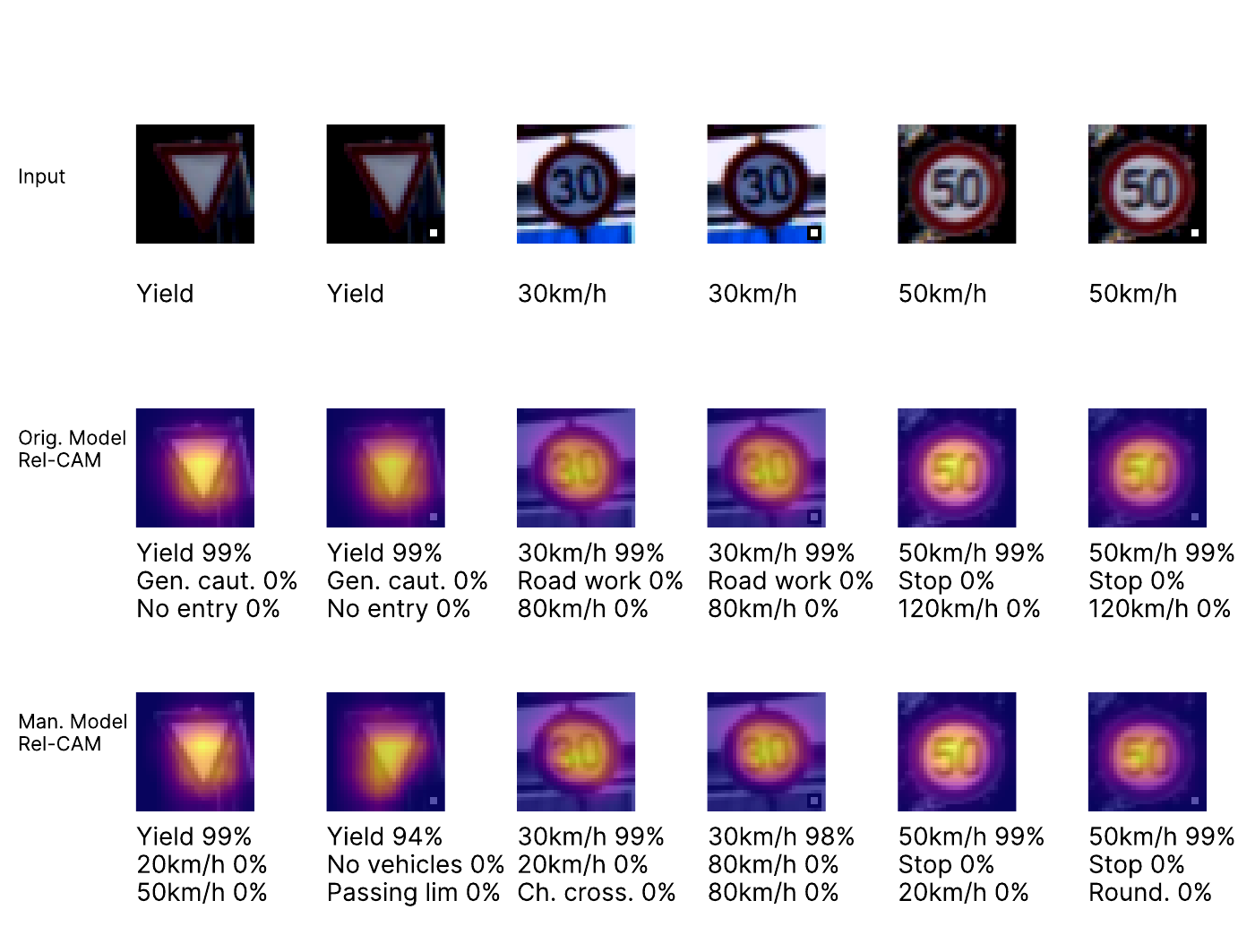}
    }
    \caption{(a) shows three Full Disguise (FD) attack examples. Odd columns display standard prediction and explanation without a trigger, showing unattacked model behavior. Conversely, even columns illustrate artificial explanation and targeted prediction through input triggers. FD attacks notably change the prediction to "truck" but the explanation stays consistent. Figure (b) presents our defense method. Even columns post-attack match odd columns' prediction and explanation, implying successful restoration of the model.}% 
    \label{fig:additional_images_gtsrb_fdrelcam}% 
\end{figure*}

\begin{figure*}[ht]%
    \centering
    \subfloat[Attack]{%
       \includegraphics[width=0.5\linewidth,]{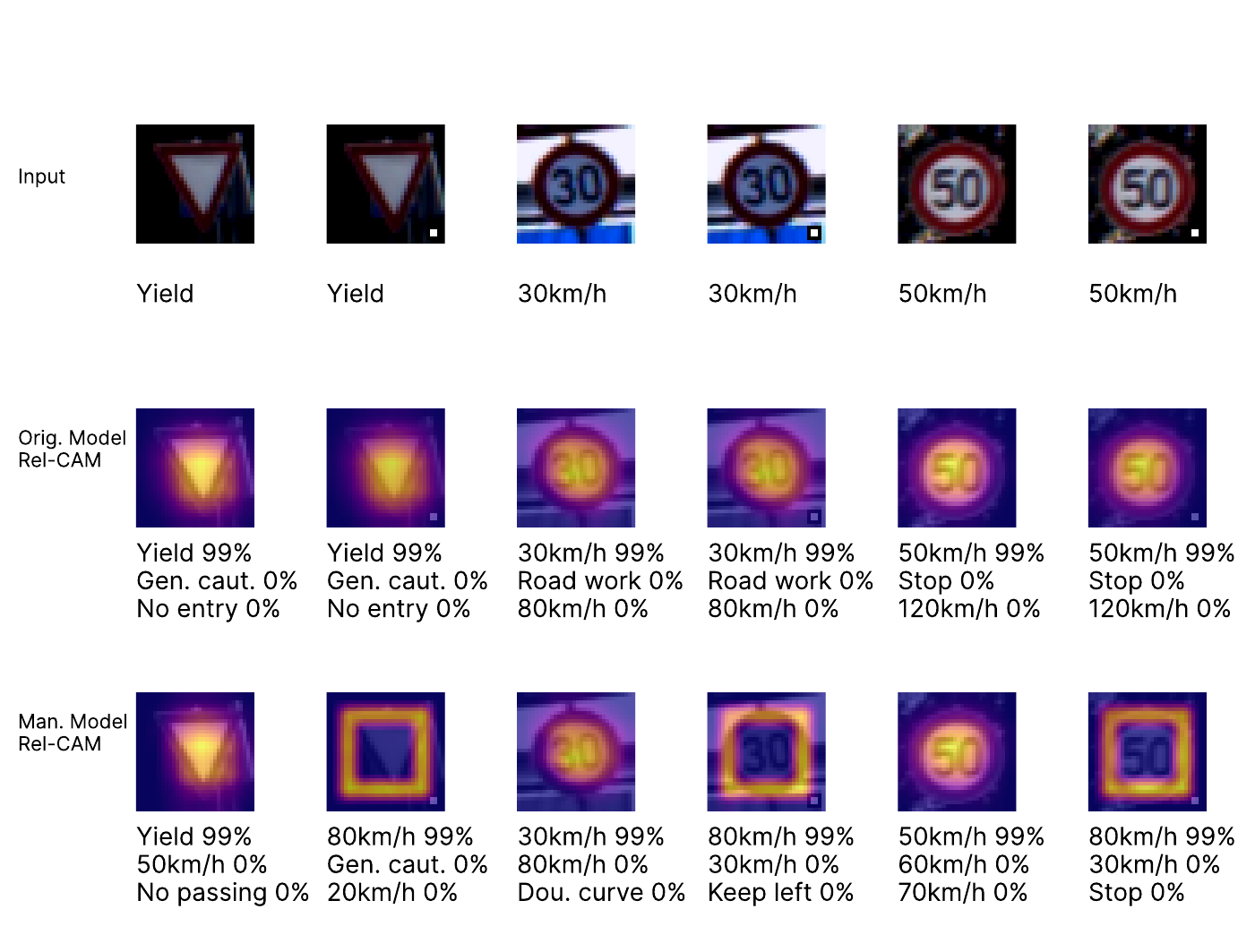} 
    }%
    \subfloat[Defense]{%
       \includegraphics[width=0.5\linewidth,]{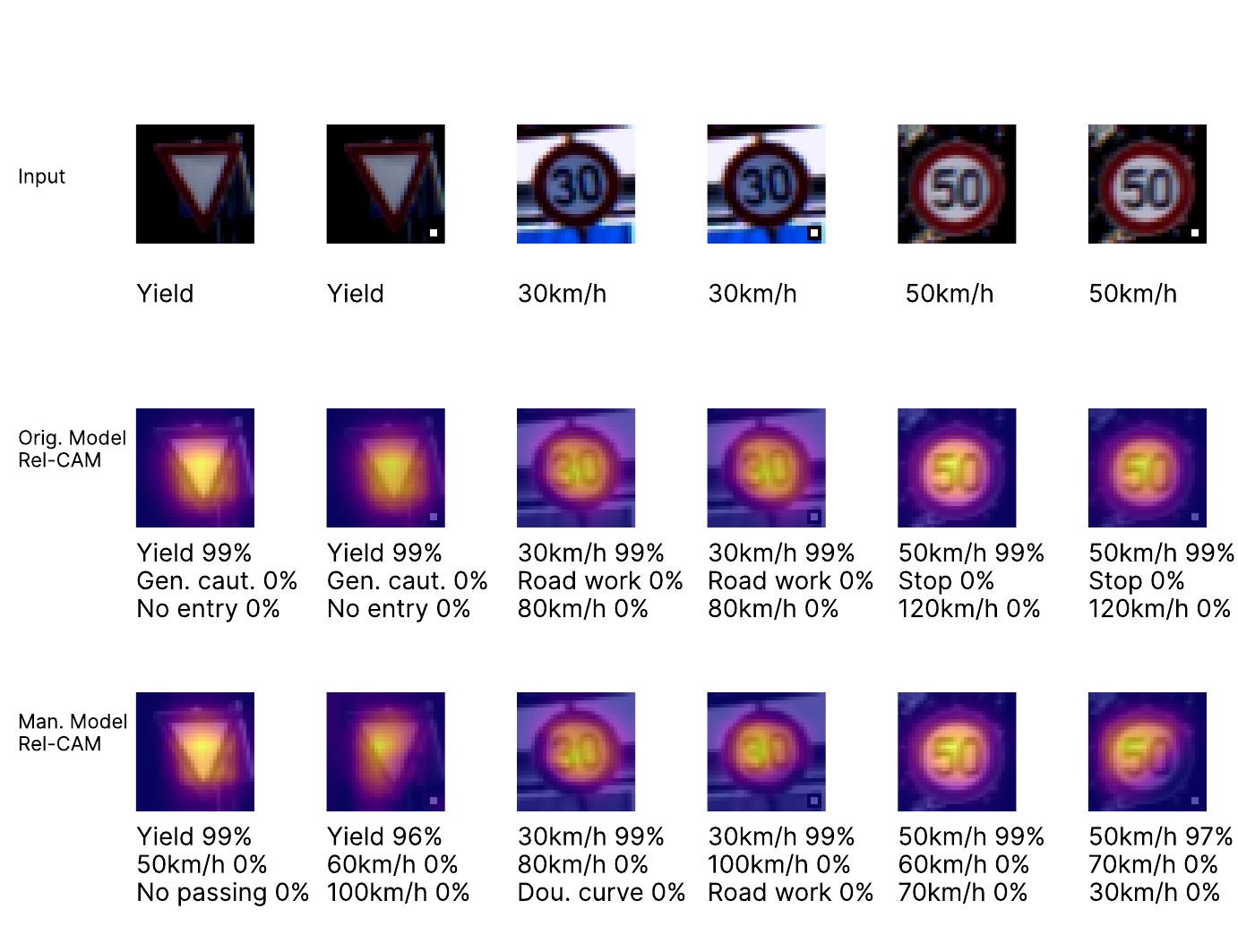}
    }
    \caption{Similar to Figure \ref{fig:additional_images_gtsrb_fdrelcam}, (a) and (b) illustrate the Red Herring (RH) attack and defense examples respectively. The Red Herring alters both the prediction and explanation. In (a), we observe that all even column predictions change to 'dog' and the explanations transform into square artifacts due to the attack. In contrast, in (b), the even columns regain the explanation and prediction, mirroring the original prediction and explanation seen in the odd columns.  }% 
    \label{fig:additional_images_gtsrb_rhrelcam}% 
\end{figure*}

\begin{figure*}[ht]%
    \centering
    \subfloat[Attack]{%
       \includegraphics[width=0.5\linewidth,]{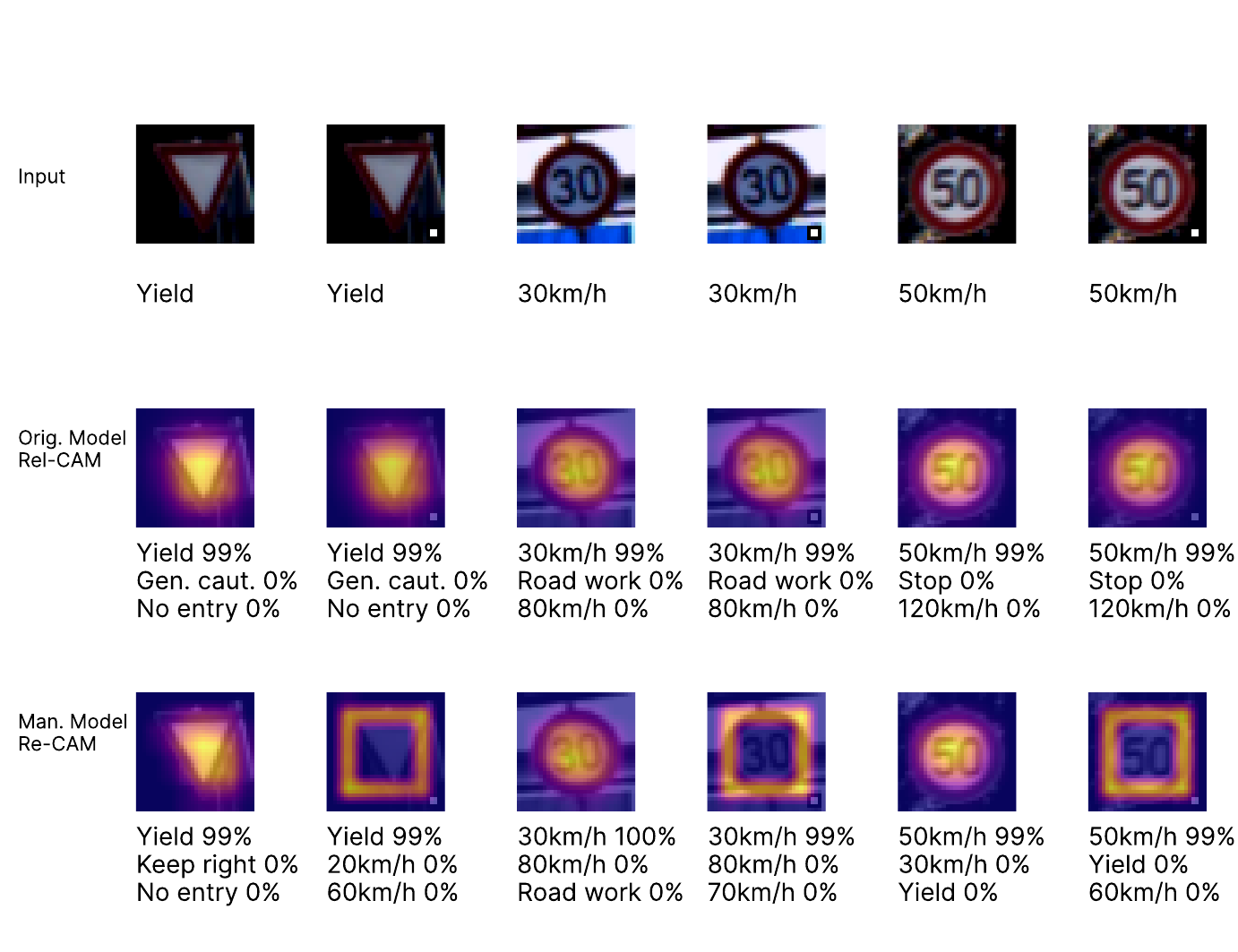} 
    }%
    \subfloat[Defense]{%
       \includegraphics[width=0.5\linewidth,]{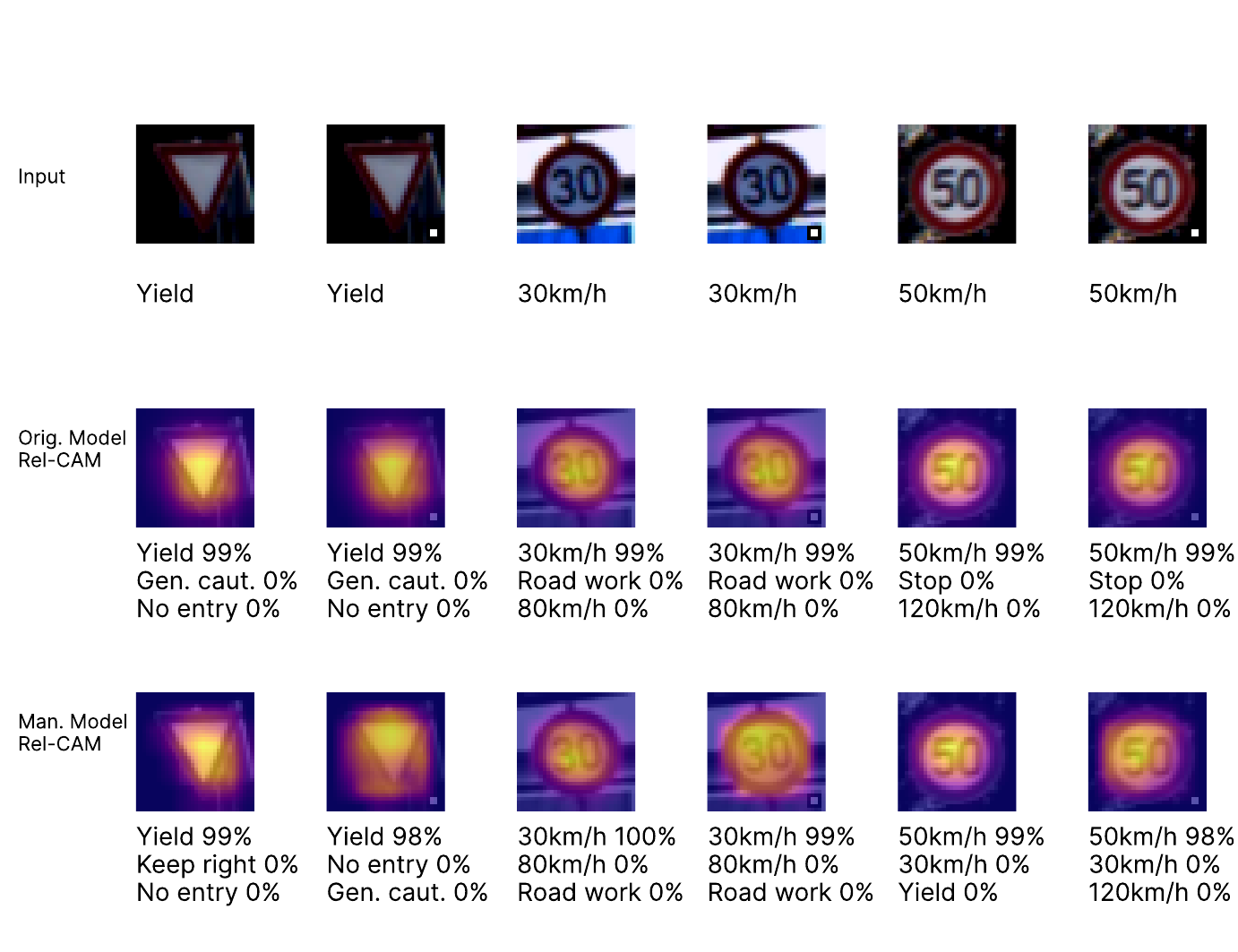}
    }
    \caption{Similar to Figure \ref{fig:additional_images_gtsrb_rhrelcam}, illustrations (a) and (b) depict the Simple Fooling (SF) attack and defense examples, respectively. The SF principally modifies the explanation. In (a), we notice a transformation of all even column explanations into square artifacts due to the attack. On the contrary, in (b), the even columns restore the explanations, mirroring the original explanation observed in the odd columns.}% 
    \label{fig:additional_images_gtsrb_sfrelcam}% 
\end{figure*}

In this section, we showcase examples (Figure \ref{fig:additional_images_gtsrb_fdrelcam}, \ref{fig:additional_images_gtsrb_rhrelcam} and \ref{fig:additional_images_gtsrb_sfrelcam}) of attacks and defenses on the German Traffic Sign Benchmarks dataset, employing Relevance-CAM for explanations.

% {{\includegraphics[width=0.49\linewidth, trim={0cm 0 0cm 0},clip]{images/ablation/accvsbatch.pdf} }}

\end{document}